\documentclass[10pt,twocolumn,letterpaper]{article}

\usepackage[pagenumbers]{cvpr} 

\usepackage[table]{xcolor}

\usepackage{pifont}
\newcommand{\cmark}{\ding{51}} 
\newcommand{\xmark}{\ding{55}} 

\usepackage{threeparttable}
\usepackage{dblfloatfix}
\usepackage{xspace}

\newcommand{\method}{VoxelFix\xspace}

\definecolor{cvprblue}{rgb}{0.21,0.49,0.74}
\usepackage[pagebackref,breaklinks,colorlinks,allcolors=cvprblue]{hyperref}

\def\paperID{126} 
\def\confName{3DV\xspace}
\def\confYear{2027\xspace}

\title{\method: Post-Hoc Semantic Correction of Completed 3D Voxel Maps}

\author{ Sunesh Praveen Raja Sundarasami\textsuperscript{1,2,*} \quad Taehyoung Kim\textsuperscript{1,*,\textdagger} \quad Johannes Scherer\textsuperscript{1} \quad Tomaž Cotič\textsuperscript{1,3} \\ Sivasubiramaniam Subbiah\textsuperscript{1,4} \quad Andreas Greiner\textsuperscript{1,5} \quad Paul Spannaus\textsuperscript{1} \quad Sebastian Houben\textsuperscript{2} \\[0.8em] \textsuperscript{1}Fraunhofer IVI \quad \textsuperscript{2}Hochschule Bonn-Rhein-Sieg \quad \textsuperscript{3}University of Bologna \quad \textsuperscript{4}FAU \quad \textsuperscript{5}THI }

\begin{document}
\maketitle

\begingroup \renewcommand{\thefootnote}{\fnsymbol{footnote}} \footnotetext[1]{Equal contribution.} \footnotetext[2]{Corresponding author: \texttt{taehyoung.kim@ivi.fraunhofer.de}} \endgroup

\begin{abstract}
Semantic 3D maps are increasingly constructed automatically for aerial robotics by integrating learned semantic predictions into 3D representations. While this avoids costly manual 3D annotation, errors in the perception and mapping pipeline can persist in the resulting map, reducing its reliability for downstream autonomous tasks. Existing 3D semantic map refinement methods either rely on the original observations, treat occupancy as part of the prediction problem, or apply non-learned local regularization to completed maps. Instead, we study post-hoc semantic correction, asking whether semantic accuracy can be recovered directly from the completed map while keeping its geometry and occupancy fixed. We introduce \method, a graph-based model that corrects voxel labels based on local geometry and neighboring semantic information. To obtain training pairs, we corrupt contiguous regions of annotated OccuFly maps according to class confusions observed in upstream maps. We evaluate \method on completed OccuFly maps generated from predictions of four independently trained 2D segmentation models. \method consistently improves mIoU by 4.23--5.00 percentage points, with gains broadly distributed across the evaluated semantic classes and particularly strong improvements for tree, roof, and wall. Results on an independently reconstructed out-of-distribution aerial scene further suggest that the learned correction can transfer beyond the environments seen during training.
\end{abstract}    
\section{Introduction}
\label{sec:intro}

\begin{figure*}[t]
    \centering
    \includegraphics[
        width=\textwidth]{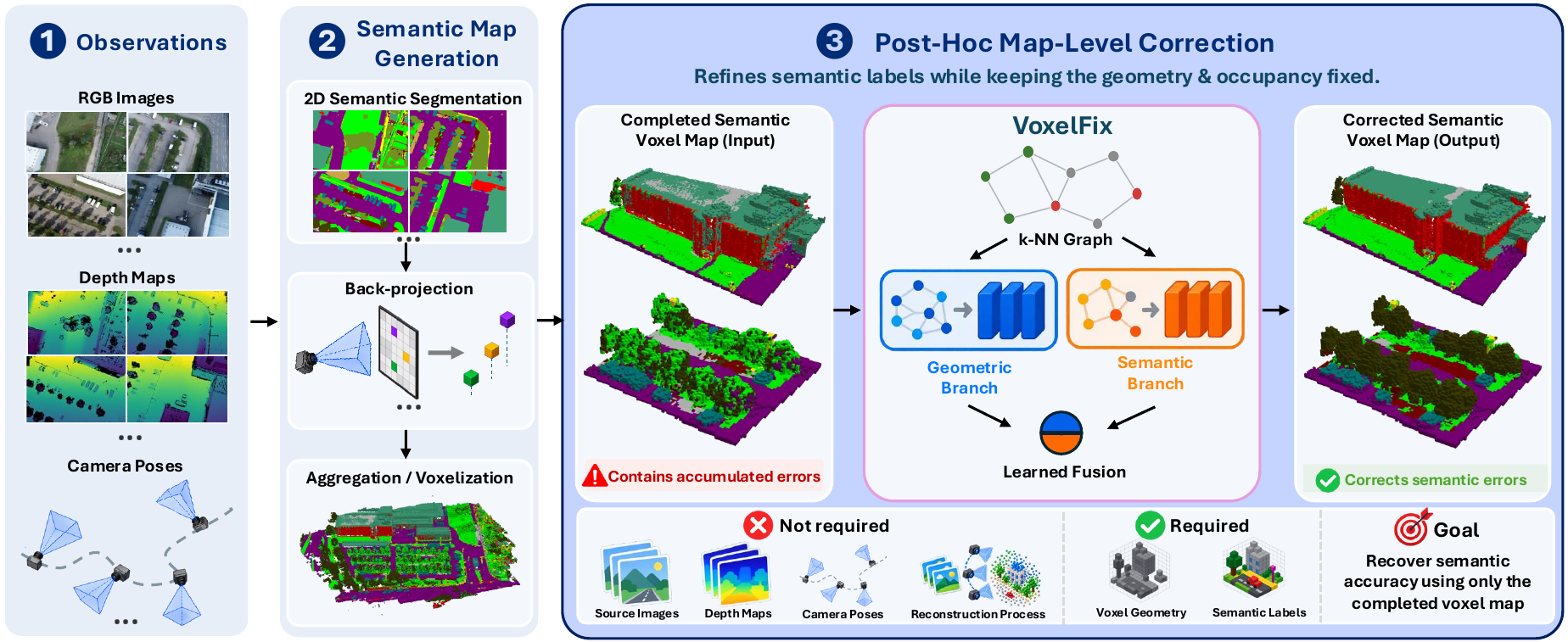}
    \vspace{-0.75em}
    \caption{\textbf{Map-only post-hoc semantic correction.} One common semantic mapping pipeline predicts 2D semantics and back-projects them using depth and camera poses before aggregating the observations into a semantic voxel map. Upstream perception and mapping errors can persist in the completed map. Our setting begins after map construction: source images, depth maps, camera poses, and intermediate mapping state are unavailable, while voxel geometry and occupancy remain fixed. \method uses only the completed voxel geometry and semantic labels to correct inconsistent semantics.}
    \label{fig:overview}
\end{figure*}

Autonomous aerial operation near the ground depends on persistent three-dimensional maps with geometric and semantic information \cite{liu2022large,liu2023active}. Semantic voxel maps represent occupied space as voxels, each with a semantic label \cite{gross2026occufly}. These labels affect downstream tasks; for example, an obstacle-aware planner and a landing-site assessment module treat the same geometry differently when labeled as vegetation, wall, or traversable ground.

High-quality semantic maps can be obtained through manual 3D annotation or by transferring manually annotated 2D labels to reconstructed 3D geometry, but both require substantial annotation effort; SemanticKITTI alone required more than 1,700 annotation hours~\cite{behley2019semantickitti}. A more scalable alternative is to derive map semantics from learned predictions, for example, by lifting 2D semantic predictions into 3D using depth and camera poses~\cite{mccormac2017semanticfusion} or projecting them onto registered LiDAR observations~\cite{jeong2018towards}. Such modular pipelines reduce the need for manual semantic annotation, but errors in perception and mapping can remain in the completed map. Once map construction is finished, the original observations and intermediate mapping state may also no longer be available. We consider this setting, in which only the completed semantic voxel map is available, and its geometry and occupancy remain fixed.

Existing refinement methods do not directly address this setting. During map construction, predictions can be refined by integrating them during fusion~\cite{gan2020bayesian,wilson2024convbki}, revisiting sequential observations~\cite{sun2018recurrent,momma2020p}, or exploiting consistency across multiple views~\cite{shi2025offboard}. These approaches require information from the mapping process that is no longer available in our setting. Learned semantic scene completion refinement operates directly in voxel space, but jointly predicts semantics and occupancy~\cite{zhang2026enhancing}. Traditional post-processing can be applied to a completed map, but typically regularizes labels through local neighborhood or smoothness criteria~\cite{hermans2014dense,jeong2018towards}. This leaves a complementary problem largely unexplored: \emph{correcting semantic labels directly from a completed map without modifying its geometry or occupancy}.

Local smoothing may be ineffective when errors span larger regions. For example, a roof region mislabeled as wall may remain unchanged when neighboring voxels share the same incorrect label. The completed map, however, still contains spatial structure and class information that can help identify such errors.

We therefore formulate post-hoc semantic correction as a voxel relabeling problem and introduce \method, a graph-based model that learns to identify and correct inconsistent semantic labels. Since paired clean and corrupted maps are not naturally available, we synthesize spatially coherent training errors from voxel-level confusion patterns of the upstream semantic predictions. At inference, \method operates solely on the completed semantic voxel map. An overview is shown in \cref{fig:overview}.

Our \textbf{contributions} are summarized as follows:
\begin{itemize}
\item We formulate map-only post-hoc semantic correction for completed aerial semantic voxel maps, refining semantics while preserving geometry and occupancy.
\item We propose \textbf{\method}, a dual-branch graph attention architecture that reasons over complementary geometric and semantic neighborhoods and integrates both through learned gated fusion.
\item We introduce a confusion-guided noise curriculum that synthesizes spatially coherent, class-dependent errors from the observed confusion patterns of upstream semantic predictions.
\item We demonstrate consistent improvements of $4.23$--$5.00$ mIoU percentage points across four upstream segmentation models on the held-out OccuFly test scene, and further evaluate transfer to an independently reconstructed OOD scene.
\end{itemize}


\section{Related work}
\label{sec:related}

\noindent\textbf{Semantic Voxel Map Construction.}
Semantic voxel maps enrich geometric 3D representations with semantic labels. A common construction strategy lifts per-frame semantic predictions into 3D using depth and camera pose and aggregates them across observations \cite{mccormac2017semanticfusion,rosinol2020kimera}. Large-scale aerial semantic datasets also provide human-annotated semantics on photogrammetric or LiDAR point clouds \cite{can2021semantic,hu2022sensaturban,wang2025uavscenes}, while synthetic scenes \cite{chen2022stpls3d} and transferred 2D annotations \cite{gross2026occufly} provide alternative sources of semantic 3D data.

\noindent\textbf{Improving Semantic Map Quality.} Once semantic errors are present in a map, existing methods differ mainly in when correction is performed and what information they require. \cref{tab:related} compares these settings. Traditional post-processing can operate directly on completed 3D maps: local neighborhood refinement targets semantic inconsistencies \cite{jeong2018towards,wei2020semantic}, while CRFs \cite{hermans2014dense} and graph-based optimization \cite{huang2019semantic} regularize labels using spatial and appearance relationships. These methods are directly applicable to our setting, but rely on predefined neighborhoods, similarity measures, or smoothness criteria rather than learned error patterns.

Other approaches improve semantic quality during map construction. Bayesian kernel inference exploits spatial correlations between neighboring observations \cite{gan2020bayesian}, ConvBKI introduces learnable convolutional fusion~\cite{wilson2024convbki}, and uncertainty-aware fusion methods account for the confidence of individual semantic observations~\cite{kim2024evidential}. These methods require access to the observations being fused and therefore cannot operate on a completed map alone.

Learning-based refinement similarly often relies on information available during map construction. Recurrent semantic mapping, for example, updates the map over sequential observations \cite{sun2018recurrent}, P2Net exploits consistency across consecutive registered LiDAR frames \cite{momma2020p}, and OccFiner uses multiple frames and views to refine semantic scene completion \cite{shi2025offboard}. ESSC-RM is closer to our setting because it refines coarse semantic occupancy directly in voxel space \cite{zhang2026enhancing}, but its input is a semantic scene completion (SSC) prediction in which occupancy and semantics are jointly predicted. We instead refine only the semantic labels of a completed map with fixed occupancy.

\noindent\textbf{Graph-Based 3D Reasoning.} Graph neural networks provide a natural representation for irregular 3D data by connecting elements through spatial or feature-based neighborhoods. Superpoint Graph reasons over geometrically homogeneous regions~\cite{landrieu2018large}, while DGCNN dynamically constructs neighborhoods in learned feature space~\cite{wang2019dynamic}. Graph attention further learns the relative contribution of neighboring nodes during message passing~\cite{velivckovic2017graph,brody2021attentive}. We use spatial and feature-based neighborhoods with graph attention to correct semantic labels in completed voxel maps.

\noindent\textbf{Error Modeling for Semantic Refinement.} 
Class-dependent label noise is commonly modeled using a transition matrix that describes the probability of one class being observed as another \cite{patrini2017making,hendrycks2018using}. For dense prediction, however, errors can extend over contiguous regions rather than occur as independent label flips \cite{ye2022robust, tadokoro2025bayesian, yao2023learning}. We use voxel-level confusion patterns to guide spatially coherent class substitutions during training.

\begin{table}[t]
    \centering
    \begin{threeparttable}
    \caption{\textbf{Comparison of approaches for improving semantic map quality.}
    \textit{No Obs.} and \textit{No Pose} indicate that original observations and pose information are not required;
    \textit{Map} denotes operation on a completed semantic map, and
    \textit{Fixed} indicates that geometry and occupancy are not modified.}
    \label{tab:related}

    \small
    \setlength{\tabcolsep}{2.5pt}
    \renewcommand{\arraystretch}{1.05}

    \begin{tabular}{lccccc}
        \toprule
        Method & No Obs. & No Pose & Map & Fixed & Learned \\
        \midrule
        SemanticFusion \cite{mccormac2017semanticfusion}
            & \xmark & \xmark & \xmark & \cmark & \xmark \\
        P2Net \cite{momma2020p}
            & \xmark & \xmark & \xmark & \cmark & \cmark \\
        OccFiner \cite{shi2025offboard}
            & \xmark & \xmark & \xmark & \xmark & \cmark \\
        ESSC-RM \cite{zhang2026enhancing}
            & \cmark & \cmark & \xmark\tnote{\dag} & \xmark & \cmark \\
        \midrule
        CRF (baseline)
            & \cmark & \cmark & \cmark & \cmark & \xmark \\
        KNN (baseline)
            & \cmark & \cmark & \cmark & \cmark & \xmark \\
        \midrule
        \textbf{Ours}
            & \cmark & \cmark & \cmark & \cmark & \cmark \\
        \bottomrule
    \end{tabular}

    \begin{tablenotes}[flushleft]
        \footnotesize
        \item[\dag] Operates on a coarse SSC prediction, not a completed semantic map.
    \end{tablenotes}
    \end{threeparttable}
\end{table}
\section{Method}
\label{sec:method}

\subsection{Problem Formulation}
\label{sec:problem}

We consider a completed semantic voxel map produced by an upstream mapping process. The map contains $N$ occupied voxels on a regular grid with voxel size $r_v$,
\begin{equation}
\mathcal{M} =
\bigl\{(\mathbf{c}_i,\hat{y}_i)\bigr\}_{i=1}^{N},
\qquad
\mathbf{c}_i \in \mathbb{Z}^3,\;\;
\hat{y}_i \in \mathcal{C},
\label{eq:map}
\end{equation}
where $\mathbf{c}_i$ is the grid coordinate of voxel $i$, $\mathcal{C}=\{1,\dots,K\}$ is the set of $K$ semantic classes, and $\hat{y}_i$ is its upstream-assigned label. In our experiments, these labels are obtained by lifting 2D semantic predictions into 3D and aggregating them across views. We denote the ground-truth and corrected labels by $y_i$ and $\tilde{y}_i$, respectively.

Two constraints define our setting. First, $\mathcal{M}$ is the only available input: once reconstruction is complete, the observations and fusion state are no longer accessible. Second, geometry and occupancy are fixed, so any performance change arises solely from semantic relabeling. We therefore seek a correction function $f_\theta$ with learnable parameters $\theta$,
\begin{equation}
    f_\theta:\; \mathcal{M} \;\longmapsto\; \{\tilde{y}_i\}_{i=1}^{N},
    \qquad \tilde{y}_i \in \mathcal{C}.
    \label{eq:task}
\end{equation}

\subsection{Voxel Graph Representation}
\label{sec:graph}

A completed map is an irregular, sparse set of occupied cells. Hence, dense volumetric convolution wastes computation on empty space, and a fixed 26-neighborhood stencil ties the receptive field to the grid resolution. We instead use an attributed graph $\mathcal{G}=(\mathcal{V},\mathcal{E})$, as is common for point cloud segmentation \cite{wang2019graph,li2020tgnet,landrieu2018large}. The node set $\mathcal{V}$ holds one node per occupied voxel, and the edge set $\mathcal{E}$ connects each node to its $k$ nearest neighbors under Euclidean distance between voxel centers. Because occupancy is
fixed, $\mathcal{E}$ is computed once per map and reused at every forward pass.

\noindent\textbf{Node attributes.}
Each voxel is represented by a six-dimensional geometric descriptor $\mathbf{g}_i \in \mathbb{R}^{6}$ and a learned geometry embedding $\mathbf{z}_i \in \mathbb{R}^{16}$. The geometric descriptor is derived from the structure tensor, computed as the covariance matrix of the positions of the $k_{\text{geo}}$ nearest neighbors, with eigenvalues $\lambda_1 \!\geq\!
\lambda_2 \!\geq\! \lambda_3$ normalized to sum to one, and with $\mathbf{n}_i$ the eigenvector of $\lambda_3$, i.e.\ the local surface normal, whose $z$-component is $n_{i,z}$,
\begin{equation}
    \mathbf{g}_i =
    \bigl[\;
    \bar{z}_i,\;
    \lambda_2\!-\!\lambda_3,\;
    \lambda_1\!-\!\lambda_2,\;
    \lambda_3,\;
    |n_{i,z}|,\;
    \log(1+|\mathcal{N}_r(i)|)
    \;\bigr],
    \label{eq:geo}
\end{equation}
whose six channels are, in order, the voxel height $\bar{z}_i$ normalized to $[0,1]$ across the map, planarity, linearity and sphericity in the sense of the standard dimensionality features \cite{demantke2012dimensionality,weinmann2015semantic}, the verticality of the surface normal, and a density term counting the voxels $\mathcal{N}_r(i)$ within radius $r$, which compensates for the varying sampling density of aerial reconstructions \cite{hackel2016fast}. Channels are standardized per-channel using training-split statistics only. These features are label-independent: they describe the local geometry of the voxel, not its current label.

The learned embedding $\mathbf{z}_i$ is the penultimate activation of a sparse volumetric U-Net built on Minkowski convolutions \cite{choy20194d,graham2018submanifold}. The network is first pretrained for voxel classification on a large synthetic aerial corpus STPLS3D \cite{chen2022stpls3d} and subsequently fine-tuned on the OccuFly training scenes using geometric inputs only. Its classifier is then discarded, and the network is frozen during graph training. The resulting embedding captures multi-scale geometric context unavailable to Eq.~\eqref{eq:geo}, and because $\mathbf{z}_i$ is computed from geometry and occupancy only, it does not directly depend on the corrupted semantic label.

\noindent\textbf{Edge attributes.}
Each edge carries
\begin{equation}
    \mathbf{e}_{ij} =
    \bigl[\;
    \Delta_{ij}^{x},\;
    \Delta_{ij}^{y},\;
    \Delta_{ij}^{z},\;
    \tau_{ij},\;
    \lVert \boldsymbol{\Delta}_{ij} \rVert_2
    \;\bigr],
    \quad
    \boldsymbol{\Delta}_{ij} = \mathbf{c}_j - \mathbf{c}_i ,
    \label{eq:edgeattr}
\end{equation}
where $\tau_{ij} \in \{1,2,3\}$ encodes the number of spatial axes involved in the relative displacement, extending the usual face-, edge-, and corner-adjacency notion beyond immediate grid neighbors. Keeping the signed displacement separate from distance allows attention to distinguish directions as well as proximity \cite{wang2021egat}, which is useful in aerial scenes where vertical and horizontal relationships carry different geometric meanings.

\noindent\textbf{Semantic priors.}
Geometry alone does not capture whether a voxel is consistent with its surrounding semantic context. We therefore derive two priors from the annotated training maps: class co-occurrence captures which labels typically occur as neighbors, while class prototypes capture the geometry typically associated with each label. The fixed co-occurrence matrix $\mathbf{A} \in \mathbb{R}^{K \times K}$ is built from class adjacencies across the training edges. After symmetrization and row normalization, $A_{ab}$ gives the probability that a neighbor of class $a$ belongs to class $b$. For geometry, each class $c$ is represented by the mean $\boldsymbol{\mu}_c$ and standard deviation $\boldsymbol{\sigma}_c$ of its geometric descriptors $\mathbf{g}$ over all training voxels of that class \cite{li2020prototypical}. For a voxel currently labeled $\hat{y}_i$, we measure its geometric agreement with that label as
\begin{equation}
    \boldsymbol{\delta}_i =
    \frac{\mathbf{g}_i - \boldsymbol{\mu}_{\hat{y}_i}}
         {\boldsymbol{\sigma}_{\hat{y}_i} + \epsilon},
    \qquad
    a_i = \frac{1}{1 + \lVert \boldsymbol{\delta}_i \rVert_2} \in (0,1],
    \label{eq:agreement}
\end{equation}
where $\epsilon>0$ is a small constant for numerical stability. A low $a_i$ indicates that the voxel geometry is atypical for its current label, while $\boldsymbol{\delta}_i$ retains the descriptor channels along which the mismatch occurs. Each edge additionally receives
\begin{equation}
    \mathbf{s}_{ij} =
    \bigl[\;
    A_{\hat{y}_i \hat{y}_j},\;
    \mathbf{1}[\hat{y}_i = \hat{y}_j],\;
    |a_i - a_j|
    \;\bigr],
    \label{eq:semedge}
\end{equation}
encoding class compatibility from the co-occurrence prior, label agreement, and the difference in geometric agreement between its endpoints.

\noindent\textbf{Network input.} Nodes are finally described by
$\mathbf{x}^{\text{in}}_i =
[\,\mathbf{g}_i
\,\|\, \mathbf{z}_i
\,\|\, \mathrm{onehot}(\hat{y}_i)
\,\|\, a_i
\,\|\, \boldsymbol{\delta}_i\,]
\in \mathbb{R}^{29+K}$,
where
$\mathrm{onehot}(\hat{y}_i) \in \{0,1\}^{K}$
encodes the current label. This makes the label under correction an explicit input, so the model reasons about the \emph{current} labeling rather than predicting semantics from geometry alone. Edges are described either by $\mathbf{e}_{ij}$ alone or by
$[\,\mathbf{e}_{ij} \,\|\, \mathbf{s}_{ij}\,]$;
these define complementary geometric and semantic edge views over the shared graph topology, which are consumed by the two branches of \method.

\subsection{\method}
\label{sec:dualgat}

\begin{figure*}[t]
\centering
\includegraphics[width=\textwidth,trim=0 250 0 0,clip]{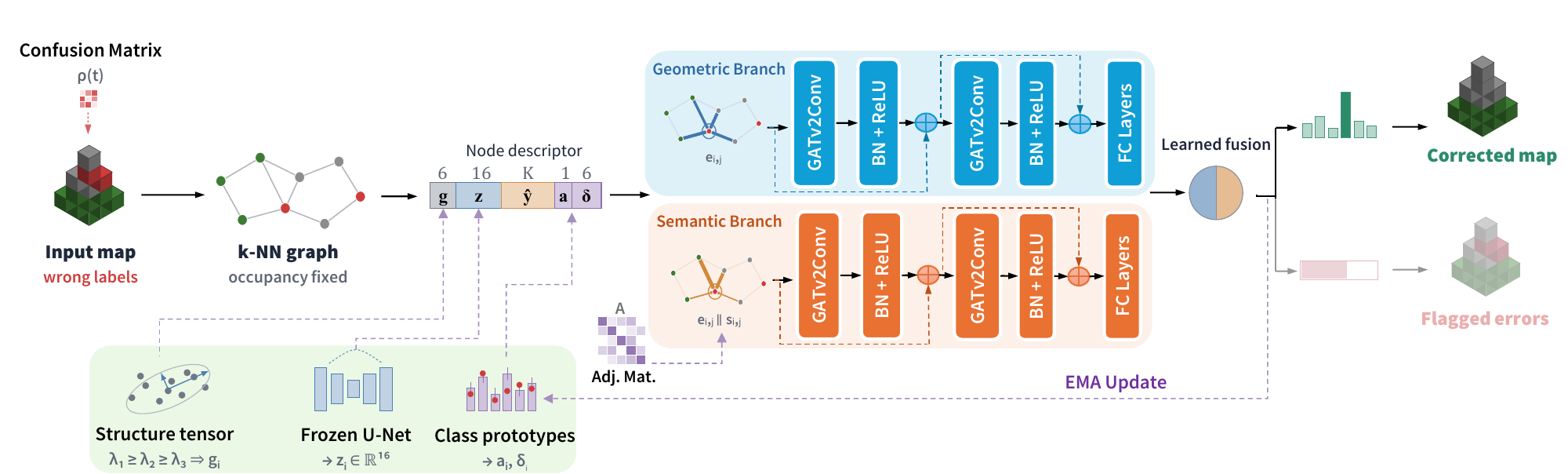}
\vspace{-0.75em}
\caption{\textbf{\method.} Occupied voxels form a fixed $k$-NN graph whose node descriptors combine local geometry $\mathbf{g}_i$, frozen volumetric embeddings $\mathbf{z}_i$, current labels $\hat{y}_i$, and class-conditioned geometric agreement features $a_i,\boldsymbol{\delta}_i$. The geometric branch attends to geometric edge attributes $\mathbf{e}_{ij}$, while the semantic branch additionally uses $\mathbf{s}_{ij}$ containing class compatibility from the co-occurrence prior $\mathbf{A}$ and label-consistency cues. Their outputs are fused by a voxel-wise gate for semantic relabeling. Dashed paths denote training-only corruption, prototype updates, and auxiliary error-detection supervision.}
\label{fig:arch}
\end{figure*}

The correction model, shown in \cref{fig:arch}, projects the voxel features into a shared representation, processes them through parallel geometric and semantic GATv2 branches, and fuses their outputs for correction and error detection.

\noindent\textbf{Input projection.}
The heterogeneous components of $\mathbf{x}^{\text{in}}_i$ are projected independently before being combined into a shared $d$-dimensional state,
\begin{equation}
    \mathbf{h}^{(0)}_i = \mathbf{W}\bigl[\,
    \mathbf{P}_g \mathbf{g}_i \,\|\,
    \mathbf{P}_z \mathbf{z}_i \,\|\,
    \mathbf{P}_y \mathrm{onehot}(\hat{y}_i) \,\|\,
    \mathbf{P}_a a_i \,\|\,
    \mathbf{P}_\delta \boldsymbol{\delta}_i \,\bigr].
    \label{eq:proj}
\end{equation}
Separate projections allow each feature group to be transformed before fusion.

\noindent\textbf{Dual branches.}
Both branches operate on the same graph but use different edge information. The geometric branch attends to $\mathbf{e}_{ij}$, whereas the semantic branch augments it with  $\mathbf{s}_{ij}$. Each branch stacks two residual GATv2 layers \cite{brody2021attentive}, with attention coefficients
\begin{equation}
    \alpha_{ij} = \operatorname*{softmax}_{j \in \mathcal{N}(i)}
    \Bigl( \mathbf{a}^{\!\top} \mathrm{LeakyReLU}
    \bigl( \mathbf{W}[\,\mathbf{h}_i \,\|\, \mathbf{h}_j \,\|\,
    \mathbf{u}_{ij}\,] \bigr) \Bigr),
    \label{eq:attn}
\end{equation}
where $\mathbf{u}_{ij}$ denotes the edge representation of the respective branch. Unlike standard GAT \cite{velivckovic2017graph}, GATv2 allows neighbor ranking to depend on the source node, which is useful for assessing local consistency. Each layer averages two attention heads and is followed by batch normalization, ReLU, dropout, and a residual connection.

\noindent\textbf{Gated fusion.}
The two branch outputs are combined using a voxel-wise gate. Each branch produces a gate score, which is normalized across the two branches,
\begin{equation}
    w^{b}_i = \sigma\bigl( \mathrm{MLP}_b(\mathbf{h}^{b}_i) \bigr),
    \quad
    \mathbf{h}^{\text{fused}}_i = \!\!\sum_{b \in \{\text{geo},\,\text{sem}\}}\!\!
    \frac{w^{b}_i}{w^{\text{geo}}_i + w^{\text{sem}}_i}\, \mathbf{h}^{b}_i .
    \label{eq:fusion}
\end{equation}
The relative contribution of geometric and semantic context can therefore vary between voxels within the same map.

\noindent\textbf{Heads and objective.}
The fused state feeds a correction head producing class logits over $\mathcal{C}$
and an auxiliary error-detection head estimating whether the input label is incorrect. Training minimizes
\begin{equation}
\mathcal{L}
=
\lambda_{\text{CE}}\mathcal{L}_{\text{CE}}
+
\lambda_{\text{det}}\mathcal{L}_{\text{det}},
\label{eq:loss}
\end{equation}
where $\mathcal{L}_{\text{CE}}$ is class-frequency-weighted cross-entropy for
label correction and $\mathcal{L}_{\text{det}}$ is binary cross-entropy on the
error mask $m_i=\mathbf{1}[\hat{y}_i\neq y_i]$, with a per-map positive-class
weight to account for the imbalance between correct and erroneous voxels.
Class prototypes are initialized from ground-truth statistics and updated during training by an exponential moving average using the predicted labels $\tilde{y}_i$. The resulting prototypes are fixed at inference.

\subsection{Confusion-Guided Noise Curriculum}
\label{sec:noise}

Because paired clean and corrupted maps are not naturally available, we synthesize training errors using a heuristic derived from voxel-level confusion patterns of the completed upstream maps. The resulting class-transition matrix $\mathbf{T}$ guides label substitutions
within spatially contiguous voxel patches, producing regional errors such as roof-to-wall or road-to-grass confusion rather than independent label flips \cite{patrini2017making,ye2022robust}. Most substitutions are sampled from $\mathbf{T}$, with a small fraction sampled from the scene class distribution to introduce additional variation. The corruption rate is annealed over training,
\begin{equation}
\rho(t)=\rho_{\max}
-(\rho_{\max}-\rho_{\min})\frac{t}{T_\mathrm{max}},
\label{eq:noise_schedule}
\end{equation}
where $t$ is the current training epoch, $T_{\mathrm{max}}$ is the max number of training epochs, $\rho_{\max}=0.50$, and $\rho_{\min}=0.05$. The model therefore encounters larger regional errors early and smaller residual errors later. The corrupted map is used as input, while the clean labels supervise correction and error detection. Patch construction and sampling details are provided in the supplementary material.


\section{Experiments}
\label{sec:experiments}

\begin{figure*}[b]
    \centering
    \setlength{\tabcolsep}{1pt} 
    \renewcommand{\arraystretch}{0.5} 
    
    \begin{tabular}{ccccccc}
        \footnotesize Input & 
        \footnotesize KNN & 
        \footnotesize CRF & 
        \footnotesize Geometry & 
        \footnotesize MinkUNet & 
        \footnotesize VoxelFix (Ours) & 
        \footnotesize GT \\[2pt] 
        
        \includegraphics[width=0.138\textwidth]{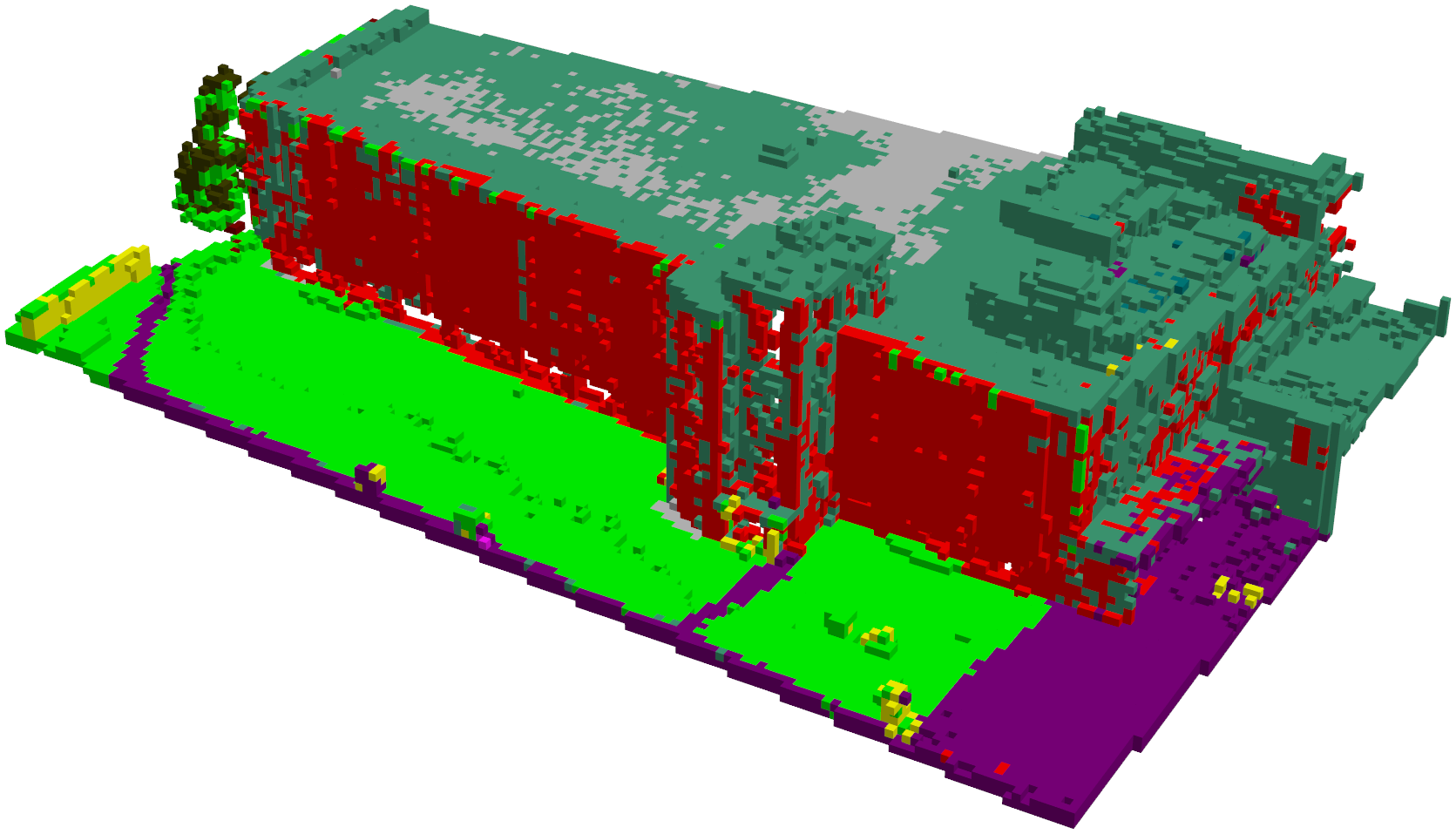} &
        \includegraphics[width=0.138\textwidth]{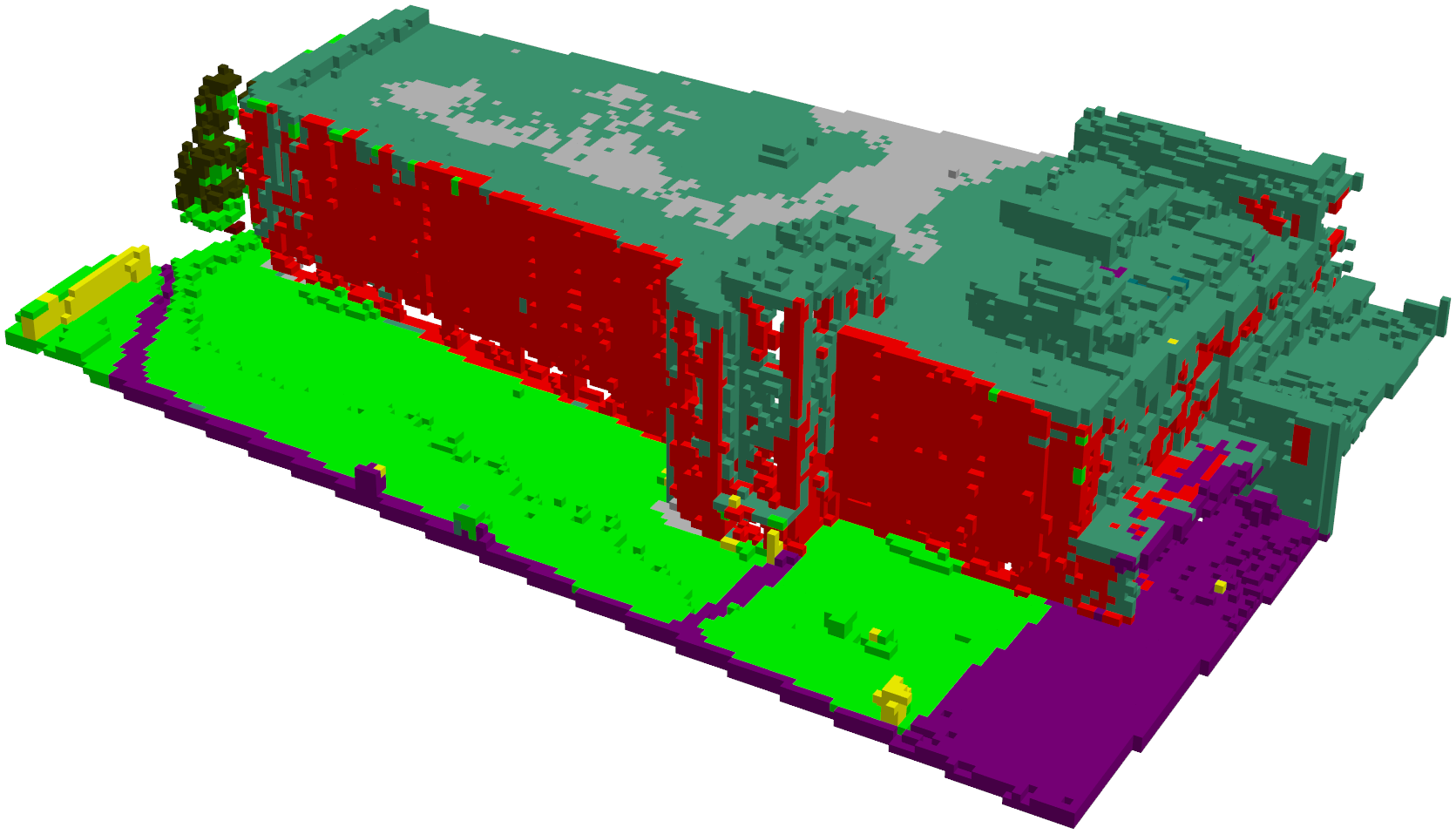} &
        \includegraphics[width=0.138\textwidth]{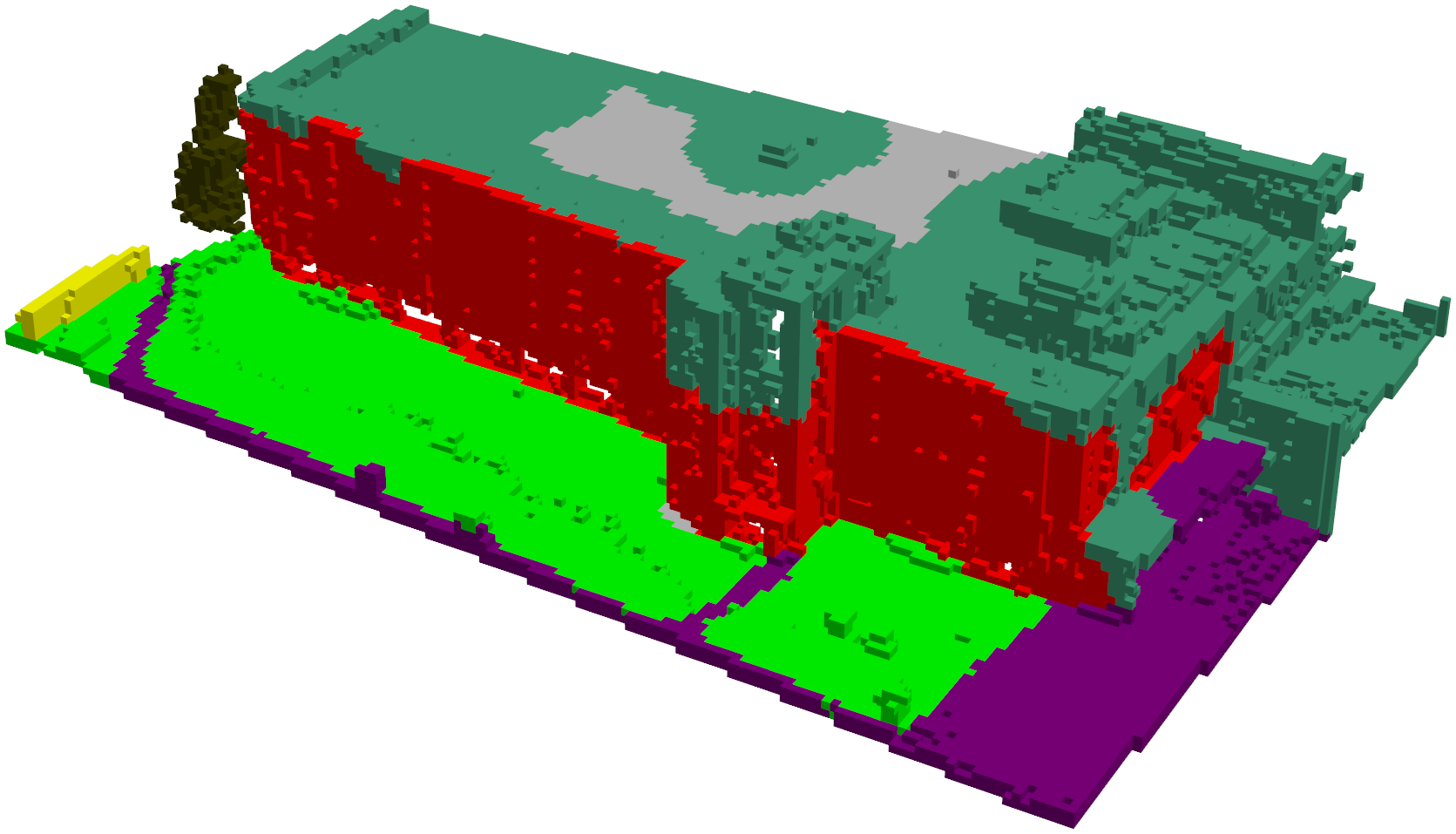} &
        \includegraphics[width=0.138\textwidth]{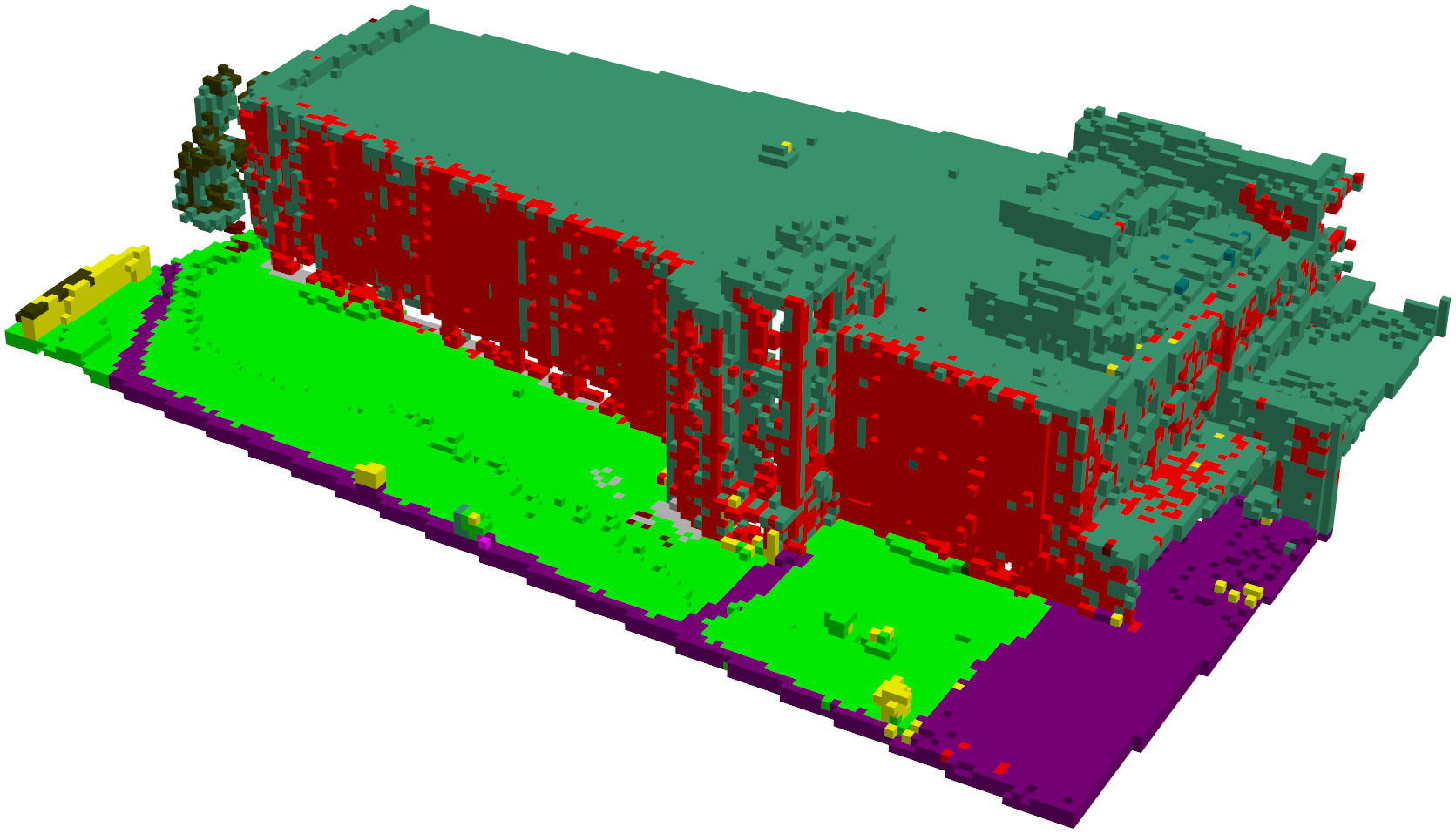} &
        \includegraphics[width=0.138\textwidth]{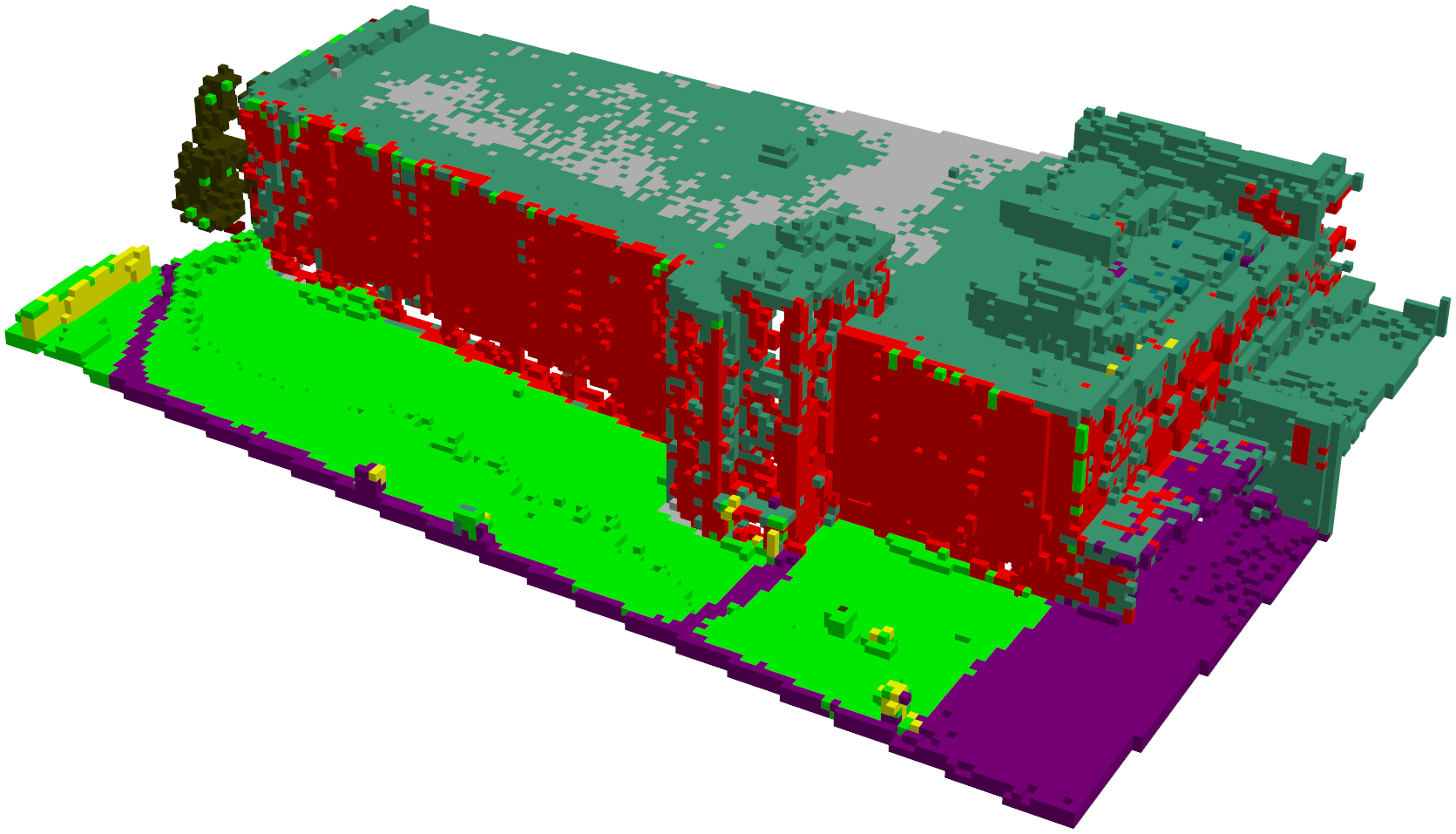} &
        \includegraphics[width=0.138\textwidth]{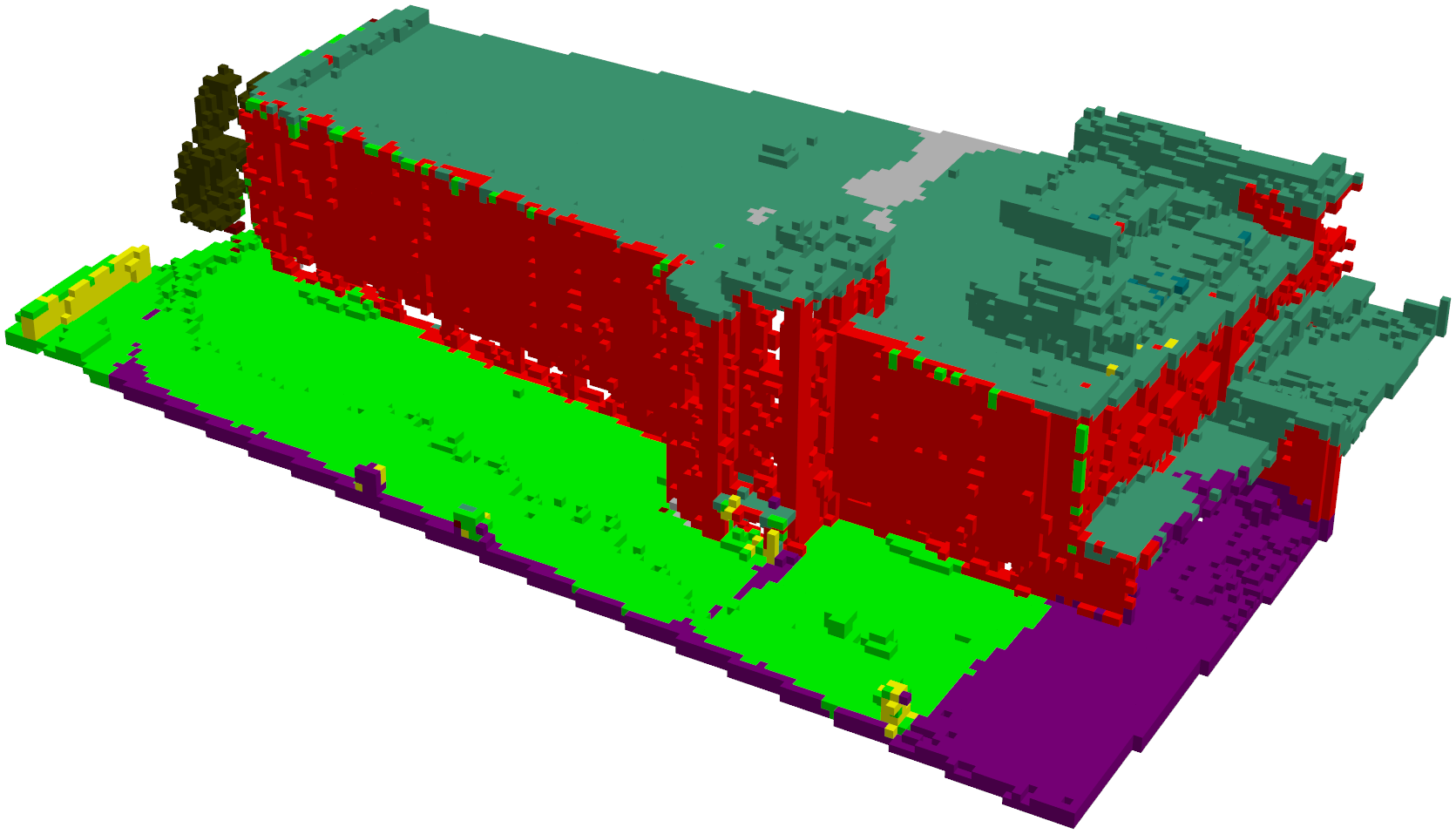} &
        \includegraphics[width=0.138\textwidth]{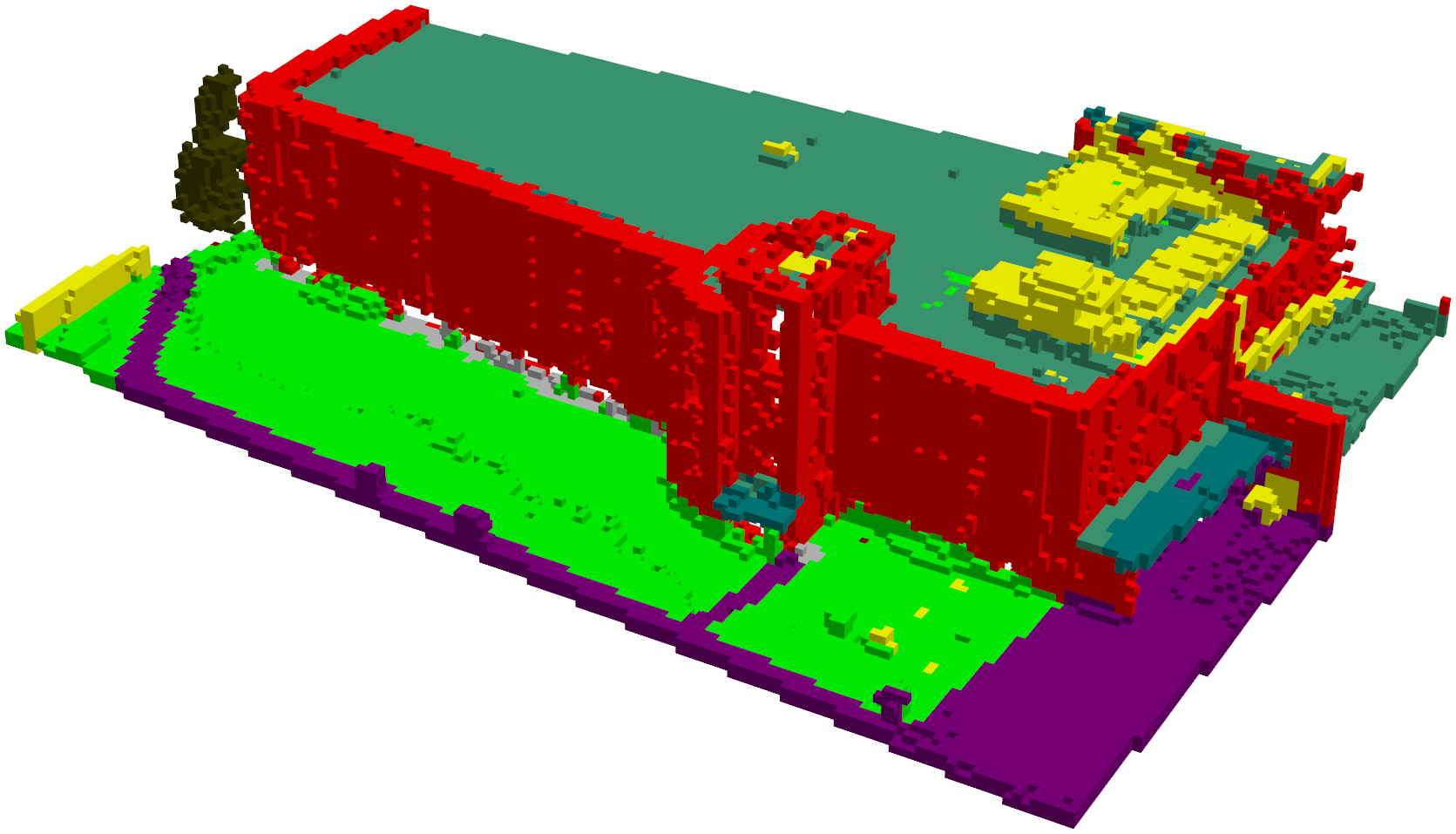} \\
        
        \includegraphics[width=0.138\textwidth]{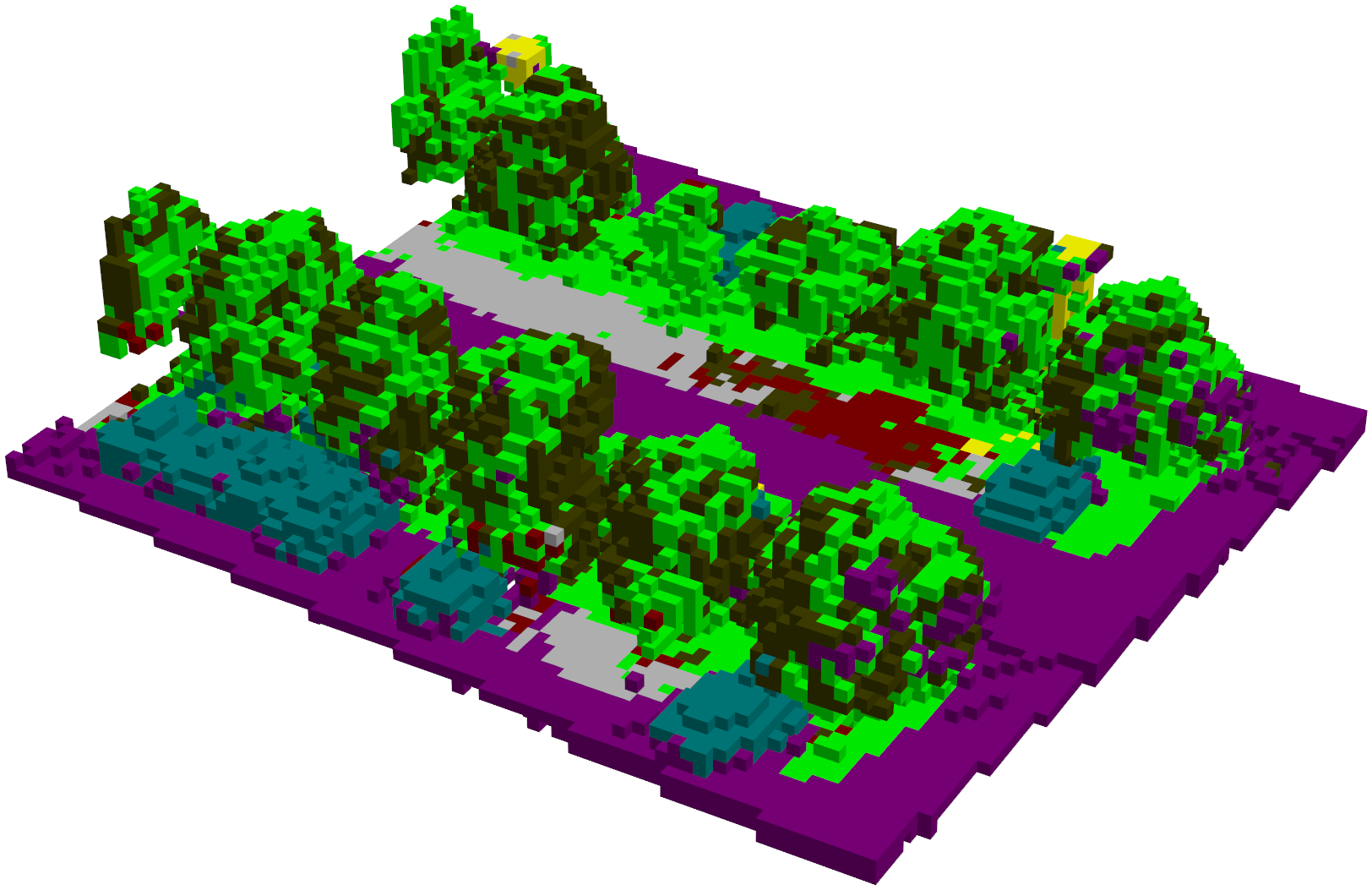} &
        \includegraphics[width=0.138\textwidth]{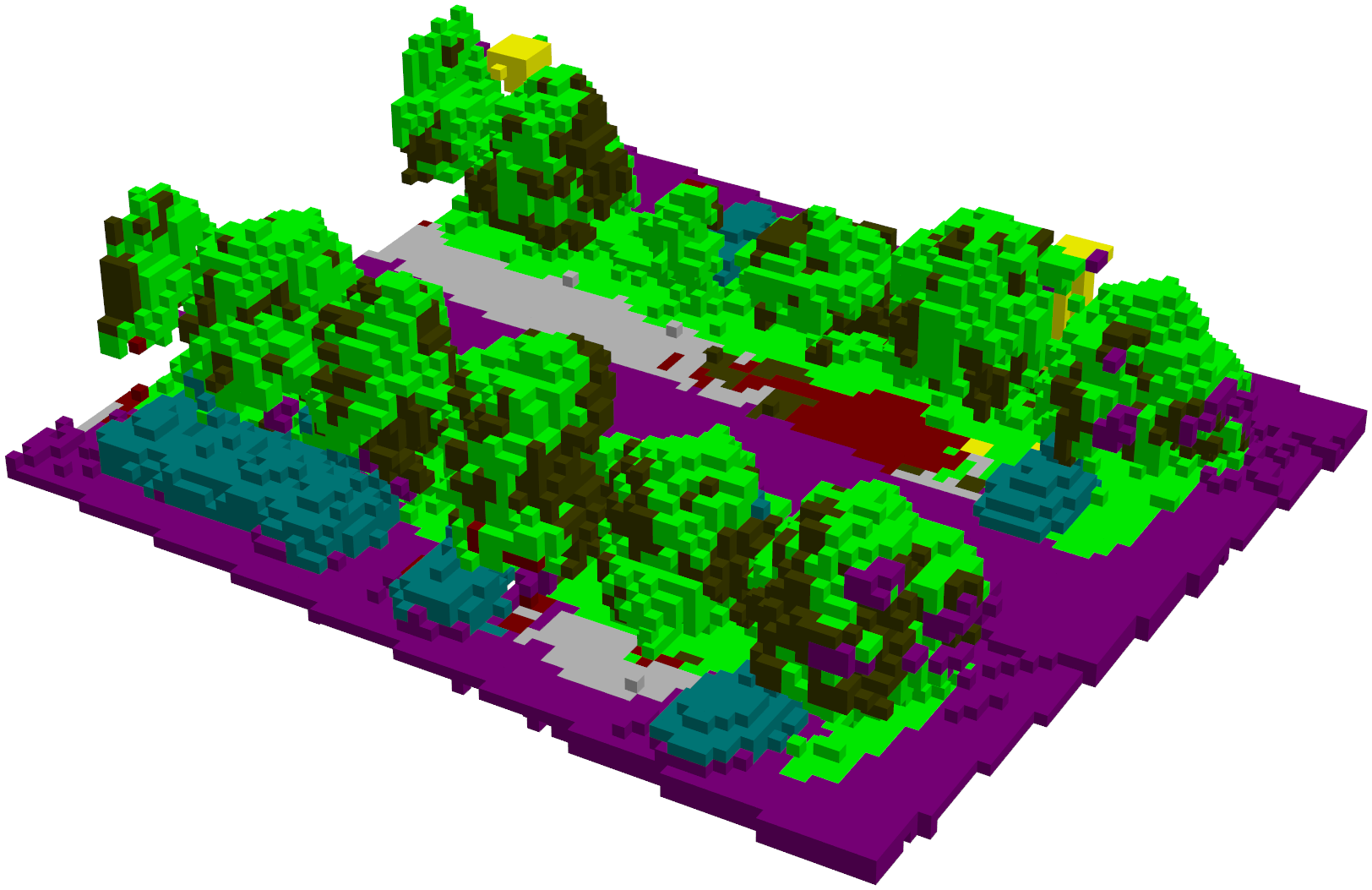} &
        \includegraphics[width=0.138\textwidth]{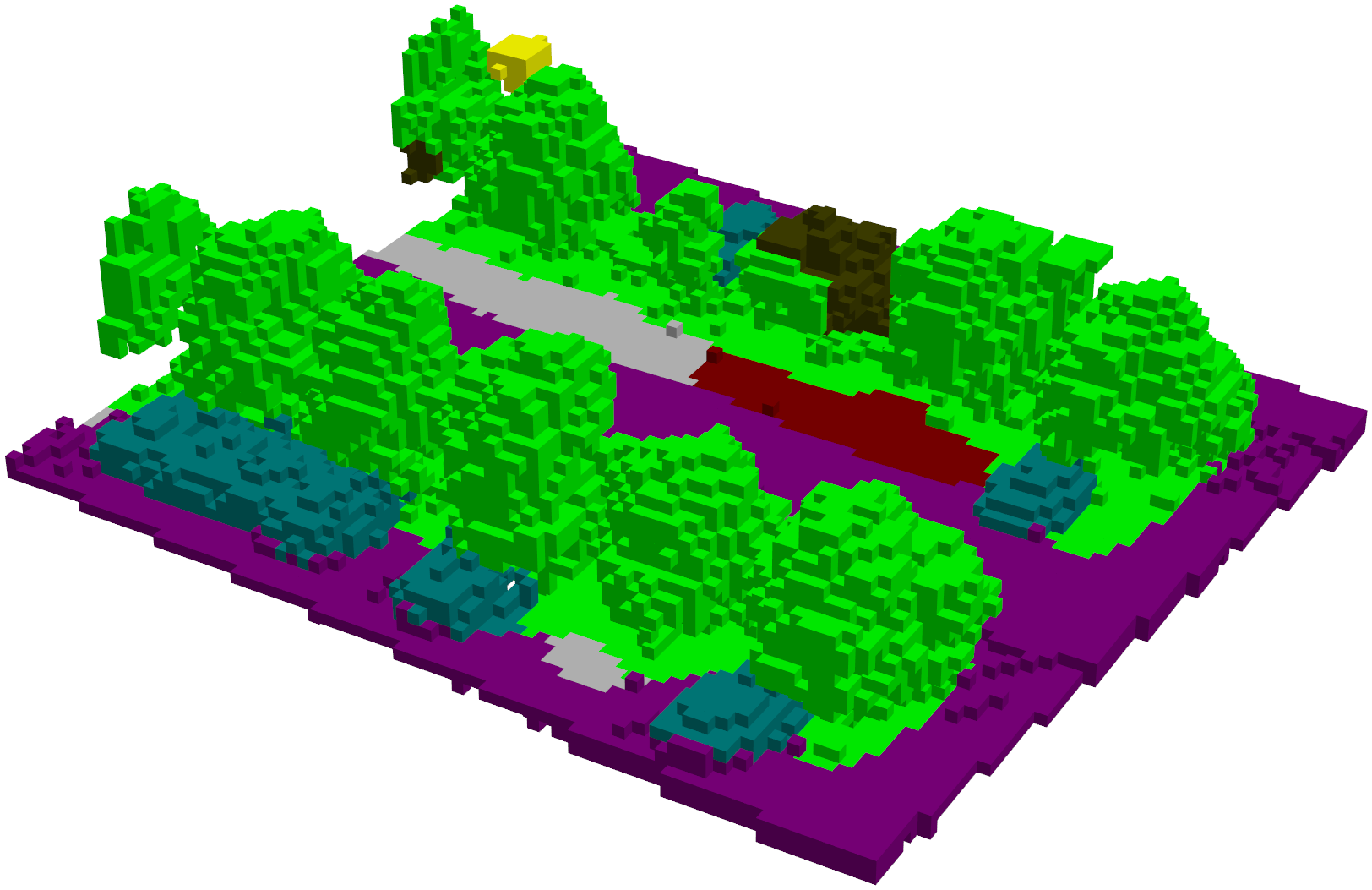} &
        \includegraphics[width=0.138\textwidth]{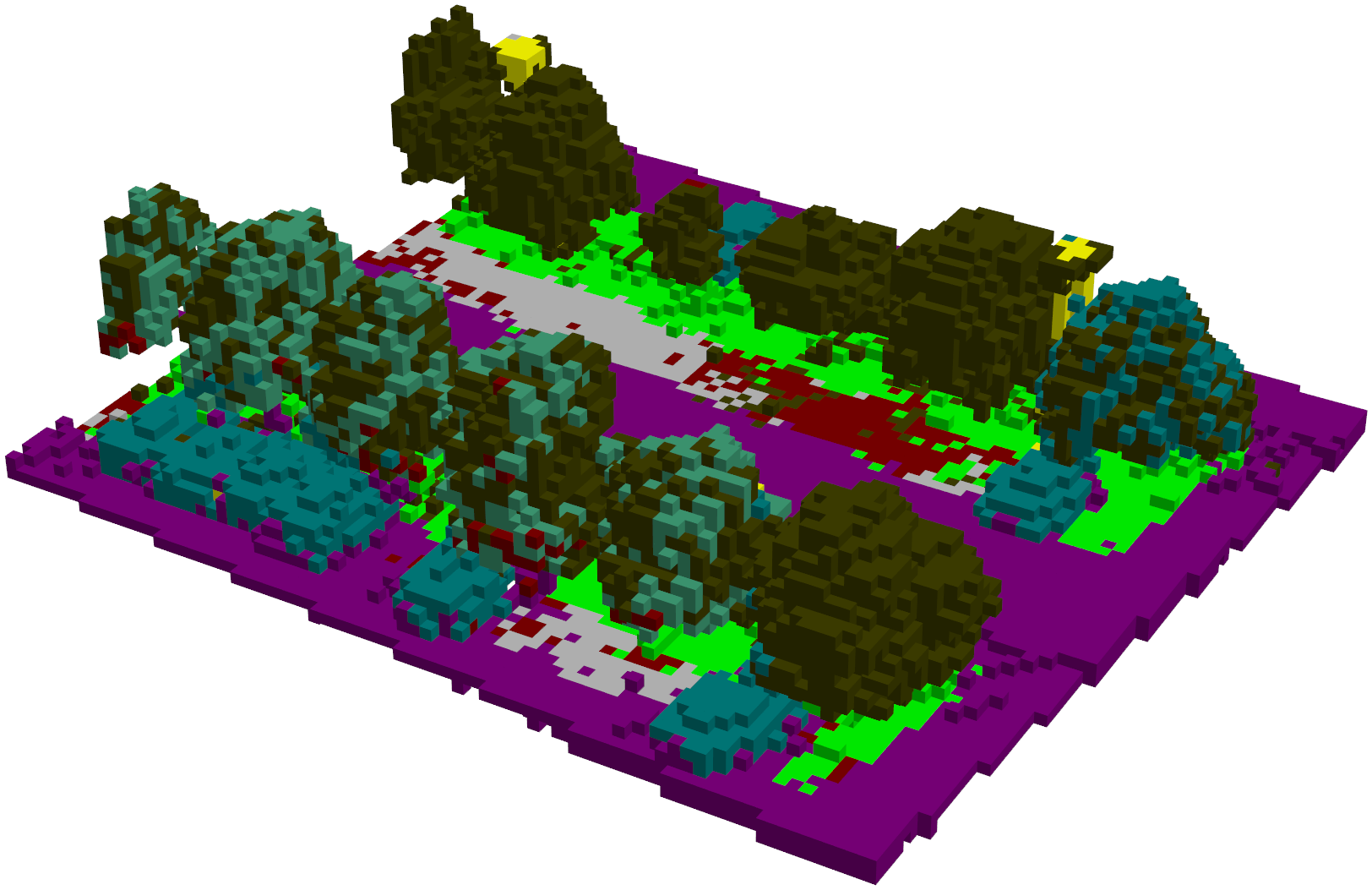} &
        \includegraphics[width=0.138\textwidth]{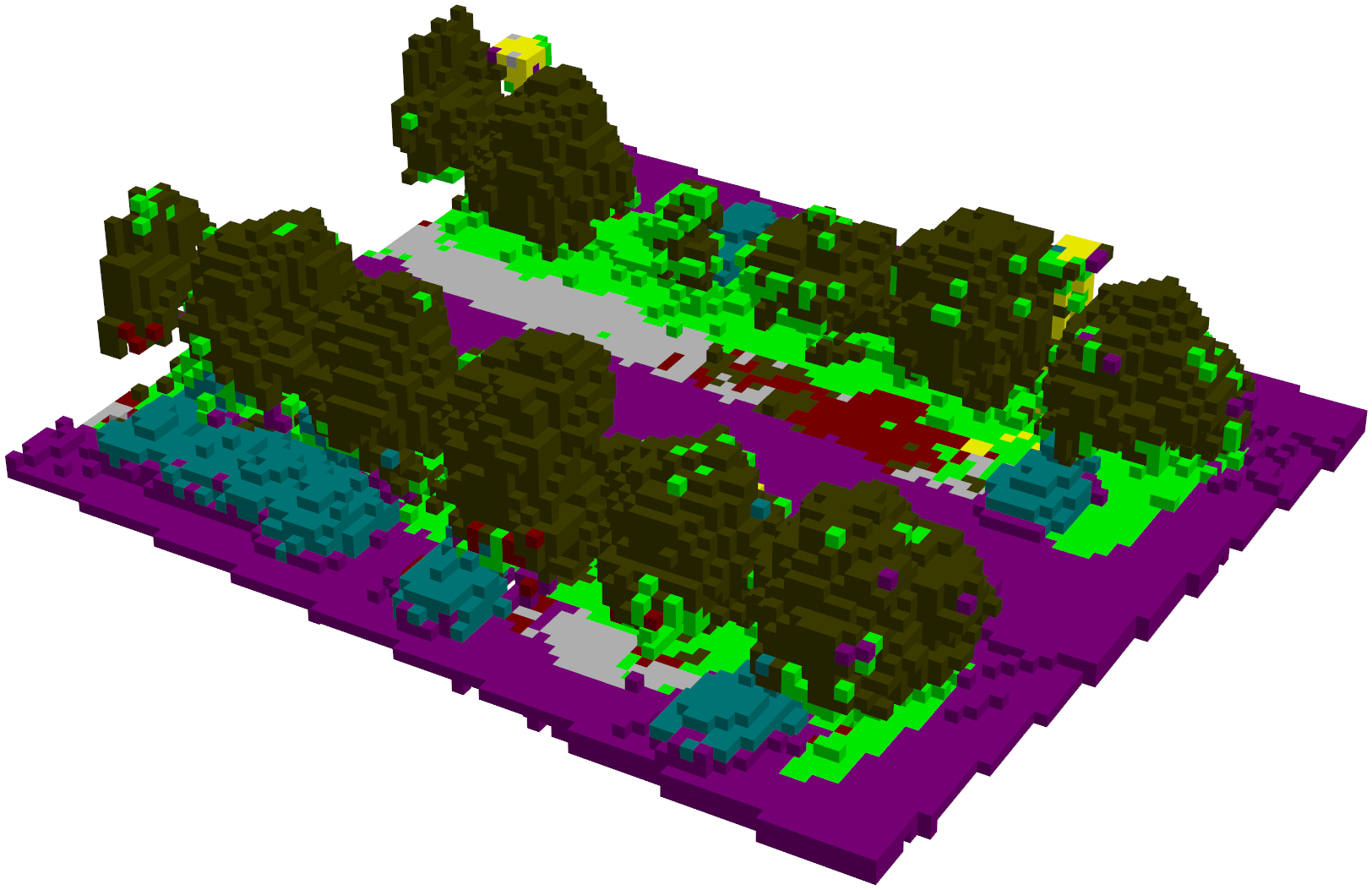} &
        \includegraphics[width=0.138\textwidth]{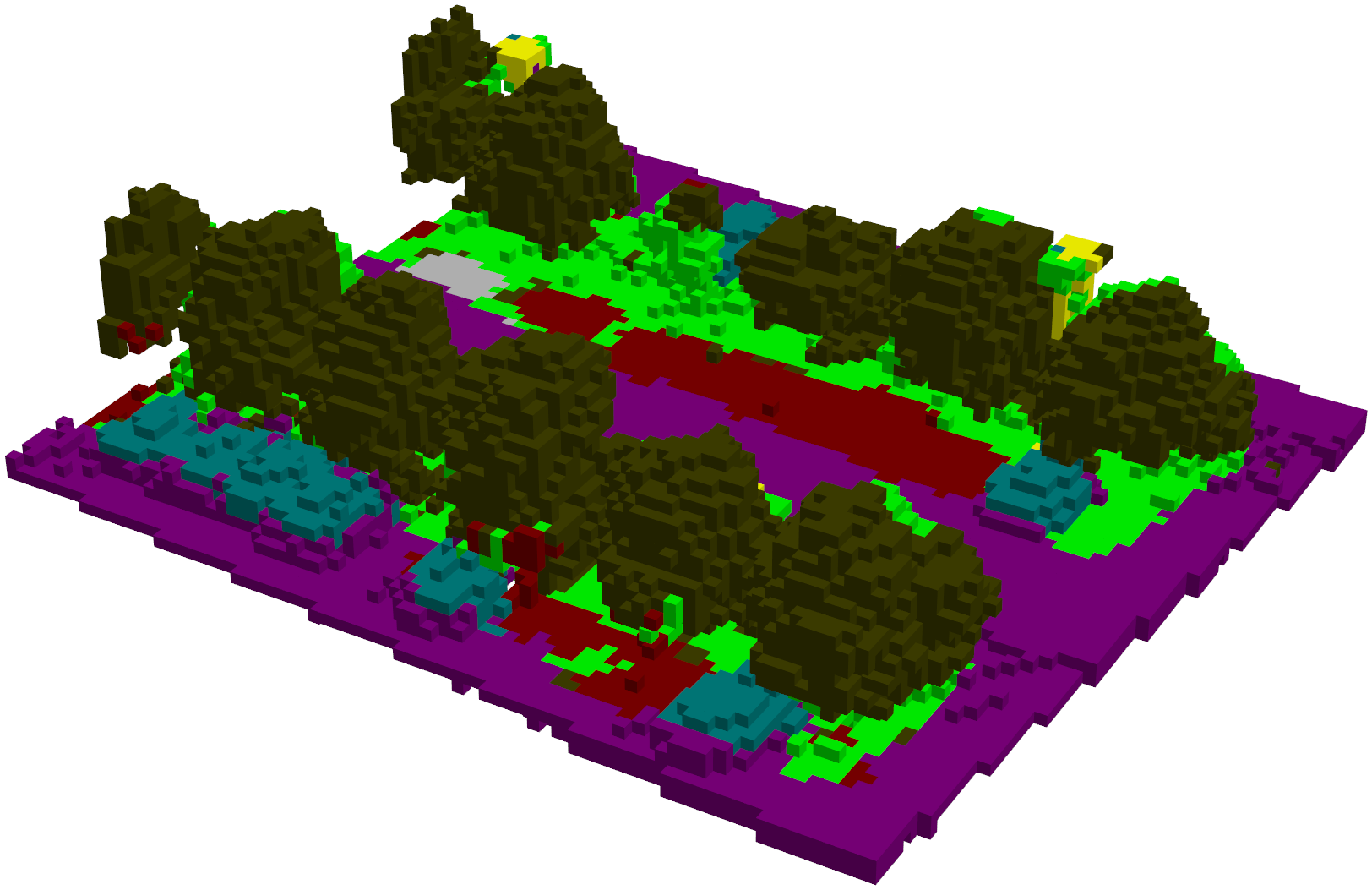} &
        \includegraphics[width=0.138\textwidth]{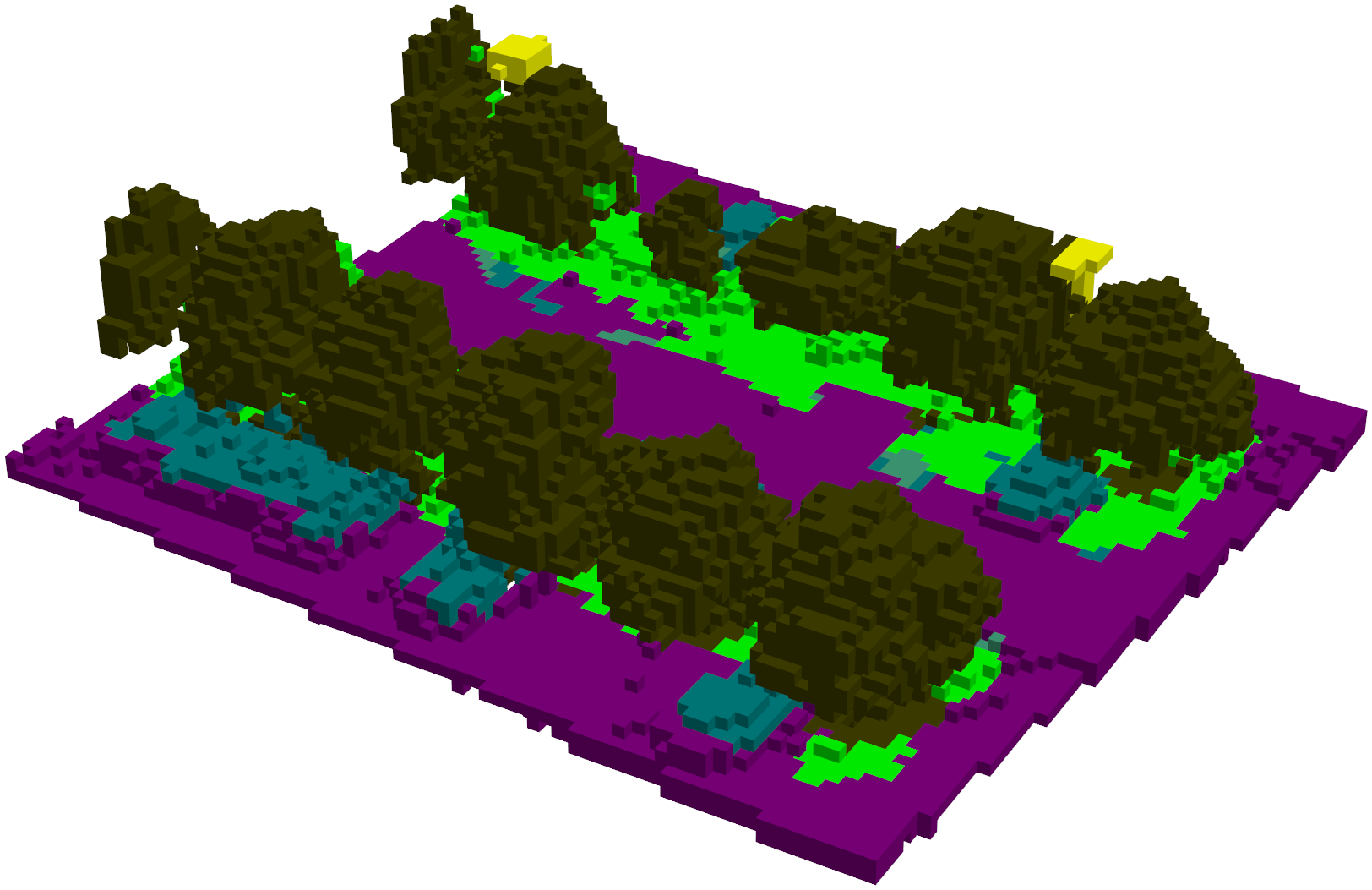} \\
        
        \includegraphics[width=0.138\textwidth]{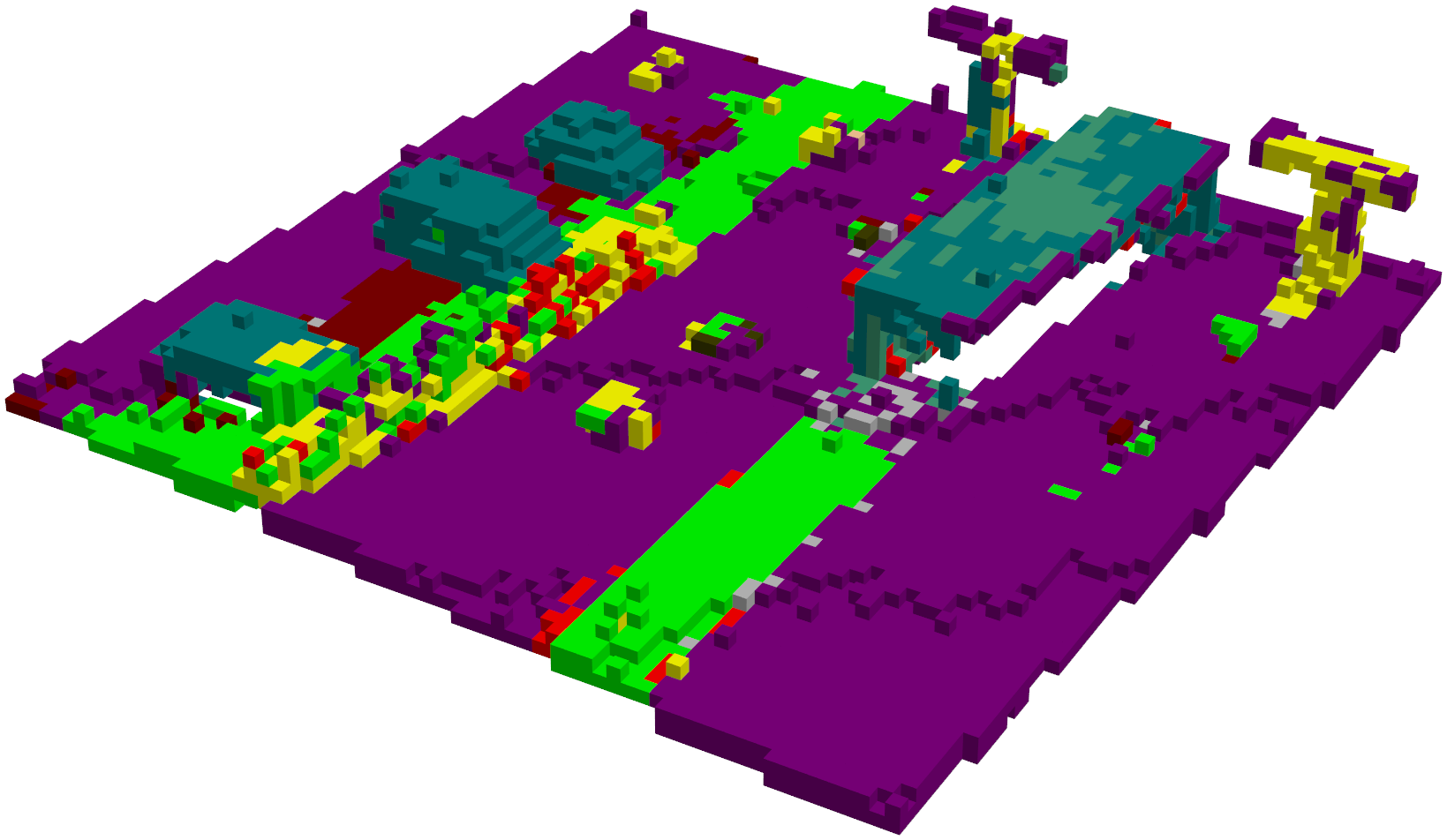} &
        \includegraphics[width=0.138\textwidth]{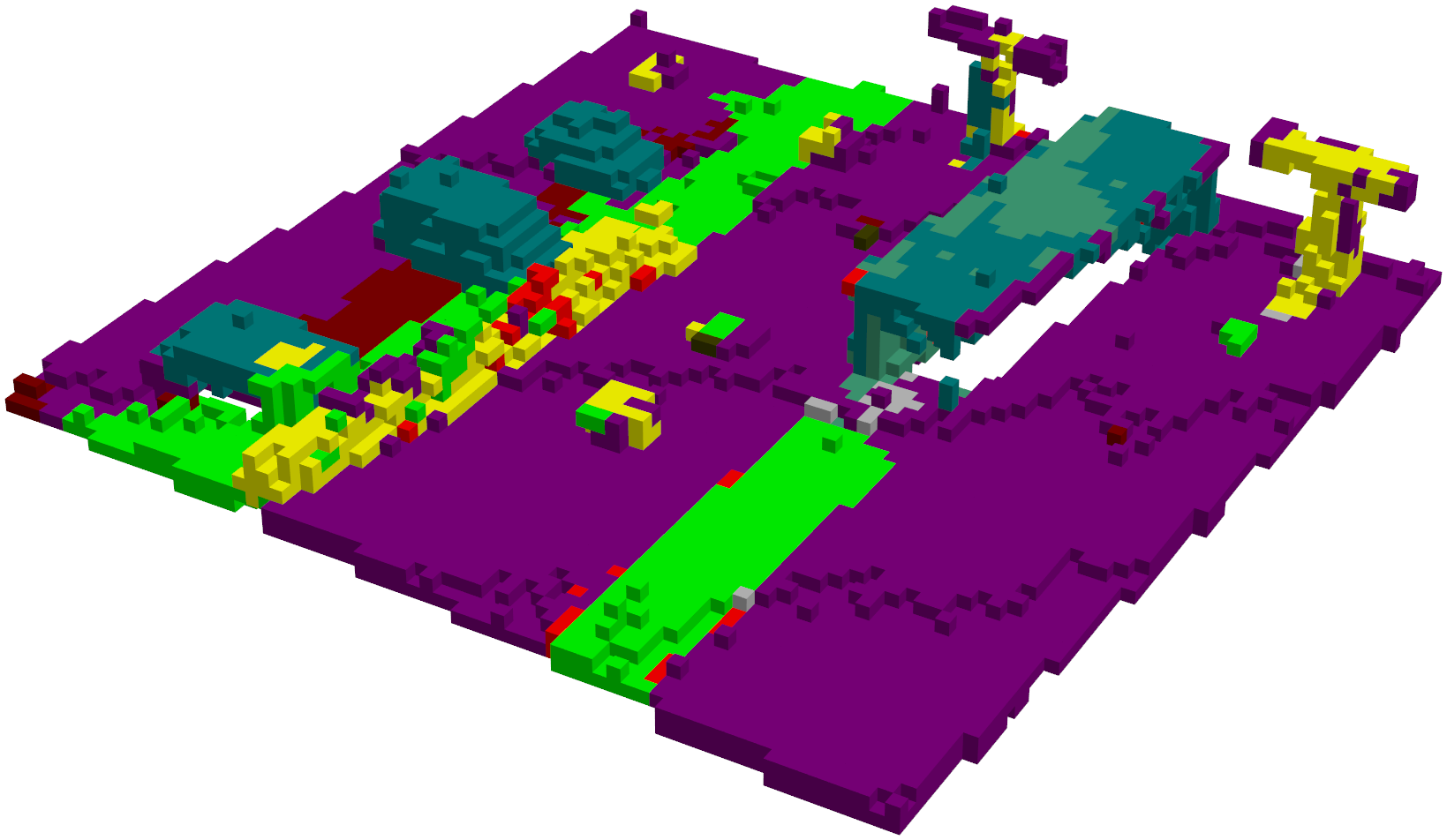} &
        \includegraphics[width=0.138\textwidth]{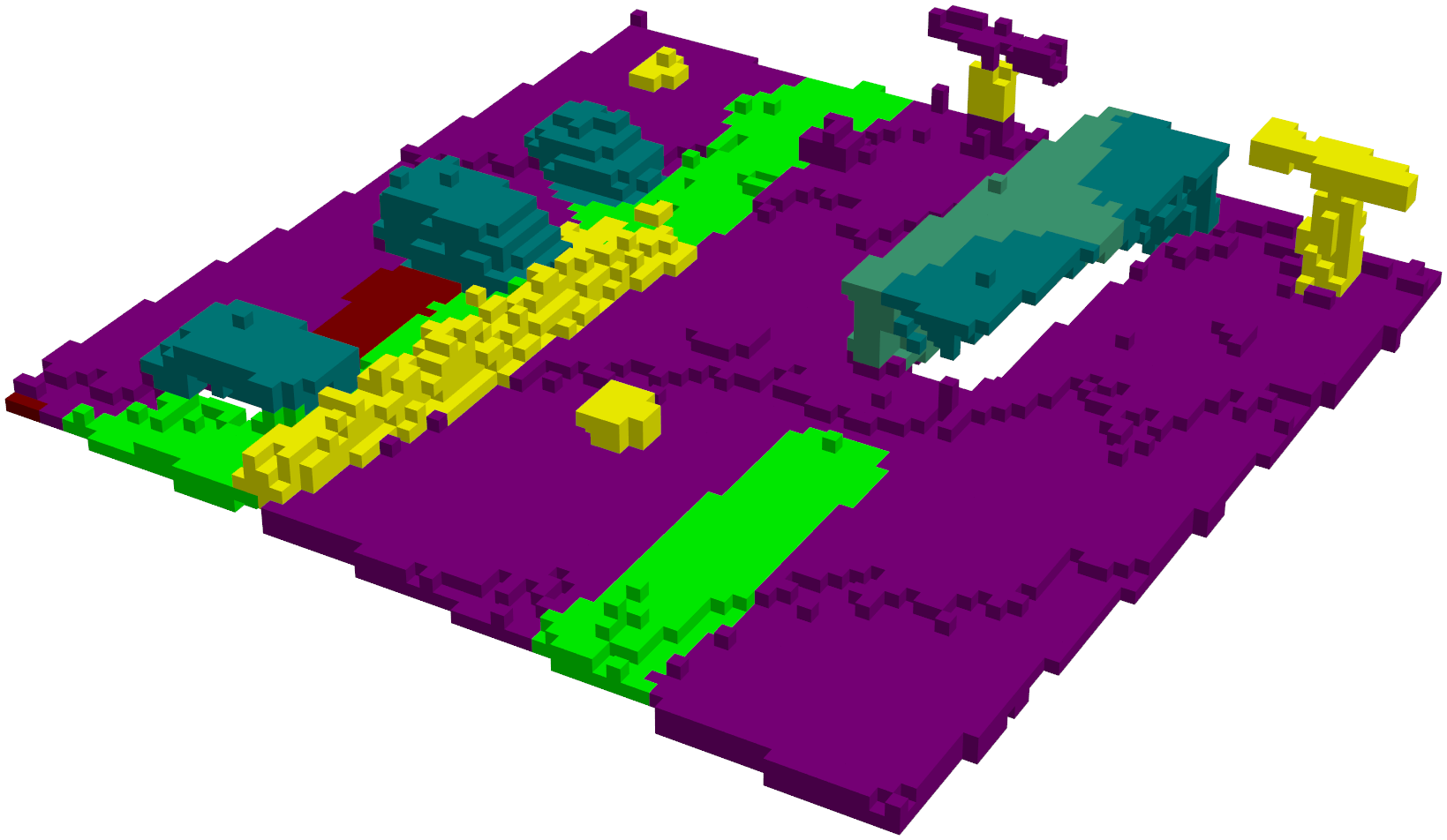} &
        \includegraphics[width=0.138\textwidth]{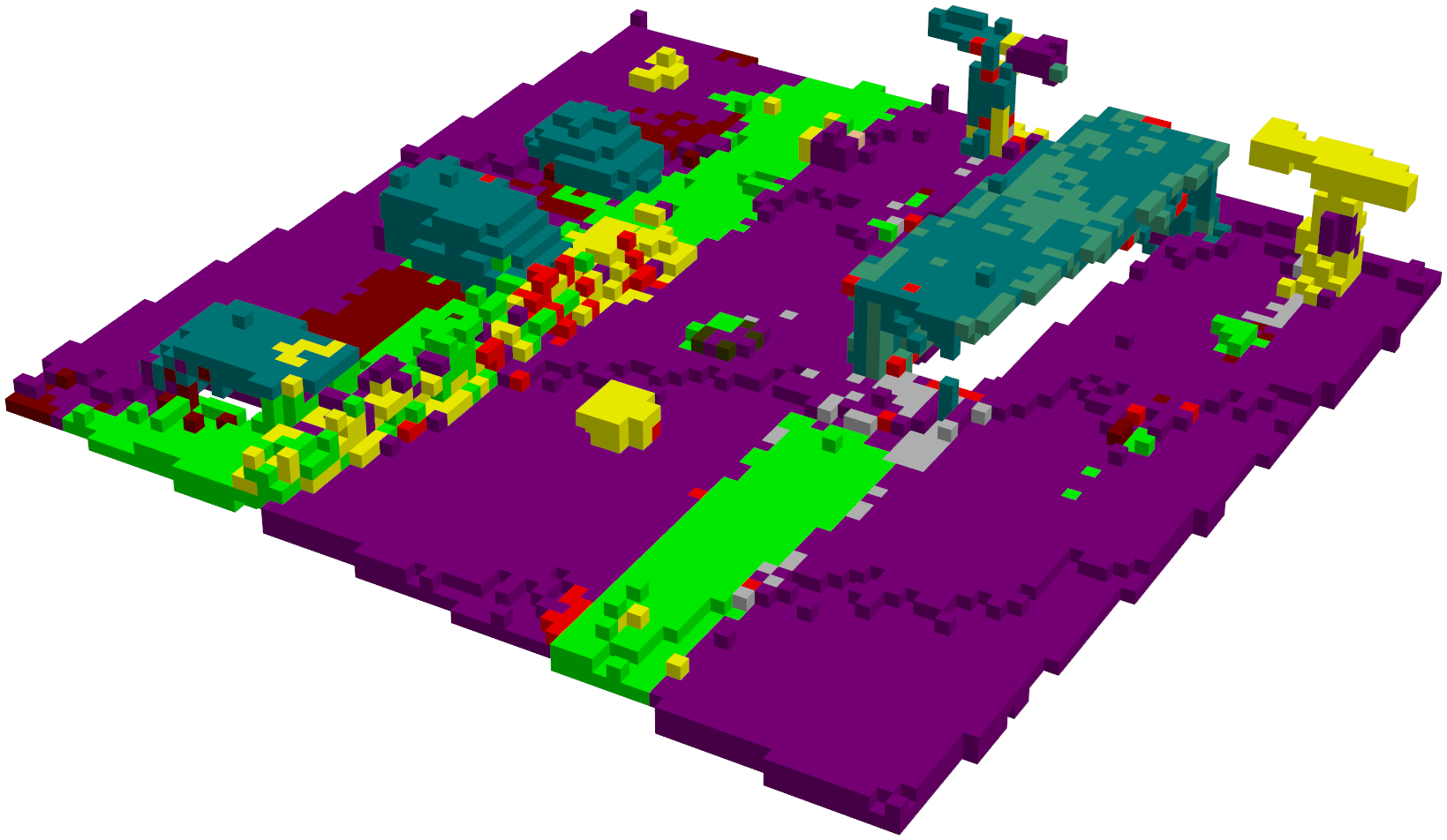} &
        \includegraphics[width=0.138\textwidth]{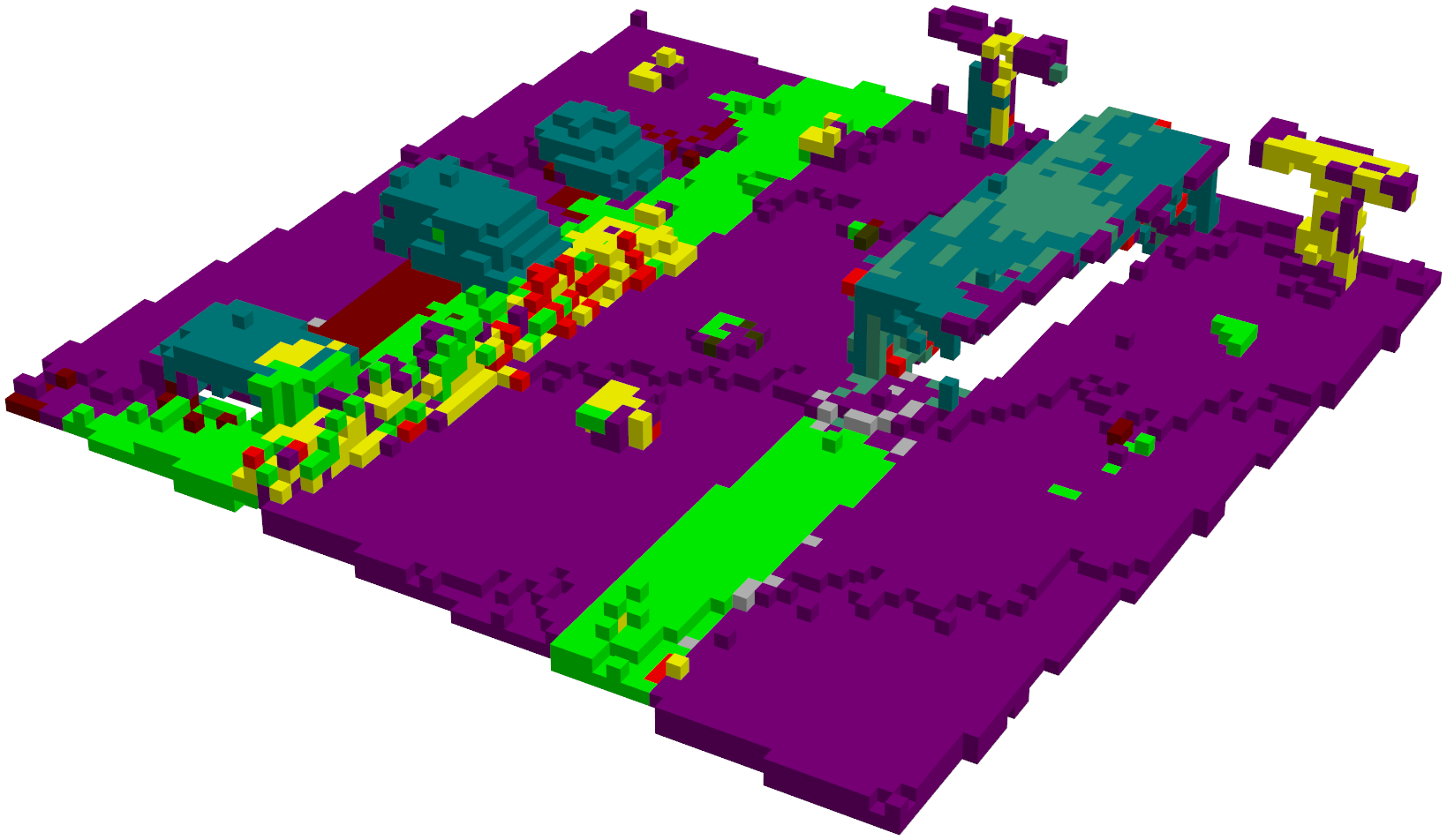} &
        \includegraphics[width=0.138\textwidth]{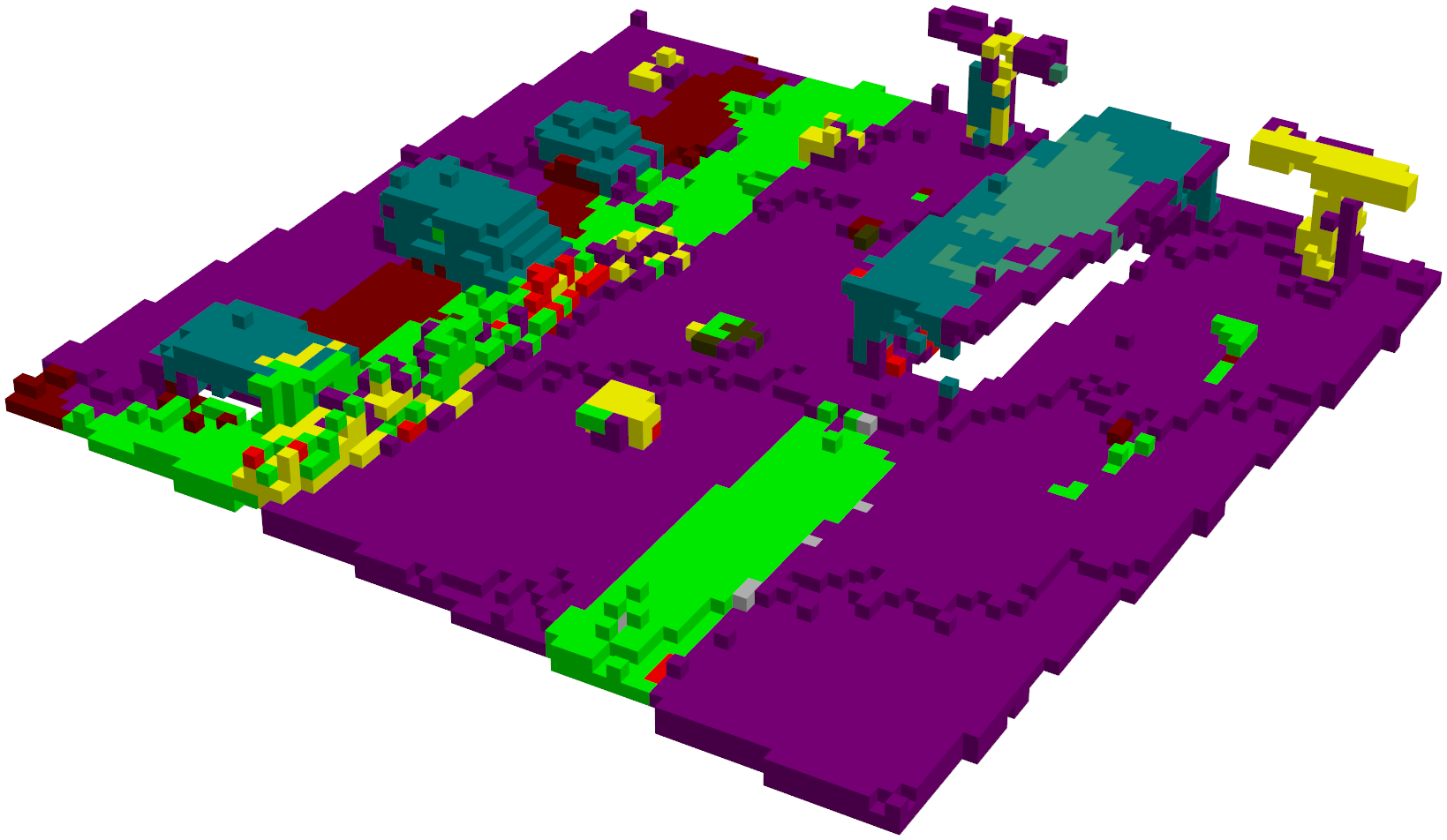} &
        \includegraphics[width=0.138\textwidth]{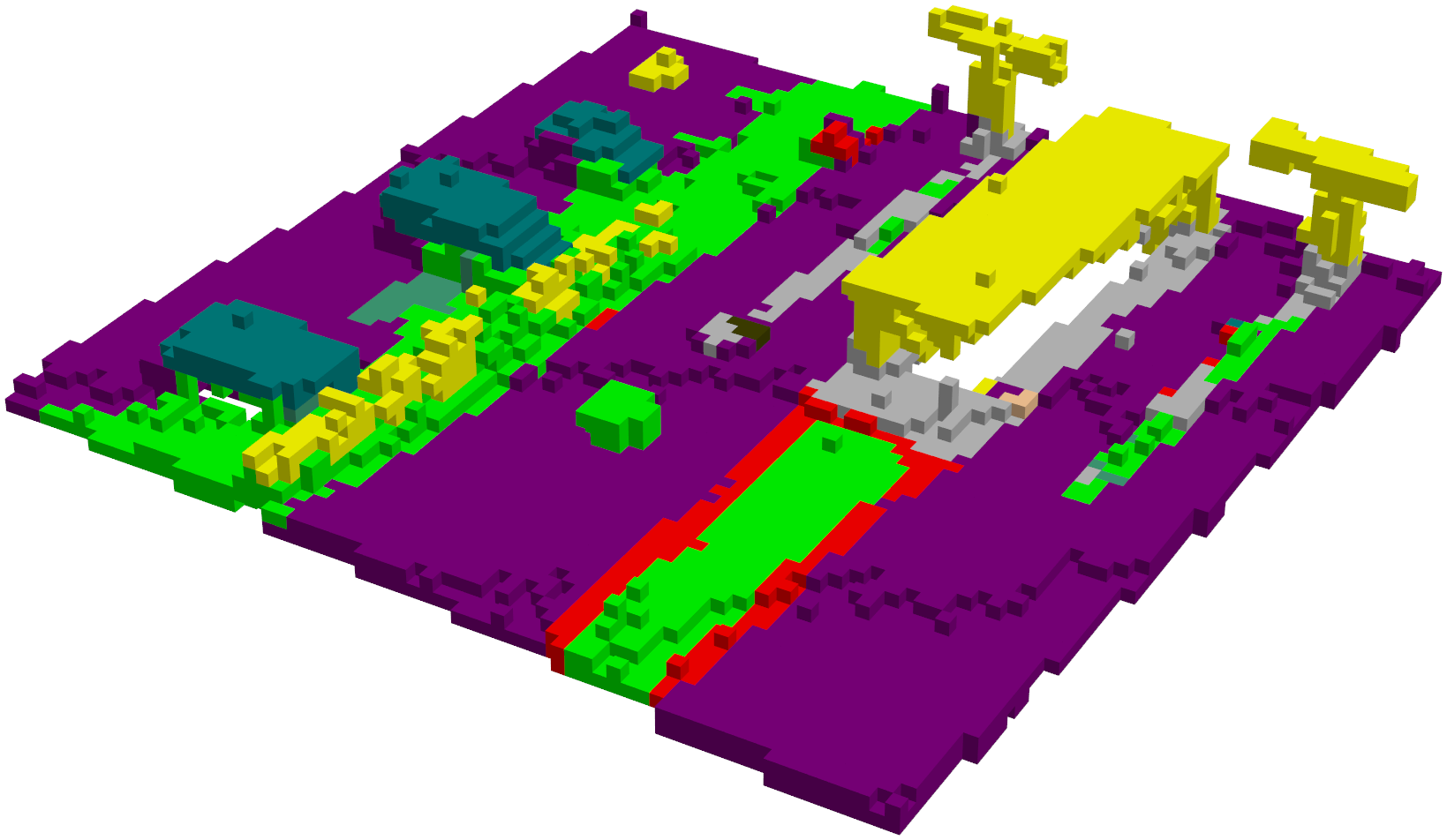} \\
    \end{tabular}
    
    \caption{\textbf{Qualitative map-level correction on the fixed OccuFly test scene.}
    From left to right: uncorrected Radix input, KNN, CRF, geometry heuristic, MinkUNet, \method, and ground truth. \method more consistently corrects spatially coherent semantic errors while preserving surrounding labels compared with local smoothing and geometry-only refinement.}
    \label{fig:qualitative_results}
\end{figure*}

\subsection{Experimental Setup}
\noindent\textbf{Datasets and upstream models.} We conduct our main experiments on nine OccuFly~\cite{gross2026occufly} scenes with scene-level ground-truth semantic voxel maps. To generate maps with different upstream error characteristics, we use four independently trained 2D segmentation models: SegFormer~\cite{xie2021segformer}, UPerNet~\cite{xiao2018unified} with Swin-v2-T~\cite{liu2022swin} and ConvNeXt-v2-T~\cite{woo2023convnext}, and Mask2Former~\cite{cheng2022masked} with Swin-T~\cite{liu2021swin}. The upstream models are trained on a unified 26-class aerial taxonomy using a combination of seven publicly available aerial datasets~\cite{icg_drone_dataset,okutswiss_dataset,chen2018large,Lee2024CARTCA,cai2025vdd,DBLP:journals/corr/abs-1810-10438,incenda_ai} and two proprietary datasets. The proprietary data are included to broaden coverage of the semantic classes represented in OccuFly, and neither contains scenes used in the in-distribution or OOD evaluation. Predictions are remapped to the OccuFly taxonomy and consolidated into 12 semantic classes for map correction by merging semantically similar categories. Full segmentation training details and class mappings are provided in the supplementary material.

\noindent\textbf{Semantic voxel map generation.} For each OccuFly scene, the remapped 2D semantic predictions are back-projected using the ground-truth depth maps and camera poses and aggregated into a sparse semantic voxel map using Radix~\cite{scherer2026socc}. For each occupied voxel, Radix maintains a class distribution that is updated by an exponential moving average of the observed semantic label frequencies, and the final voxel label is given by the maximum-probability class. After map construction, \method receives only these hard semantic labels with the fixed voxel geometry. The source images, depth maps, camera poses, and per-frame predictions are discarded. The CRF baseline additionally uses the class probabilities retained by Radix, giving it access to richer semantic information than \method. Because ground-truth depth and poses are used for map construction, our evaluation isolates semantic errors arising from upstream segmentation predictions and their aggregation, rather than errors in the reconstructed geometry.

\noindent\textbf{Baselines.} We compare \method against KNN label smoothing, CRF refinement, a geometry-based correction heuristic, and a sparse volumetric MinkUNet baseline. KNN replaces each voxel label based on its local neighborhood, while the CRF jointly considers spatial proximity and the per-voxel class probabilities retained by Radix. The geometry heuristic first partitions the map into ground and above-ground regions using Cloth Simulation Filtering~\cite{zhang2016easy}. It then applies KNN majority voting to the ground and DBSCAN~\cite{ester1996density} clustering to the above-ground objects. MinkUNet is trained for voxel-wise correction using the same training splits as \method. Full baseline parameters are provided in the supplementary material.

\noindent\textbf{Training.} The geometry encoder is prepared in two stages. We first pretrain the sparse volumetric U-Net for voxel classification on STPLS3D, then fine-tune it to the OccuFly training scenes using ground-truth voxel labels and geometry-only inputs. The classifier is then discarded and the encoder frozen; its penultimate activations provide the 16-dimensional geometry embeddings $\mathbf{z}_i$. \method is then trained separately on synthetically corrupted training maps using the objective in Eq.~\eqref{eq:loss} and the noise curriculum in \cref{sec:noise}. Validation is performed on the held-out validation scene, and the checkpoint with the best validation performance is retained. Optimizer settings, loss weights, architectural hyperparameter tuning, augmentation, and training schedules are provided in the supplementary material.

\noindent\textbf{Cross-scene protocol.}
We follow a 7/1/1 cross-scene protocol. One OccuFly scene is fixed as the in-distribution test scene. Each of the remaining eight scenes is used once for validation, with the other seven forming the corresponding training split, resulting in eight independently trained models. The fixed test scene is not used for training, validation, checkpoint selection, or estimation of model-dependent statistics. All training-derived quantities, including feature-normalization statistics, semantic priors, prototype initialization, and confusion statistics, are computed exclusively from the seven training scenes of each split. Learned-model results are evaluated on the same fixed test scene and reported as mean $\pm$ standard deviation across the eight trained models. Deterministic baselines are evaluated once on the same test map.

\noindent\textbf{Metrics.}
We report mIoU and voxel accuracy as primary metrics. To quantify refinement behavior, we additionally report the error correction rate
$\mathrm{ECR}=
\frac{|\{i:\hat y_i\neq y_i,\ \tilde y_i=y_i\}|}
     {|\{i:\hat y_i\neq y_i\}|}$,
the fraction of initially incorrect voxels corrected, and the damage rate
$\mathrm{DR}=
\frac{|\{i:\hat y_i=y_i,\ \tilde y_i\neq y_i\}|}
     {|\{i:\hat y_i=y_i\}|}$,
the fraction of initially correct voxels made incorrect.

\subsection{Map-Level Correction} 

\begingroup
\setlength{\tabcolsep}{4pt}
\begin{table}[t!]
\centering
\caption{\textbf{Map-level semantic correction on the fixed OccuFly test scene using SegFormer-MiT-B3 upstream predictions.} Learned models report mean $\pm$ std
over eight independently trained cross-validation splits. $\Delta$mIoU is
measured relative to the uncorrected Radix input map.}
\renewcommand{\arraystretch}{0.9}
\resizebox{\linewidth}{!}{
\begin{tabular}{@{}l|ccccc@{}}
\toprule
Method & mIoU$\uparrow$ & $\Delta$mIoU$\uparrow$
       & Acc.$\uparrow$ & ECR$\uparrow$ & DR$\downarrow$ \\
\midrule
Radix (input)
& 25.80
& --
& 60.46
& --
& -- \\

KNN
& 26.14
& +0.34
& 60.89
& 5.12
& 2.64 \\

CRF
& 26.19
& +0.39
& 61.19
& 10.65
& 5.75 \\

Geometry Heuristic
& 28.60
& +2.80
& 66.27
& 24.51
& 6.41 \\

MinkUNet
& 28.52$\pm$1.17
& +2.72$\pm$1.17
& 65.83$\pm$2.34
& 16.16$\pm$7.09
& \textbf{1.69$\pm$0.82} \\

\midrule
\method (Ours)
& \textbf{30.66$\pm$0.55}
& \textbf{+4.86$\pm$0.55}
& \textbf{68.34$\pm$0.62}
& \textbf{25.36$\pm$1.67}
& 3.55$\pm$0.58 \\

\bottomrule
\end{tabular}
}
\label{tab:map_correction}
\end{table}
\endgroup
\begin{table*}[t!]
    \centering
    \scriptsize
    \setlength{\tabcolsep}{2.5pt}
    \renewcommand{\arraystretch}{0.9}
    \caption{\textbf{Per-class IoU (\%) and overall performance on the fixed OccuFly test scene using SegFormer-MiT-B3 upstream predictions.} Only classes present in the test scene are shown. Best and second-best results are shown in \textbf{bold} and \underline{underlined}, respectively.}
    \resizebox{\textwidth}{!}{%
    \begin{tabular}{l|>{\centering\arraybackslash}c>{\centering\arraybackslash}c>{\centering\arraybackslash}c>{\centering\arraybackslash}c>{\centering\arraybackslash}c>{\centering\arraybackslash}c>{\centering\arraybackslash}c>{\centering\arraybackslash}c>{\centering\arraybackslash}c>{\centering\arraybackslash}c>{\centering\arraybackslash}c|>{\centering\arraybackslash}c>{\centering\arraybackslash}c}
        \toprule
        \textbf{Method}
        & \textbf{Road}
        & \textbf{Tree}
        & \textbf{Vehicle}
        & \textbf{Grass}
        & \textbf{Wall}
        & \textbf{Roof}
        & \textbf{Obstacle}
        & \textbf{Gravel}
        & \textbf{Person}
        & \textbf{Dirt}
        & \textbf{Bicycle}
        & \cellcolor{gray!20}\textbf{mIoU}
        & \textbf{Acc.} \\

        & \textcolor[RGB]{128,0,128}{\rule{1.5ex}{1.5ex}}
        & \textcolor[RGB]{64,64,0}{\rule{1.5ex}{1.5ex}}
        & \textcolor[RGB]{0,128,128}{\rule{1.5ex}{1.5ex}}
        & \textcolor[RGB]{0,255,0}{\rule{1.5ex}{1.5ex}}
        & \textcolor[RGB]{255,0,0}{\rule{1.5ex}{1.5ex}}
        & \textcolor[RGB]{64,160,120}{\rule{1.5ex}{1.5ex}}
        & \textcolor[RGB]{255,255,0}{\rule{1.5ex}{1.5ex}}
        & \textcolor[RGB]{192,192,192}{\rule{1.5ex}{1.5ex}}
        & \textcolor[RGB]{255,16,255}{\rule{1.5ex}{1.5ex}}
        & \textcolor[RGB]{128,0,0}{\rule{1.5ex}{1.5ex}}
        & \textcolor[RGB]{255,204,153}{\rule{1.5ex}{1.5ex}}
        & & \\

        \midrule
        Radix (input)
        & 68.53
        & 42.85
        & 42.41
        & 40.25
        & 18.18
        & 52.97
        & 9.37
        & 6.83
        & 0.00
        & 2.43
        & 0.00
        & \cellcolor{gray!20}25.80
        & 60.46 \\

        KNN
        & 69.50
        & 42.49
        & 43.75
        & 40.75
        & 18.16
        & 53.38
        & 10.28
        & 6.91
        & 0.00
        & 2.36
        & 0.00
        & \cellcolor{gray!20}26.14
        & 60.89 \\

        CRF
        & \textbf{71.24}
        & 41.39
        & \underline{44.71}
        & 40.68
        & 17.14
        & 53.53
        & \underline{11.31}
        & 6.71
        & 0.00
        & 1.42
        & 0.00
        & \cellcolor{gray!20}26.19
        & 61.19 \\

        Geometry Heuristic
        & \underline{70.73}
        & \textbf{66.17}
        & 31.89
        & \textbf{49.07}
        & 17.21
        & 52.56
        & \textbf{12.02}
        & \textbf{12.27}
        & 0.00
        & \underline{2.71}
        & 0.00
        & \cellcolor{gray!20}\underline{28.60}
        & \underline{66.27} \\

        MinkUNet
        & 70.48$\pm$0.18
        & \underline{61.02$\pm$8.73}
        & \textbf{45.14$\pm$0.51}
        & \underline{46.16$\pm$2.97}
        & \underline{19.05$\pm$1.07}
        & \underline{53.65$\pm$0.60}
        & 8.93$\pm$0.23
        & 6.84$\pm$0.07
        & 0.00
        & 2.48$\pm$0.10
        & 0.00
        & \cellcolor{gray!20}28.52$\pm$1.17
        & 65.83$\pm$2.34 \\

        \midrule
        VoxelFix (Ours)
        & 69.79$\pm$0.70
        & 59.48$\pm$0.73
        & 44.66$\pm$1.47
        & 45.62$\pm$0.57
        & \textbf{29.49$\pm$2.92}
        & \textbf{67.50$\pm$2.44}
        & 9.39$\pm$0.45
        & \underline{8.29$\pm$1.20}
        & 0.00
        & \textbf{3.07$\pm$0.26}
        & 0.00
        & \cellcolor{gray!20}{\textbf{30.66$\pm$0.55}}
        & \textbf{68.34$\pm$0.62} \\

        \bottomrule
    \end{tabular}%
    }
    \label{tab:perclass_iou}
\end{table*}

We first evaluate whether semantic errors can be corrected directly from a completed voxel map, without access to the source images, depth maps, camera poses, or reconstruction process. We use SegFormer-MiT-B3 for the detailed comparison; upstream-model sensitivity is evaluated separately in \cref{sec:upstream_quality}. \cref{tab:map_correction} compares \method with the uncorrected Radix map and all baselines. In addition to mIoU and voxel accuracy, we report the change in mIoU relative to the input map ($\Delta$mIoU), together with ECR and DR to distinguish successful corrections from changes that corrupt initially correct labels.

\noindent\textbf{Overall correction performance.}
All correction methods improve the uncorrected Radix map, although their behavior differs substantially. KNN and CRF provide only modest gains, whereas the geometry-based heuristic improves the mIoU by +2.80 points. \method achieves the highest mIoU at 30.66$\pm$0.55 mIoU, a gain of $+4.86\pm0.55$ points over the Radix input and +2.14 points over MinkUNet. It corrects 25.36$\pm$1.67\% of initially erroneous voxels while corrupting only 3.55$\pm$0.58\% of initially correct ones. Compared with the geometry heuristic, \method achieves a slightly higher ECR (25.36\% vs.\ 24.51\%) while substantially reducing the DR (3.55\% vs.\ 6.41\%). Overall, \method provides stronger map-level correction than local smoothing, the geometry heuristic, and learned volumetric refinement.

\noindent\textbf{Qualitative behavior.}
The examples in \cref{fig:qualitative_results} illustrate where these differences arise. Local smoothing methods primarily reinforce the dominant labels in a neighborhood and therefore struggle when an erroneous region is spatially coherent. Geometry-based correction can resolve errors that admit a clear structural interpretation, such as tree vs. grass and roof vs. wall, but is limited when geometry alone does not determine the correct class. \method more consistently recovers mislabeled roof regions and separates trees from surrounding grass, while preserving correctly labeled regions. In more ambiguous areas, such as the mixed road-and-ground-obstacle region, errors persist. This indicates that correction remains limited when the map provides insufficient evidence for the correct class.

\noindent\textbf{Class-wise behavior.}
\cref{tab:perclass_iou} further decomposes the overall improvement by semantic class. Relative to the input map, \method improves or preserves the IoU of every evaluated class, rather than increasing overall mIoU through gains in only a small subset of categories. The largest improvements occur for tree, roof, and wall, while smaller gains are observed for road, vehicle, grass, and gravel. Although competing methods achieve higher IoU for individual classes, their improvements are less consistent across the class set. Person and bicycle remain at zero IoU for all methods. These results suggest that \method provides broad map-level refinement, with particularly strong gains for classes that exhibit informative geometric and semantic context.

\subsection{Effect of Upstream Quality}
\label{sec:upstream_quality}

We next examine whether correction performance depends on the upstream
segmentation model. We construct the same fixed test scene using predictions
from four independently trained segmentation models and apply \method
to each resulting map. As shown in \cref{tab:upstream_quality}, \method
consistently improves the completed map across all four upstream models,
yielding gains of $+4.23$ to $+5.00$ mIoU percentage points despite differences in the
initial map quality and error characteristics. The correction gain does not
follow the initial mIoU monotonically, suggesting that performance depends not
only on the amount of upstream error, but also on its semantic and spatial
structure.

\begingroup
\setlength{\tabcolsep}{4pt}
\begin{table}[t!]
\centering
\caption{\textbf{Effect of the upstream segmentation model on map-level correction
on the fixed OccuFly test scene.} Corrected results show mean $\pm$ std over eight cross-validation splits.}
\renewcommand{\arraystretch}{0.9}
\resizebox{\linewidth}{!}{
\begin{tabular}{@{}l|ccc@{}}
\toprule
Upstream model & Input mIoU & Corrected mIoU$\uparrow$
               & $\Delta$mIoU$\uparrow$ \\
\midrule
SegFormer-MiT-B3
    & 25.80 & 30.66$\pm$0.55 & +4.86$\pm$0.55 \\
UPerNet-SwinV2-T
    & 22.64 & 27.25$\pm$0.74 & +4.61$\pm$0.74 \\
UPerNet-ConvNeXtV2-T
    & 24.88 & 29.11$\pm$0.47 & +4.23$\pm$0.47 \\
Mask2Former-Swin-T
    & 23.70 & 28.69$\pm$0.41 & +5.00$\pm$0.41 \\
\bottomrule
\end{tabular}
}
\label{tab:upstream_quality}
\end{table}
\endgroup

\subsection{OOD Evaluation}
\label{sec:ood}

We further evaluate whether the learned correction transfers beyond the OccuFly environments used for training. We independently reconstruct an additional aerial scene using the same semantic mapping pipeline and generate its ground-truth voxel map following the OccuFly procedure. The scene is not used for training, validation, checkpoint selection, or estimation of semantic priors. For this correction-focused OOD evaluation, we use Mask2Former, whose completed map has lower initial quality (30.95 mIoU), leaving greater headroom for post-hoc correction than SegFormer (32.66 mIoU). \cref{tab:ood} compares the resulting map against the same map-level correction baselines used in the in-distribution evaluation. \method improves the unseen map from 30.95 to 32.66 mIoU, outperforming KNN by 0.68 points. The qualitative examples in \cref{fig:ood} show that \method can correct coherent errors on previously unseen scene geometry that persist after local or geometry-only refinement. However, ambiguous regions such as the solar panels on the roof remain unresolved, indicating that OOD correction is limited when the input map provides insufficient evidence for the correct class.

\begingroup
\setlength{\tabcolsep}{4pt}
\begin{table}[t!]
\centering
\caption{\textbf{OOD map-level correction on an independently reconstructed scene
using Mask2Former upstream predictions.} The scene is not used during training
or model selection.}
\renewcommand{\arraystretch}{0.8}
\resizebox{0.85\linewidth}{!}{
\begin{tabular}{@{}l|cccc@{}}
\toprule
Method & mIoU$\uparrow$ & $\Delta$mIoU$\uparrow$
       & ECR$\uparrow$ & DR$\downarrow$ \\
\midrule
Radix (input)
& 30.95 & -- & -- & -- \\

KNN
& 31.98 & +1.04 & 11.95 & 1.65 \\

CRF
& 31.95 & +1.00 & \textbf{25.88} & 3.72 \\

Geometry Heuristic
& 31.10 & +0.16 & 7.01 & 1.87 \\

MinkUNet
& 31.58
& +0.63
& 6.41
& \textbf{0.72} \\

\midrule
\method (Ours)
& \textbf{32.66}
& \textbf{+1.71}
& 15.65
& 1.03 \\

\bottomrule
\end{tabular}
}
\label{tab:ood}
\end{table}
\endgroup

\begin{figure*}[b]
    \centering
    \setlength{\tabcolsep}{1pt} 
    \renewcommand{\arraystretch}{0.5} 
    
    \begin{tabular}{ccccccc}
        \footnotesize Input & 
        \footnotesize KNN & 
        \footnotesize CRF & 
        \footnotesize Geometry & 
        \footnotesize MinkUNet & 
        \footnotesize VoxelFix (Ours) & 
        \footnotesize GT \\[2pt] 
        
        \includegraphics[width=0.138\textwidth]{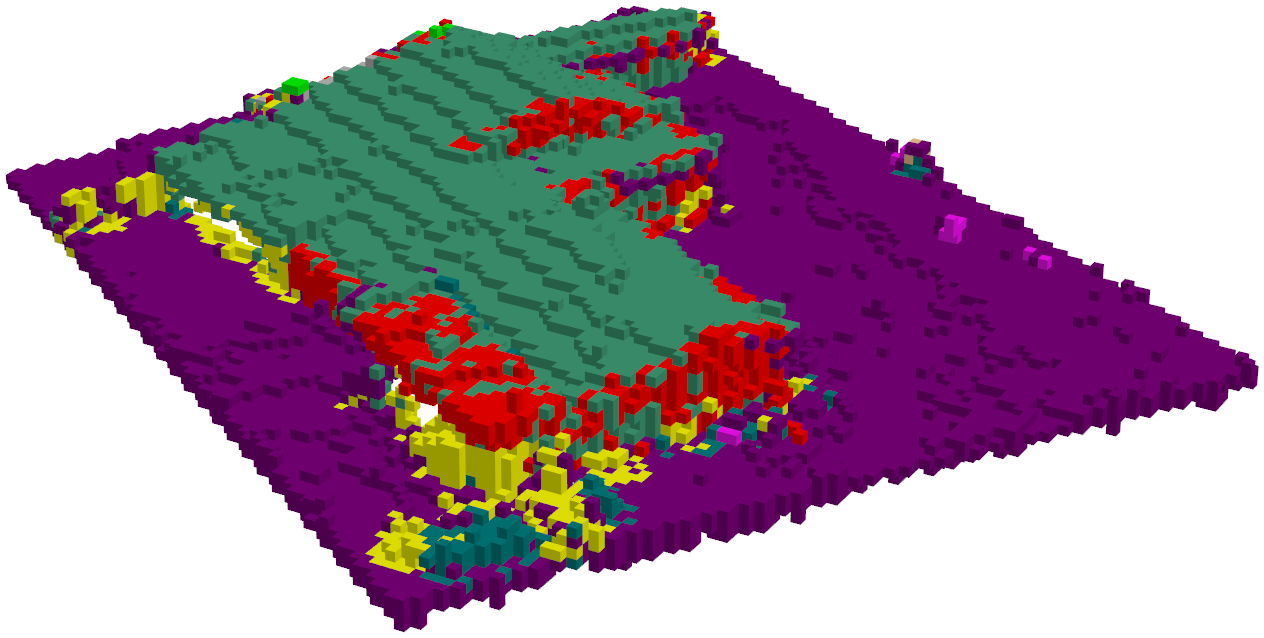} &
        \includegraphics[width=0.138\textwidth]{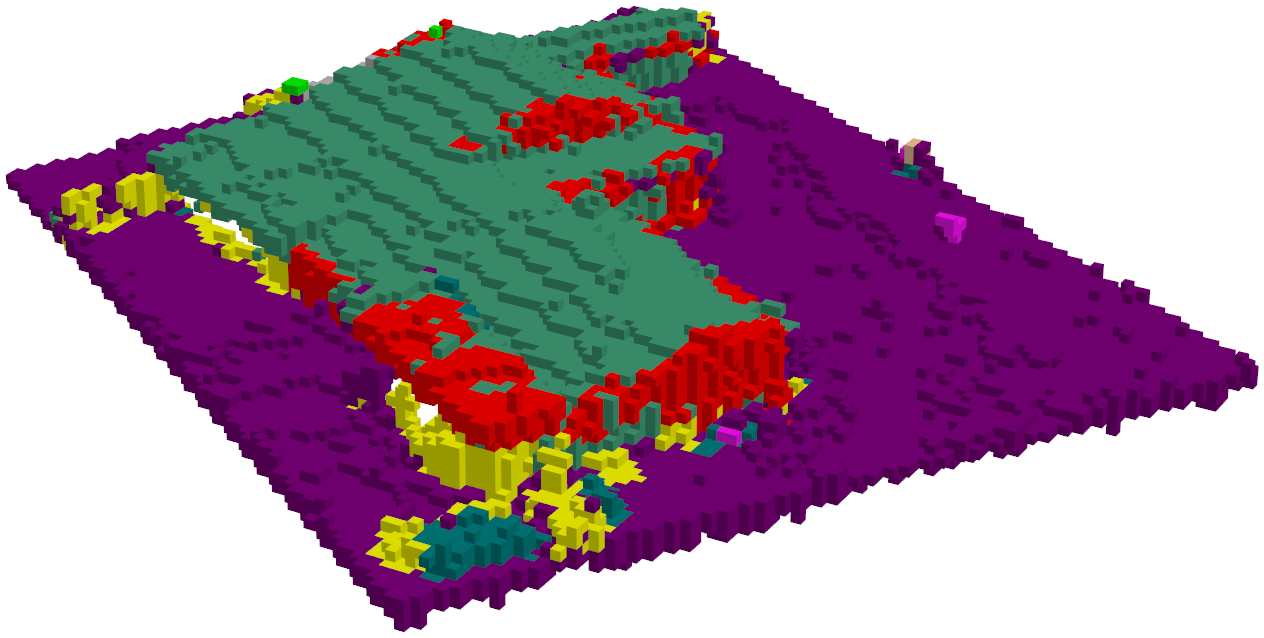} &
        \includegraphics[width=0.138\textwidth]{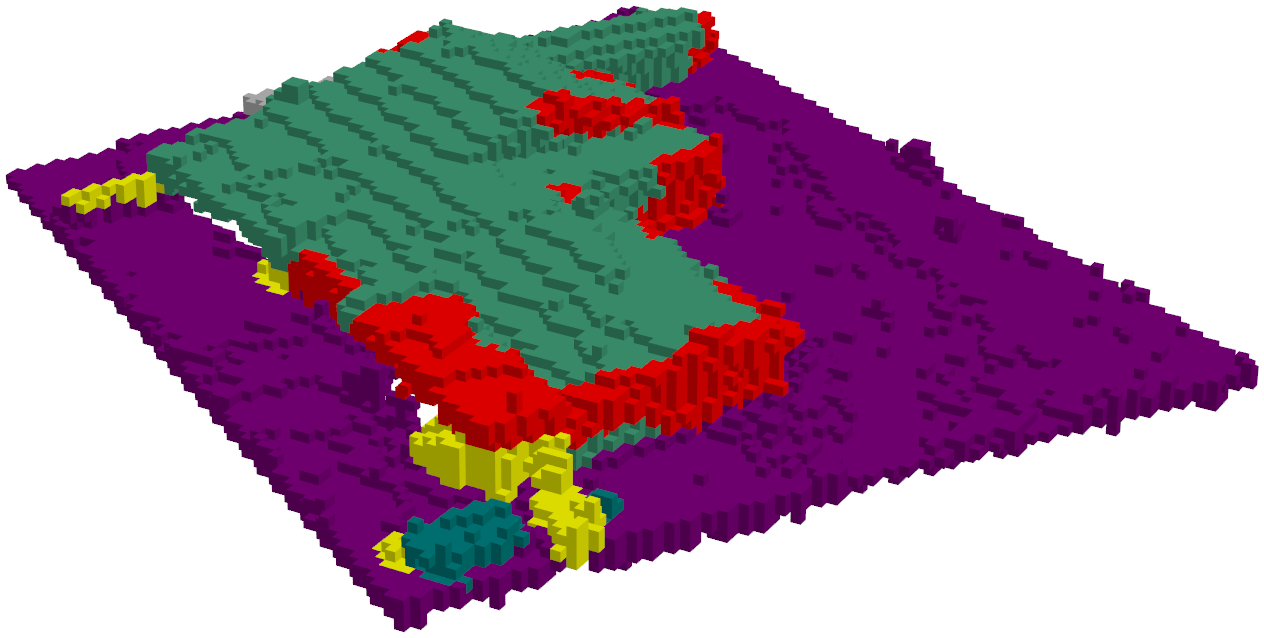} &
        \includegraphics[width=0.138\textwidth]{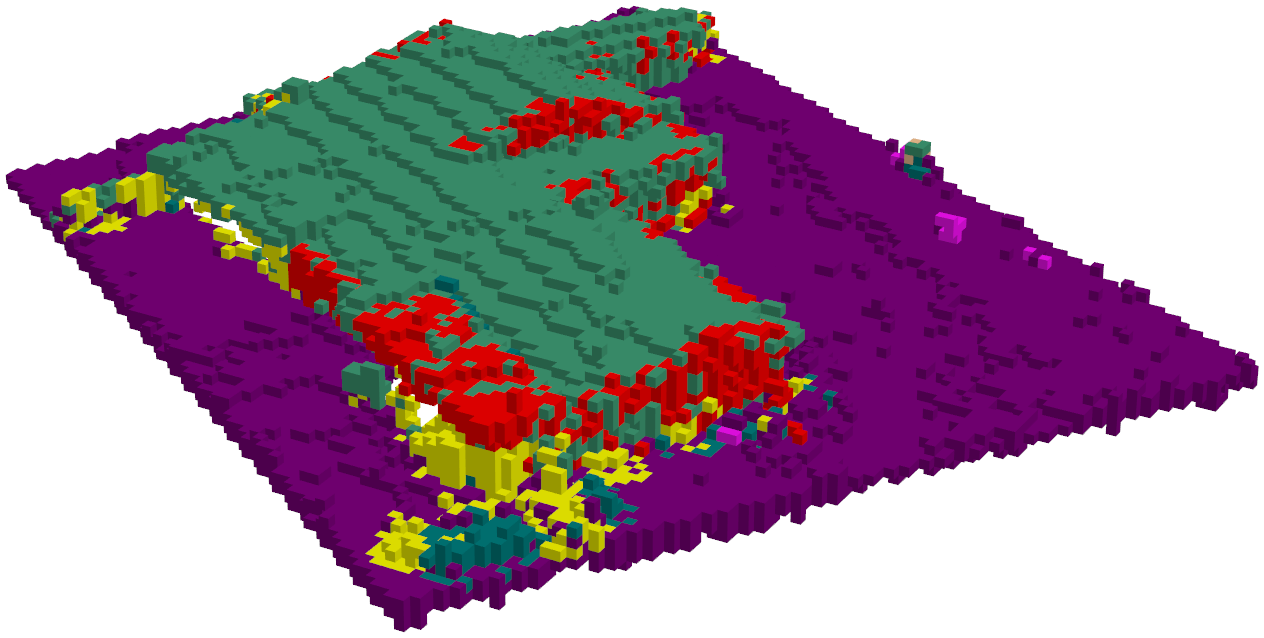} &
        \includegraphics[width=0.138\textwidth]{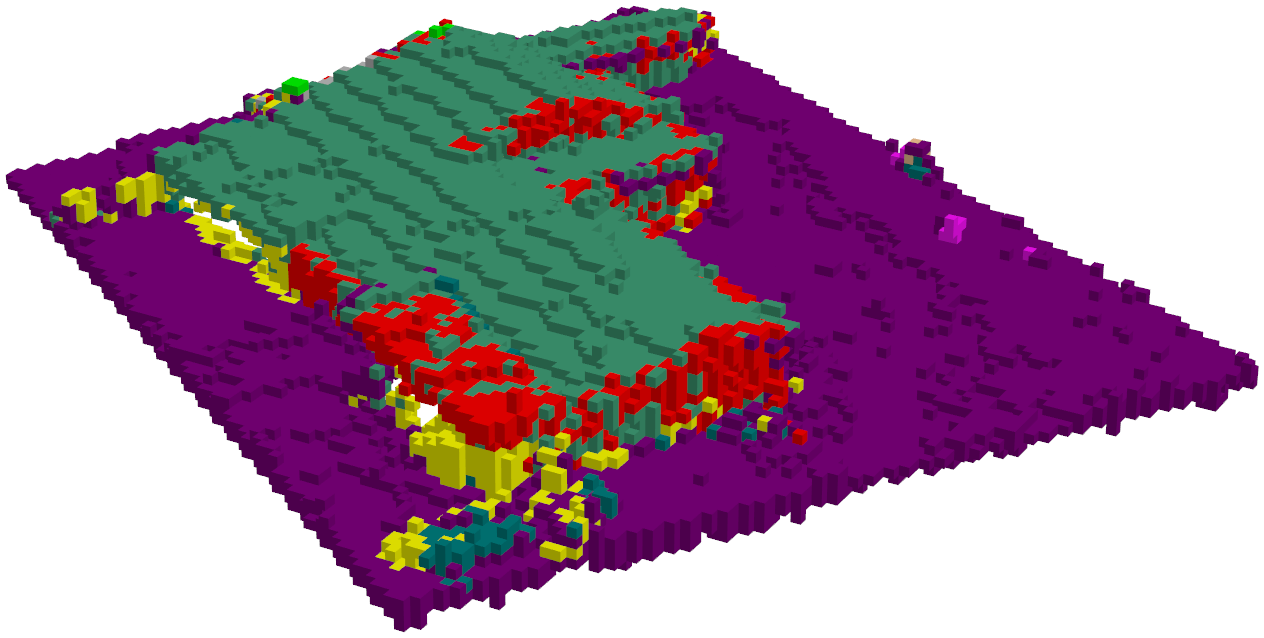} &
        \includegraphics[width=0.138\textwidth]{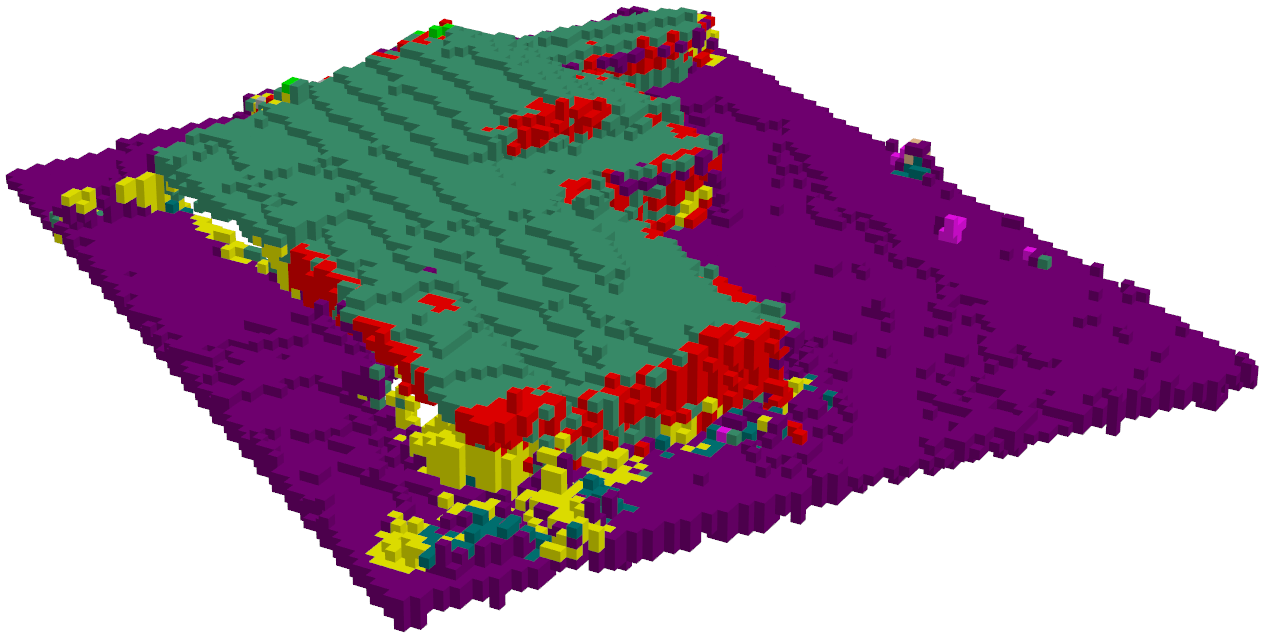} &
        \includegraphics[width=0.138\textwidth]{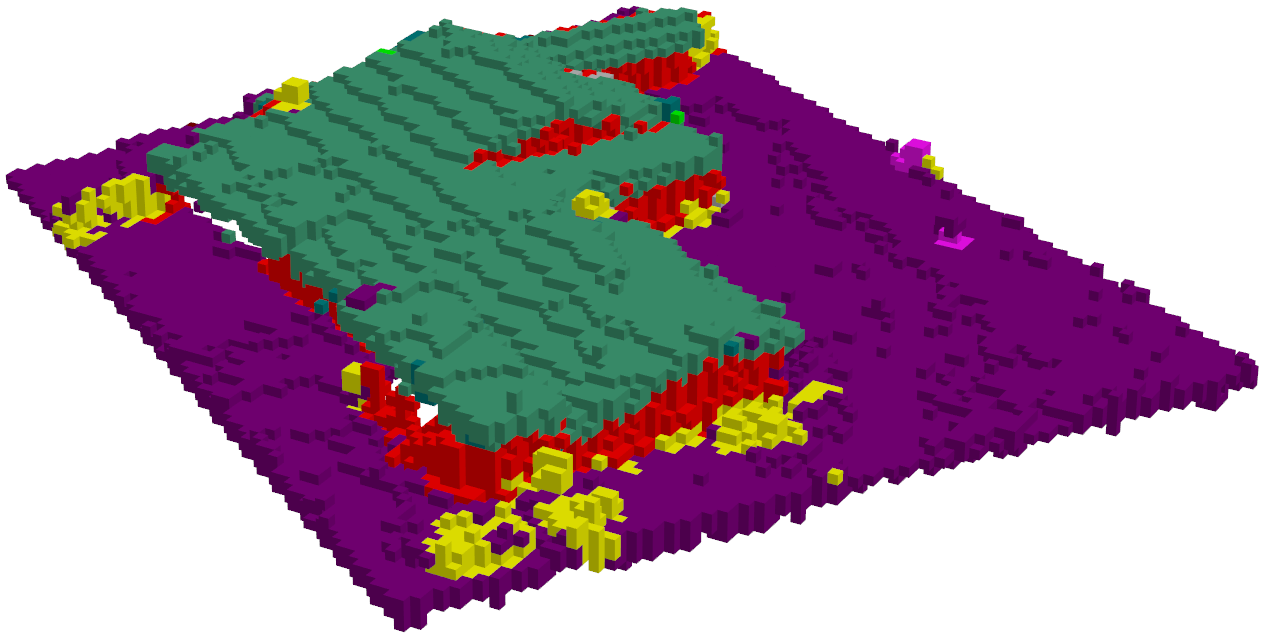} \\
        
        \includegraphics[width=0.138\textwidth]{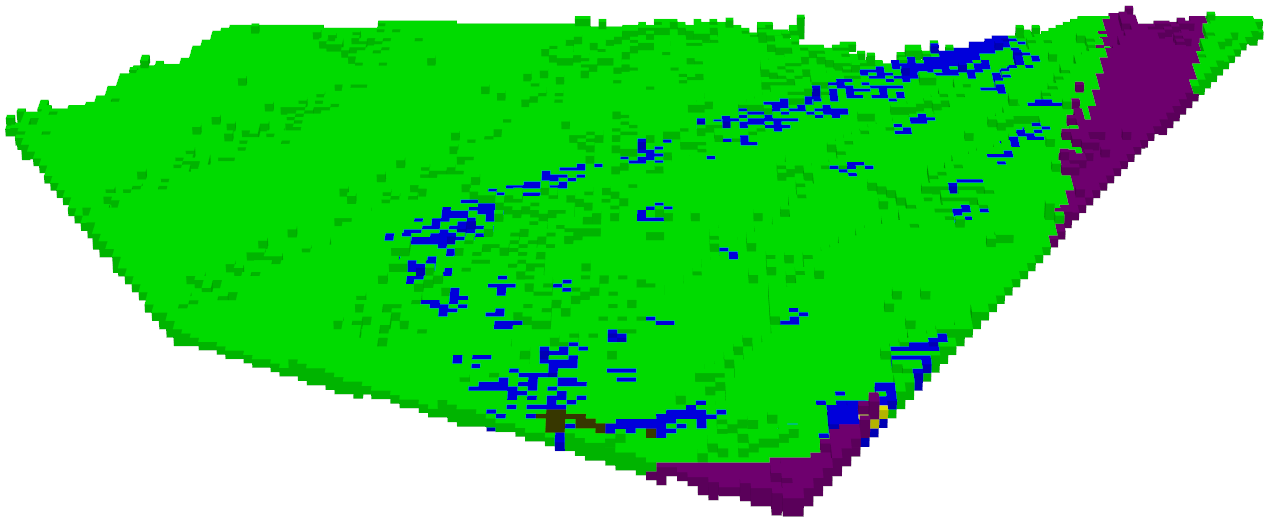} &
        \includegraphics[width=0.138\textwidth]{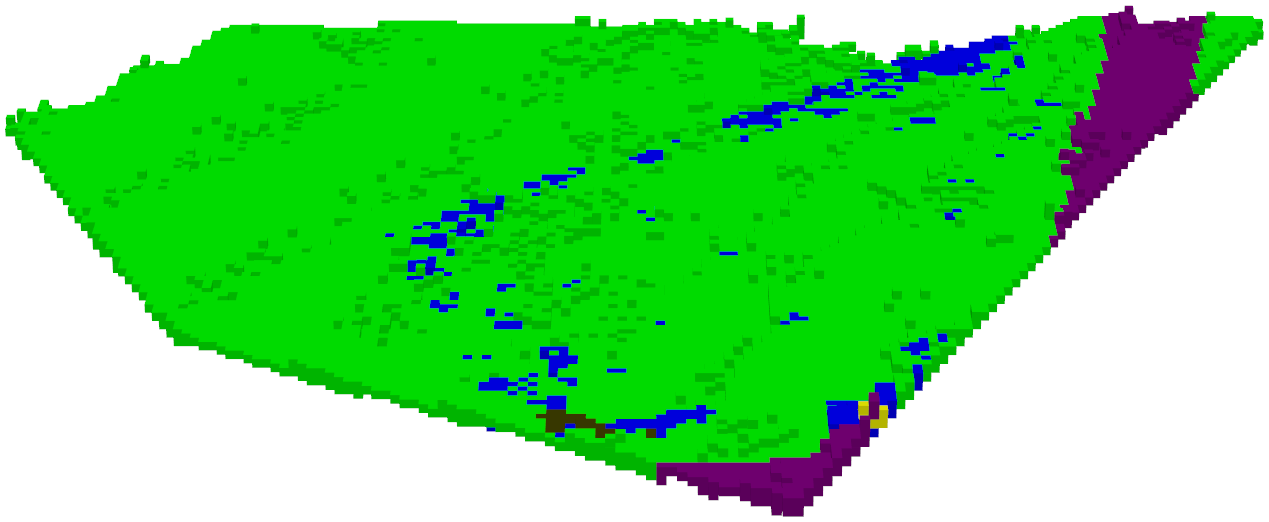} &
        \includegraphics[width=0.138\textwidth]{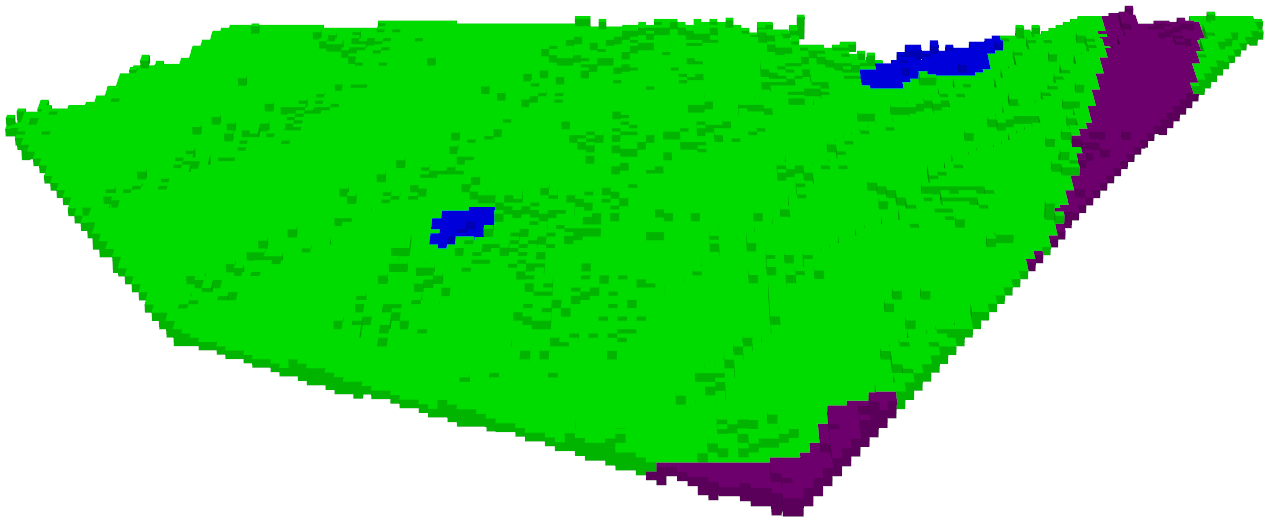} &
        \includegraphics[width=0.138\textwidth]{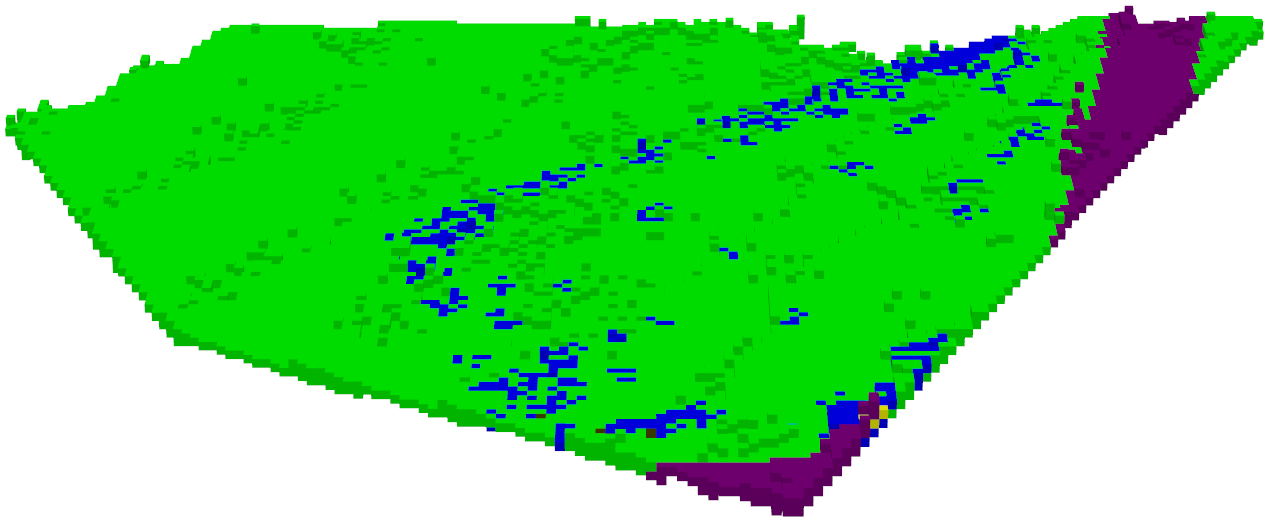} &
        \includegraphics[width=0.138\textwidth]{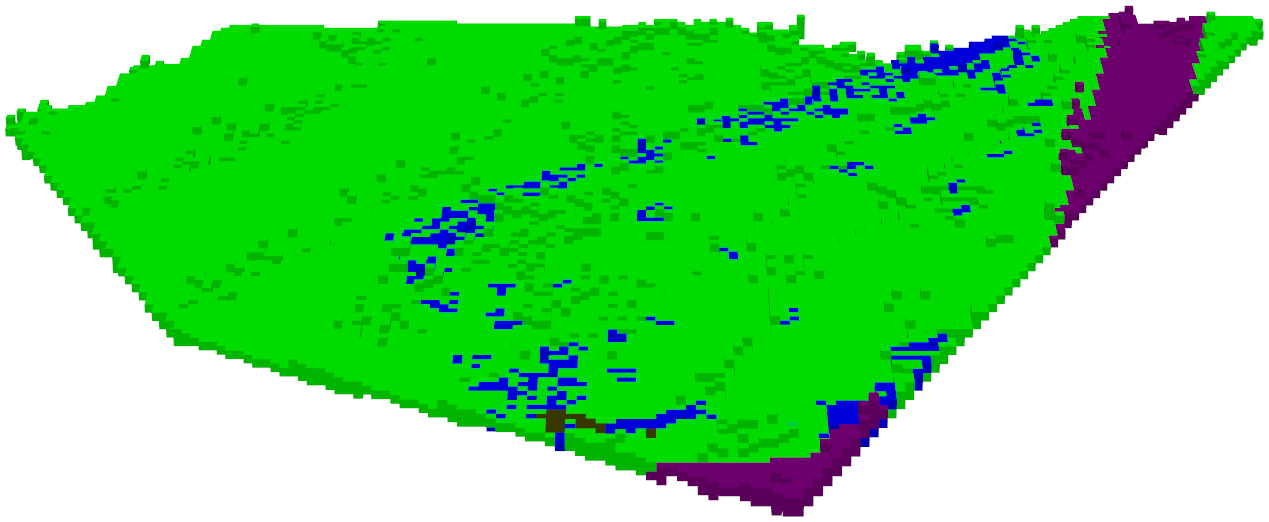} &
        \includegraphics[width=0.138\textwidth]{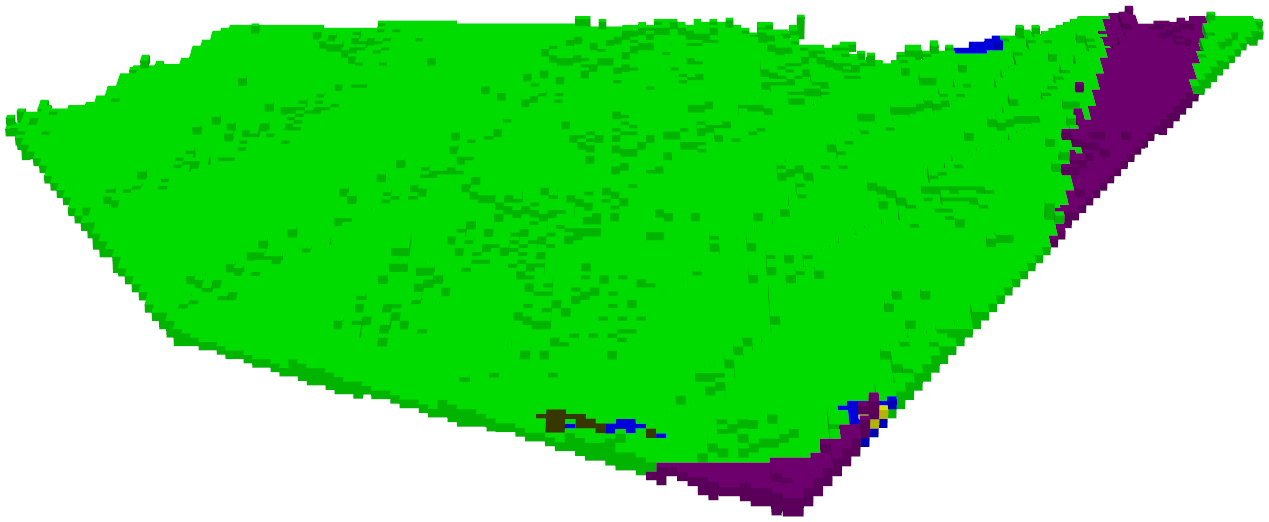} &
        \includegraphics[width=0.138\textwidth]{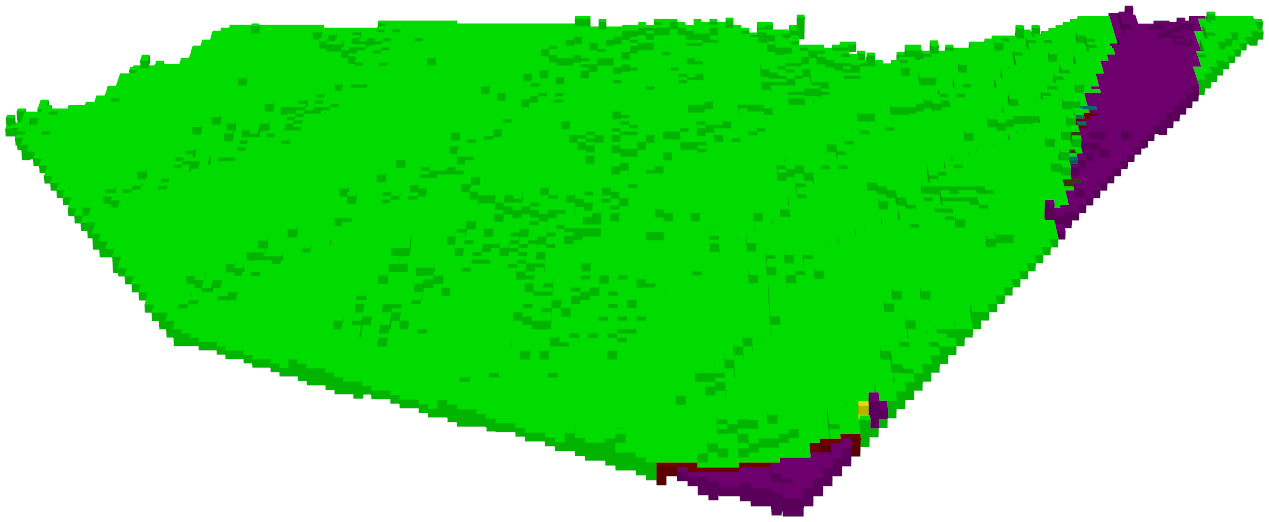} \\
        
        \includegraphics[width=0.138\textwidth]{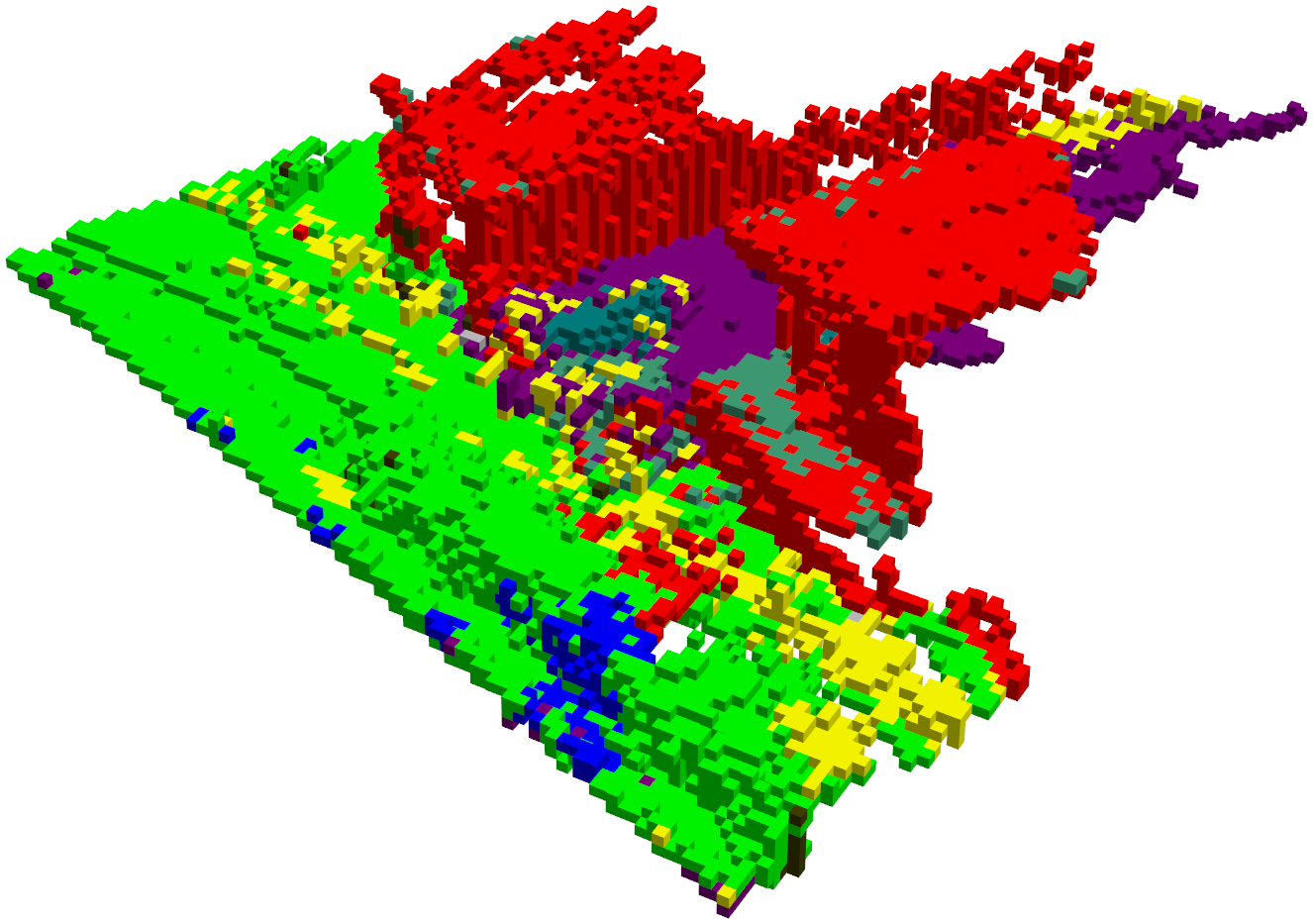} &
        \includegraphics[width=0.138\textwidth]{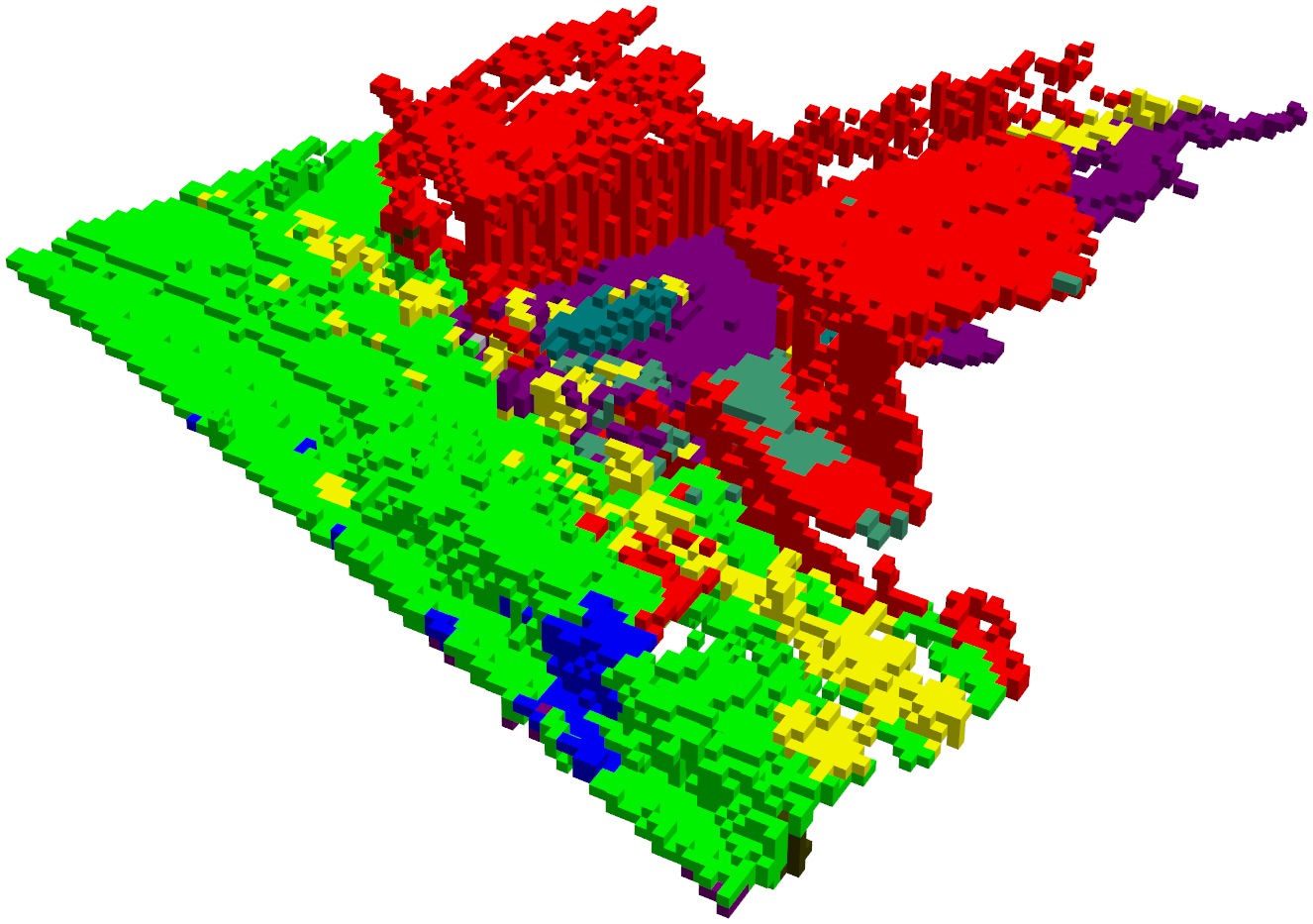} &
        \includegraphics[width=0.138\textwidth]{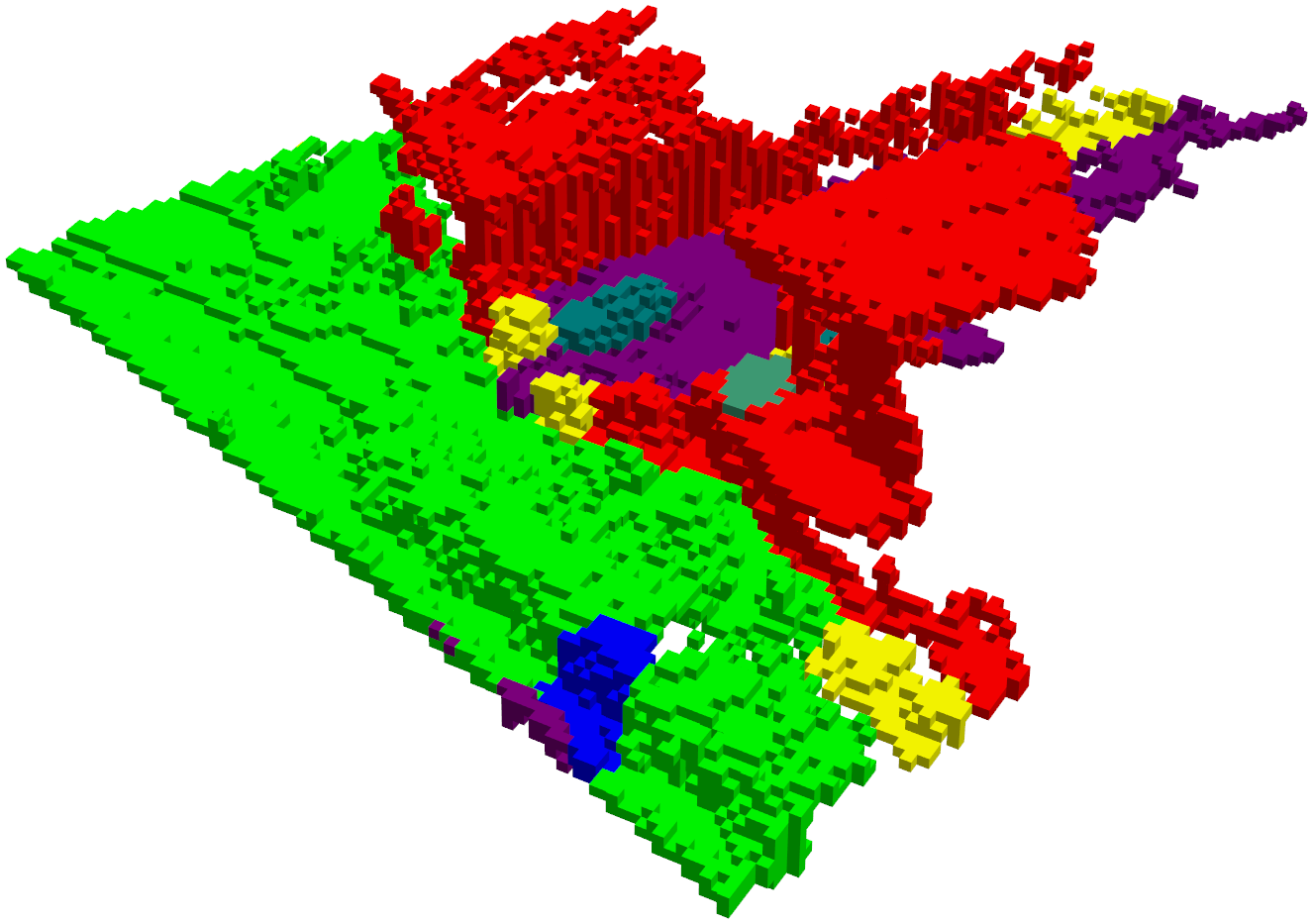} &
        \includegraphics[width=0.138\textwidth]{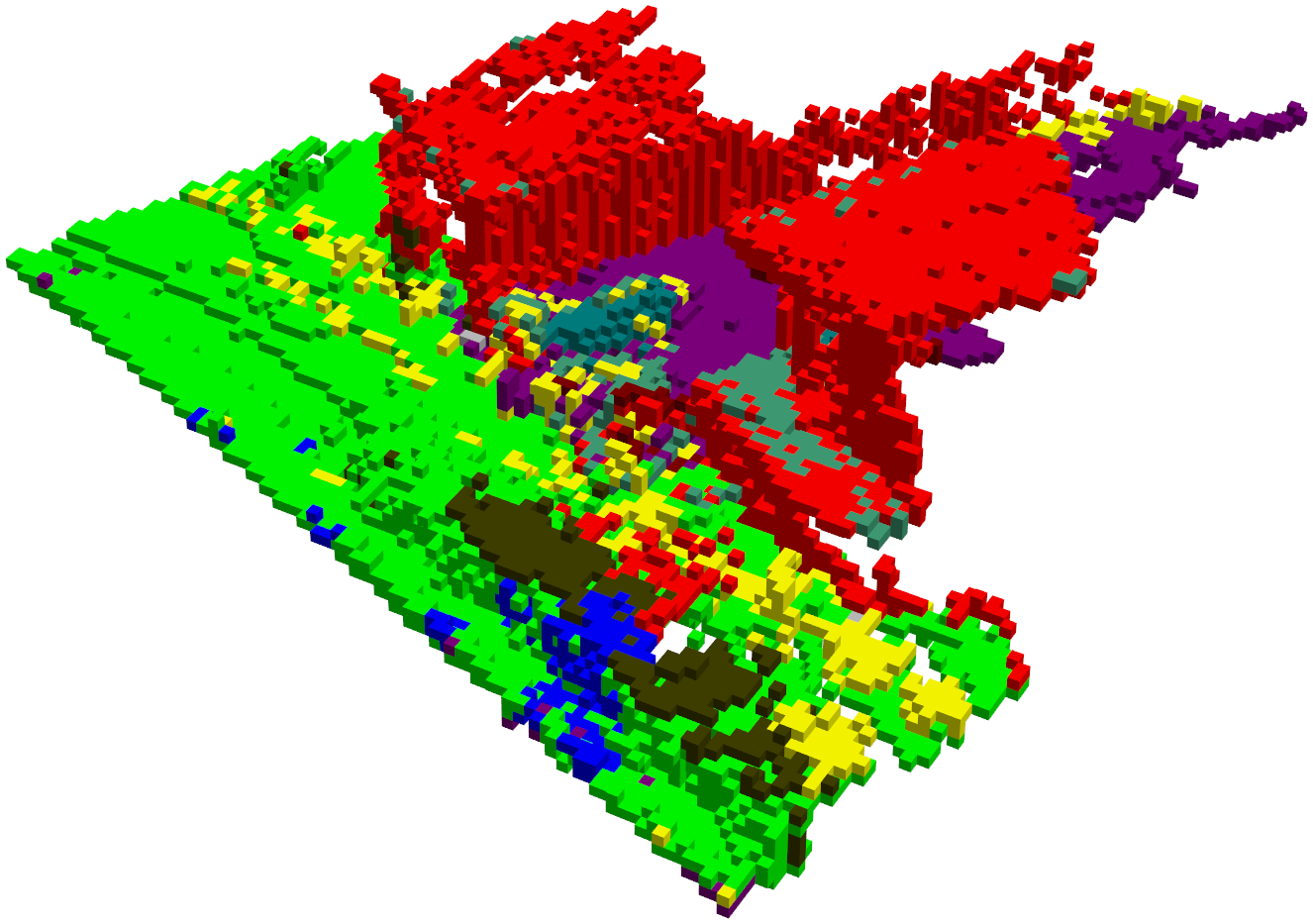} &
        \includegraphics[width=0.138\textwidth]{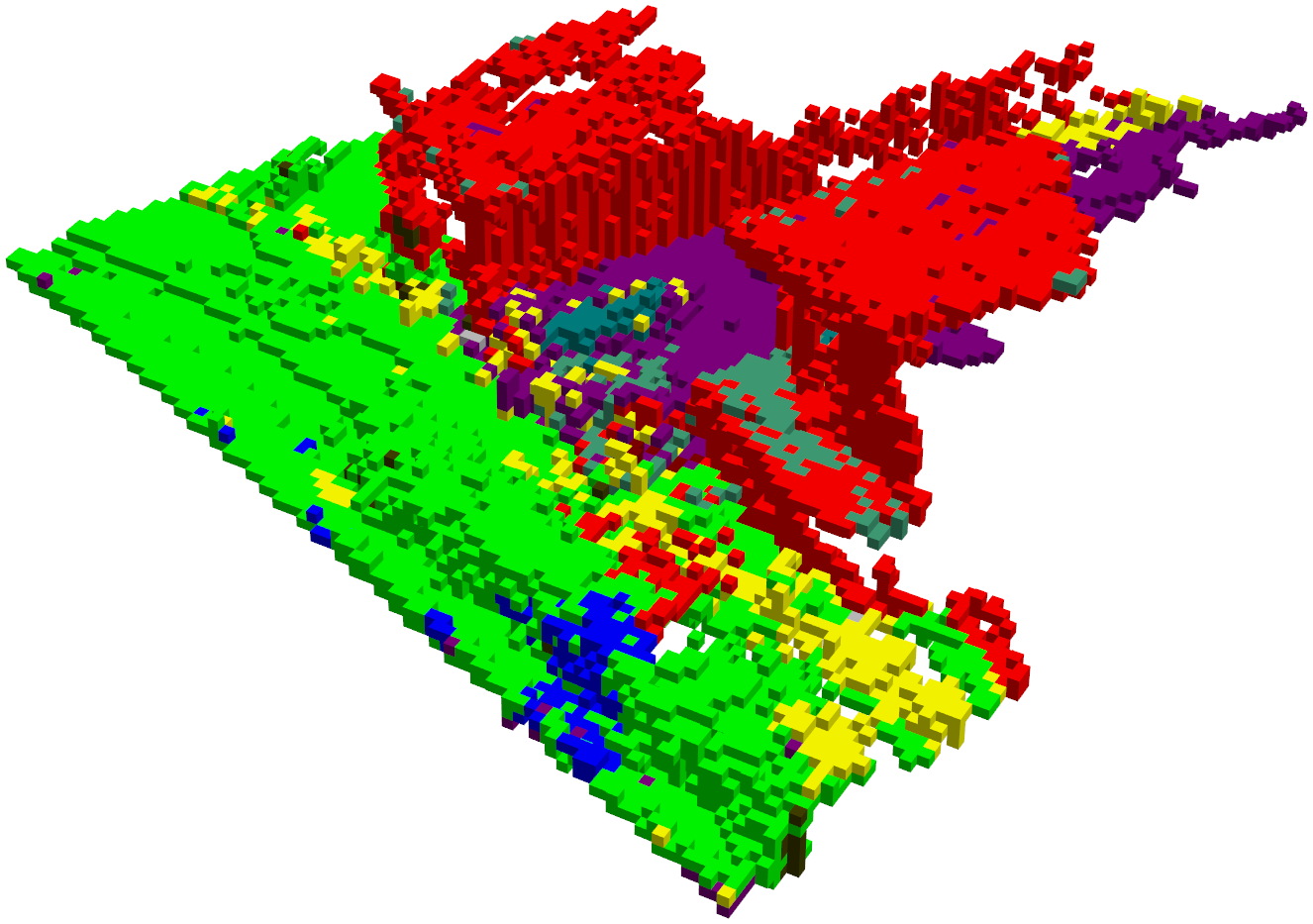} &
        \includegraphics[width=0.138\textwidth]{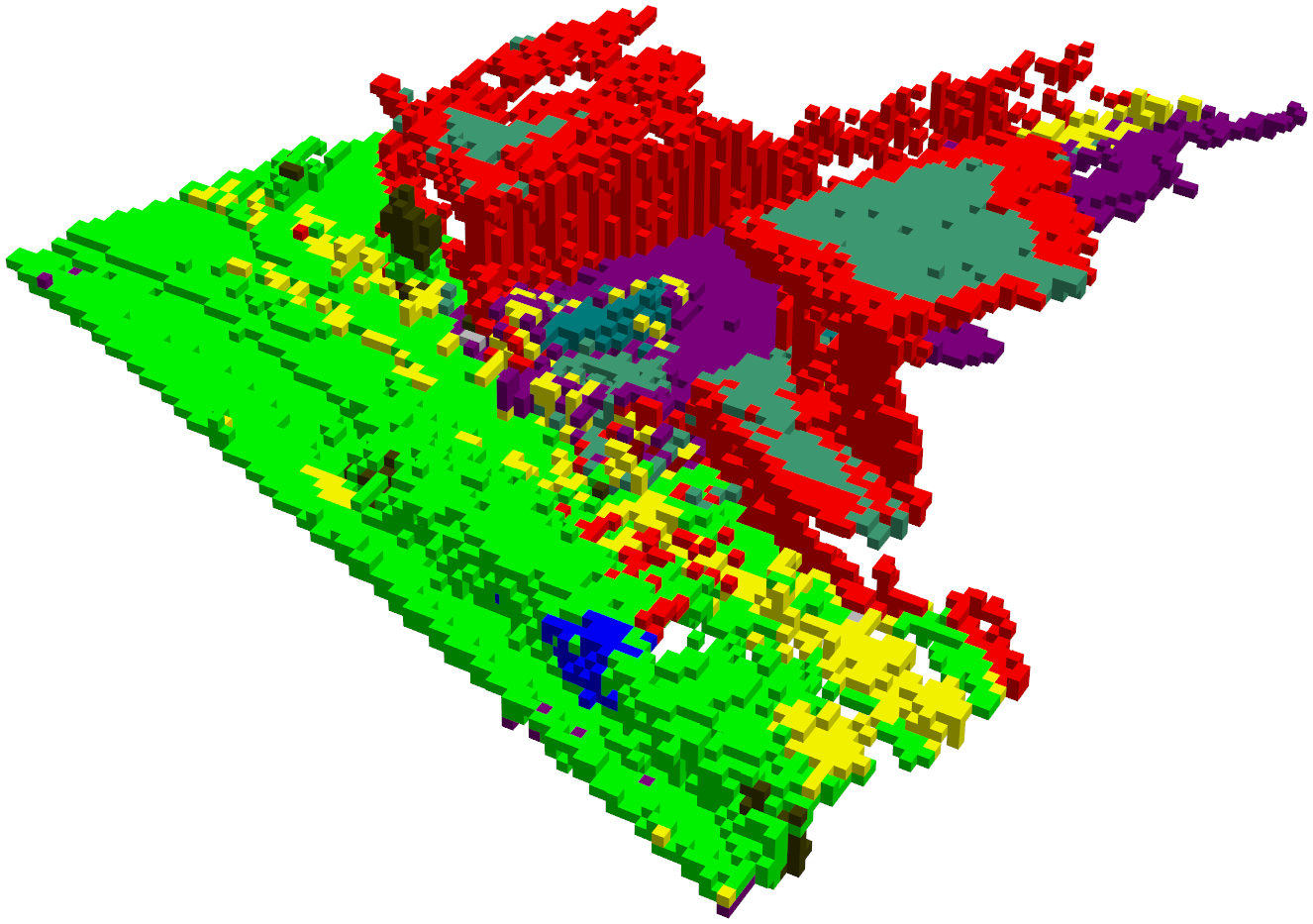} &
        \includegraphics[width=0.138\textwidth]{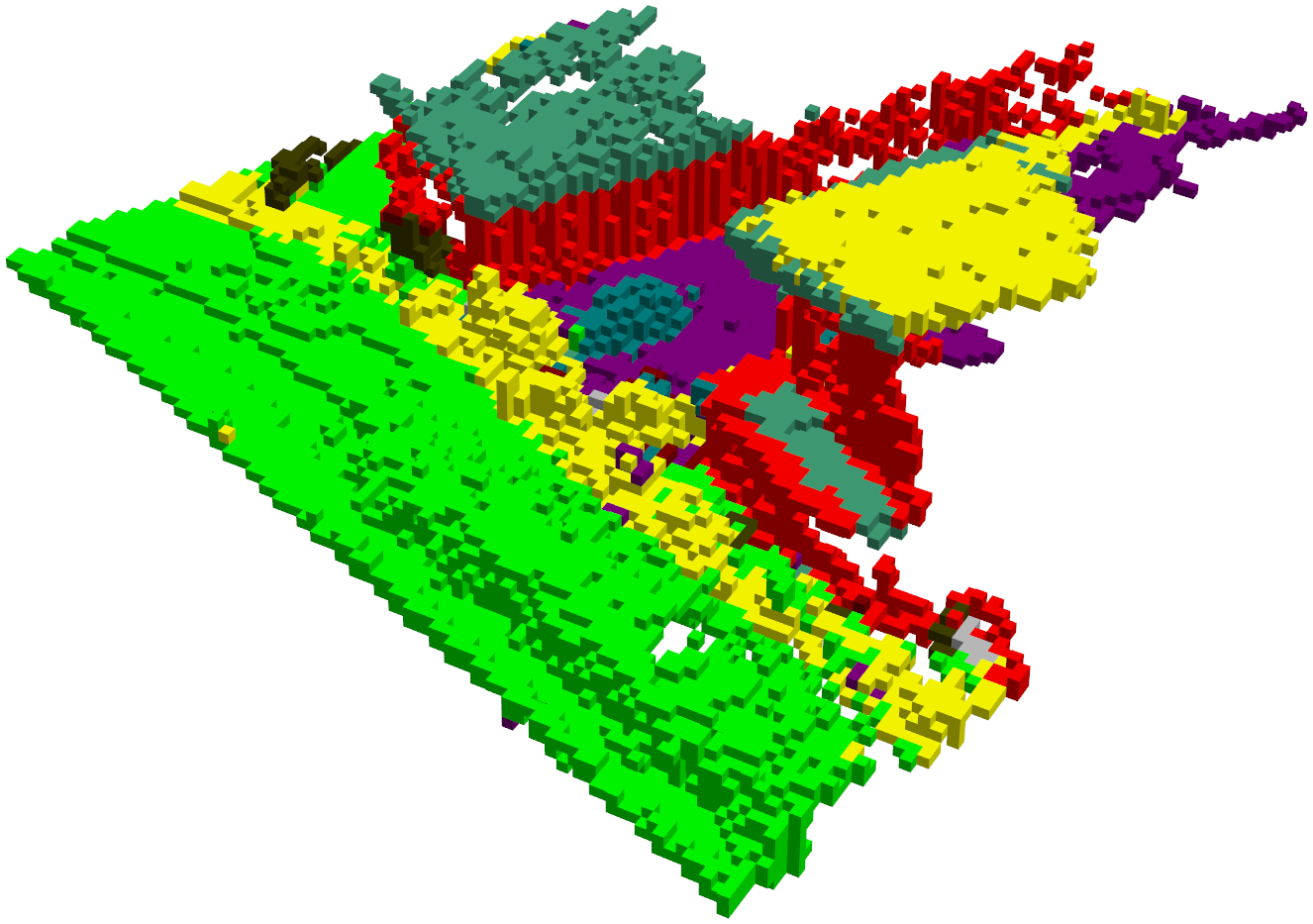} \\
    \end{tabular}
    \vspace{-0.75em}
    \caption{\textbf{Qualitative map-level correction on the OOD aerial scene using Mask2Former upstream predictions.}
    From left to right: uncorrected Radix input, KNN, CRF, geometry heuristic, MinkUNet, \method, and ground truth. \method corrects several semantic inconsistencies on previously unseen scene geometry while largely preserving correctly labeled regions.}
    \label{fig:ood}
\end{figure*}

\subsection{Ablation Study}
\label{sec:ablation}

Finally, we ablate the main components of \method to assess their contribution to map-level correction. \cref{tab:ablation} shows that all tested components contribute positively. The largest degradation occurs when the noise curriculum is restricted to low corruption levels or reversed, reducing mIoU by 4.01 and 3.62 points, respectively. Removing the auxiliary detection loss or the confusion-guided corruption results in further drops of 2.21 and 2.18 points, highlighting the importance of structured training supervision. Removing the learned volumetric geometry embeddings or the geometric branch each reduces mIoU by approximately 2 points, while a shallower GAT, ungated fusion, and removal of semantic priors lead to smaller but consistent decreases. Overall, the full model performs best, indicating that both the corruption strategy and complementary geometric--semantic reasoning contribute to effective map-level correction.
\begingroup
\setlength{\tabcolsep}{4pt}
\begin{table}[t!]
\centering
\caption{\textbf{Ablation of the main \method components on the fixed OccuFly test scene.} Results report mean $\pm$ std across the four upstream segmentation models, evaluated on a single held-out split of the 7/1/1 protocol.}
\renewcommand{\arraystretch}{0.9}
\resizebox{\linewidth}{!}{
\begin{tabular}{@{}l|ccc@{}}
\toprule
Variant & mIoU$\uparrow$ & $\Delta$mIoU$\uparrow$ & ECR$\uparrow$ \\
\midrule
w/o UNet embeddings (6D geometry only)
    & 27.43$\pm$1.43 & +3.18$\pm$0.36 & 12.31$\pm$2.01 \\
single-layer GAT 
    & 27.65$\pm$1.43 & +3.39$\pm$0.21 & 13.79$\pm$1.60 \\
w/o geometric branch
    & 27.43$\pm$1.47 & +3.17$\pm$0.49 & 14.77$\pm$3.00 \\
w/o gated fusion
    & 27.88$\pm$1.47 & +3.63$\pm$0.39 & 15.81$\pm$2.68 \\
reversed noise curriculum (0.05$\to$0.50)
    & 25.80$\pm$1.35 & +1.55$\pm$0.19 & 6.73$\pm$1.24 \\
low-noise curriculum (0.10$\to$0.02)
    & 25.41$\pm$1.20 & +1.15$\pm$0.28 & 4.95$\pm$1.16 \\
w/o confusion-guided noise
    & 27.24$\pm$1.38 & +2.98$\pm$0.31 & 12.84$\pm$1.70 \\
w/o semantic priors
    & 27.94$\pm$1.44 & +3.69$\pm$0.33 & 15.99$\pm$2.46 \\
w/o detection loss
    & 27.21$\pm$1.53 & +2.96$\pm$0.37 & 13.06$\pm$2.47 \\
\midrule
Full model
    & \textbf{29.42$\pm$1.33}
    & \textbf{+5.16$\pm$0.40}
    & \textbf{24.53$\pm$2.56} \\
\bottomrule
\end{tabular}
}
\label{tab:ablation}
\end{table}
\endgroup
\section{Conclusion}
\label{sec:conclusion}

We presented \method, a map-only approach for post-hoc semantic correction of
completed voxel maps. Rather than revisiting the observations or reconstruction
process, \method operates directly on the completed map and refines its semantic
labels while keeping geometry and occupancy fixed. By combining geometric and
semantic graph reasoning with confusion-guided training corruptions, \method
consistently improves maps produced by four different upstream segmentation
models and outperforms local, geometry-based, and learned volumetric refinement
baselines. Our results further show that useful correction can transfer to an
independently reconstructed scene, supporting post-hoc semantic correction as
a complementary stage in modular semantic mapping pipelines.

Our formulation is nevertheless bounded by the information retained in the
completed map. In particular, \method cannot recover missing geometry or correct
reconstruction errors, and its geometric reasoning assumes that the reconstructed
structure is sufficiently reliable. Our experiments isolate semantic errors by
constructing maps with ground-truth depth and camera poses; robustness to
geometric errors, therefore, remains to be evaluated. Correction also remains
challenging when the map provides insufficient evidence for the correct class,
as observed for ambiguous structures in the OOD evaluation. Future work could
address these limitations by incorporating geometric uncertainty and evaluating
correction under increasingly noisy reconstruction conditions while retaining
the map-only formulation.



{
    \small
    \bibliographystyle{ieeenat_fullname}
    \bibliography{main}
}

\clearpage
\setcounter{page}{1}
\maketitlesupplementary

\section{Additional Implementation Details}
\label{sec:supp_impl}
This section provides the architectural, optimization, and training details required to reproduce VoxelFix. Exact model and training hyperparameters are summarized in \cref{tab:supp_arch,tab:supp_training}, while the following sections describe the corresponding feature construction, training procedure, and geometry-encoder setup.

\subsection{VoxelFix Architecture}
\label{sec:supp_arch}

VoxelFix follows the dual-branch graph-attention architecture described in the main paper, comprising geometric and semantic GATv2 branches, voxel-wise gated fusion, a semantic correction head, and an auxiliary error-detection head. The complete architectural configuration is summarized in \cref{tab:supp_arch}.

The node descriptor concatenates the handcrafted geometric descriptor $\mathbf{g}_i$, the frozen geometry embedding $\mathbf{z}_i$, a $K$-dimensional one-hot encoding of the current semantic label $\hat{y}_i$, the scalar class-conditioned agreement score $a_i$, and the prototype deviation $\boldsymbol{\delta}_i$. Each feature group is independently projected before being combined into the shared representation used by both graph branches. The two branch outputs are fused using learned voxel-wise confidence weights, and the resulting representation is passed to the correction and error-detection heads.

\begin{table}[t]
    \centering
    \caption{\textbf{VoxelFix architectural configuration.}}
    \label{tab:supp_arch}
    \small
    \setlength{\tabcolsep}{5pt}
    \renewcommand{\arraystretch}{1.05}
    \begin{tabular}{lc}
        \toprule
        Parameter & Value \\
        \midrule
        Graph neighbors $k$ & 30 \\
        Structure-tensor neighbors $k_{\mathrm{geo}}$ & 50 \\
        Density radius $r$ & 3 m \\
        Shared hidden dimension $d$ & 32 \\
        GATv2 layers per branch & 2 \\
        Attention heads per layer & 2 \\
        GAT hidden dimension & 32 \\
        Dropout probability & 0.1 \\
        Geometry descriptor dimension & 6 \\
        Frozen geometry embedding dimension & 16 \\
        Fusion MLPs & $32\!\rightarrow\!16\!\rightarrow\!1$ \\
        Correction head & $32\!\rightarrow\!K$ \\
        Error-detection head & $32\!\rightarrow\!1$ \\
        Trainable parameters &  26,734\\
        \bottomrule
    \end{tabular}
\end{table}

\subsection{VoxelFix Training}
\label{sec:supp_training}

VoxelFix is trained on synthetically corrupted semantic voxel maps using the optimization configuration summarized in \cref{tab:supp_training}. Validation is performed on the held-out validation scene of each split, and the checkpoint with the lowest validation loss is retained. Class prototypes are initialized from the corresponding training scenes and updated during training using an exponential moving average.

\begingroup
\setlength{\tabcolsep}{3.5pt}
\begin{table}[t!]
    \centering
    \caption{\textbf{VoxelFix training configuration.}}
    \label{tab:supp_training}
    \footnotesize
    \renewcommand{\arraystretch}{0.9}
    \begin{tabular}{@{}lc@{}}
        \toprule
        Parameter & Value \\
        \midrule
        Optimizer & AdamW \\
        Initial learning rate & $1\times10^{-3}$ \\
        Weight decay & $1\times10^{-4}$ \\
        Epochs & 500 max., early stopping (patience 20) \\
        LR scheduler & ReduceLROnPlateau ($\times0.5$, patience 5) \\
        Batch / subgraph size & 1 scene graph / tile \\
        Max.\ tile size & $5\times10^{5}$ voxels \\
        Gradient clipping & max-norm 1.0 \\
        $\lambda_{\mathrm{CE}}$ & 1.0 \\
        $\lambda_{\mathrm{det}}$ & 0.5 \\
        Prototype EMA coefficient & 0.999 (update rate $0.001$) \\
        Checkpoint criterion & Best validation loss \\
        Random seed & 42 \\
        \bottomrule
    \end{tabular}
\end{table}
\endgroup

\paragraph{Compute and training budget.}
All experiments were executed on a single NVIDIA A100 80\,GB GPU.
The geometry encoder was trained in two stages, consisting of STPLS3D
pretraining followed by OccuFly fine-tuning, after which the frozen geometry encoder was reused across cross-scene splits. VoxelFix was then trained independently for each cross-scene split using the corresponding frozen geometry encoder. Approximate wall-clock costs for each stage are reported in \cref{tab:supp_compute}.

\begin{table}[t]
    \centering
    \caption{\textbf{Training compute and approximate wall-clock cost.}}
    \label{tab:supp_compute}
    \small
    \begin{tabular}{lc}
        \toprule
        Stage & Configuration / time \\
        \midrule
        Hardware & 1$\times$ NVIDIA A100 80\,GB \\
        STPLS3D pretraining & $\sim$8 h \\
        OccuFly fine-tuning & $\sim$1 h \\
        VoxelFix training & $\sim$3 h per split \\
        \bottomrule
    \end{tabular}
\end{table}

\paragraph{Augmentation.}
No additional geometric augmentation is applied during VoxelFix training. Training-time variation is introduced through the synthetic semantic corruption described in \cref{sec:supp_noise}, which generates spatially contiguous label errors according to the confusion-guided transition distribution. The corresponding corruption ranges and probabilities are reported in \cref{sec:supp_noise}.

\paragraph{Hyperparameter selection.}
Hyperparameters were selected using only the corresponding training and validation scenes. The fixed test scene was not used for hyperparameter or checkpoint selection, feature normalization, prototype initialization, or confusion-statistics estimation. The final configuration used in all experiments is reported in \cref{tab:supp_arch,tab:supp_training}.

\section{Geometry Encoder}
\label{sec:supp_geo_encoder}

VoxelFix incorporates a learned geometry representation to complement the handcrafted geometric descriptor. We obtain this representation using a sparse volumetric U-Net trained in two stages: pretraining on STPLS3D, followed by a single fine-tuning on OccuFly. Importantly, semantic labels are used only as supervision for training the geometry encoder; its input consists exclusively of geometric voxel features. After fine-tuning, the classification head is discarded and the resulting 16-dimensional per-voxel embeddings are kept fixed during VoxelFix training.

\subsection{Sparse Volumetric U-Net}
Both STPLS3D and OccuFly are voxelized at a resolution of $0.5$\,m. For each occupied voxel, we compute a six-dimensional geometric descriptor 
comprising normalized height, planarity, linearity, sphericity, verticality, and local point density. These geometry-only descriptors form the input to the sparse U-Net. The network consists of four encoder stages and a symmetric decoder with additive skip connections, followed by a dedicated 16-dimensional embedding head. The complete architecture is summarized in \cref{tab:supp_geo_arch}.

\begingroup
\setlength{\tabcolsep}{3.5pt}
\begin{table}[t]
    \centering
    \caption{\textbf{Geometry-encoder architecture.}}
    \label{tab:supp_geo_arch}
    \footnotesize
    \renewcommand{\arraystretch}{0.92}
    \begin{tabular}{@{}l p{0.54\linewidth}@{}}
        \toprule
        Parameter & Value \\
        \midrule
        Sparse U-Net backbone &
        MinkowskiEngine sparse U-Net, additive skip connections \\
        
        Input voxel features &
        6D geometric descriptor (height, planarity, linearity,
        sphericity, verticality, density) \\
        
        Encoder stages &
        4 (1 stride-1 stem + 3 stride-2 blocks) \\
        
        Channel configuration &
        $6\!\rightarrow\!32\!\rightarrow\!64\!\rightarrow\!128
        \!\rightarrow\!256$ (encoder),
        $256\!\rightarrow\!128\!\rightarrow\!64\!\rightarrow\!32$
        (decoder) \\
        
        Output embedding dimension & 16 \\
        Voxel size / quantization & 0.5\,m \\
        \bottomrule
    \end{tabular}
\end{table}
\endgroup

\subsection{STPLS3D Pretraining}
The geometry encoder is first pretrained for voxel-wise semantic classification on STPLS3D to learn transferable geometric representations
from a larger collection of outdoor scenes. Training uses only the six-dimensional geometric descriptor as input, while the grouped semantic classes provide the supervision targets. To improve robustness to incomplete aerial observations, we additionally employ partial-scan augmentation, which randomly removes part of the observed scene. The complete pretraining configuration is reported in \cref{tab:supp_stpls3d}.

\begingroup
\setlength{\tabcolsep}{3.5pt}
\begin{table}[t]
    \centering
    \caption{\textbf{STPLS3D pretraining configuration for the geometry encoder.}}
    \label{tab:supp_stpls3d}
    \footnotesize
    \renewcommand{\arraystretch}{0.92}
    \begin{tabular}{@{}lc@{}}
        \toprule
        Parameter & Value \\
        \midrule
        Training split & 60 / 6 / 1 scenes (train / val / test, random) \\
        Number of classes & 8 (grouped) \\
        Optimizer & AdamW \\
        Learning rate & $1\times10^{-3}$ \\
        Weight decay & $1\times10^{-4}$ \\
        Number of epochs & 1000 max., early stop (patience 250) \\
        Loss & Cross-entropy, inverse-frequency class weights \\
        Batch size & 1 scene \\
        Augmentation & Partial-scan augmentation, $p=0.5$ \\
        \bottomrule
    \end{tabular}
\end{table}
\endgroup

\subsection{OccuFly Fine-Tuning}
The pretrained geometry encoder is adapted to the OccuFly domain once and then reused for all cross-scene splits. Fine-tuning uses the seven training scenes of split~1, with Scene 6 for checkpoint selection; the fixed test scene (Scene 8) is excluded from fine-tuning and from all training-derived statistics. Geometric input features are normalized using statistics computed from these seven scenes only, and the same statistics are applied whenever the frozen encoder is evaluated. During fine-tuning the encoder, decoder, and embedding head remain trainable, while the classification head is reinitialized for the OccuFly supervision taxonomy. Because a single encoder is shared across splits, it was trained on the scene used for validation in splits~2--8; validation loss in those splits is therefore mildly optimistic, while the fixed test scene remains unseen
throughout. The optimization and augmentation settings are summarized in
\cref{tab:supp_occufly_finetune}. After fine-tuning, the classification head is discarded and the geometry encoder is frozen for all subsequent VoxelFix training.

\begingroup
\setlength{\tabcolsep}{3pt}
\begin{table}[t]
    \centering
    \caption{\textbf{OccuFly fine-tuning configuration for the geometry encoder.}}
    \label{tab:supp_occufly_finetune}
    \footnotesize
    \renewcommand{\arraystretch}{0.92}
    \begin{tabular}{@{}p{0.32\linewidth}p{0.62\linewidth}@{}}
        \toprule
        Parameter & Value \\
        \midrule
        Optimizer &
        AdamW \\

        Learning rate &
        $5\times10^{-4}$, cosine decay after 5-epoch warmup \\

        Weight decay &
        $1\times10^{-5}$ \\

        Number of epochs &
        200 max., early stop (patience 40) \\

        Loss &
        Cross-entropy \\

        Class weighting &
        Effective-number ($\beta=0.999$), clipped to $[0.25,4.0]$ \\

        Batch size &
        2 scenes \\

        Augmentation &
        Random $z$-rotation, translation jitter ($\sigma=0.2$\,m),
        scale ($0.9$--$1.1\times$), and $x/y$ flips \\

        Checkpoint selection &
        Best validation loss \\
        \bottomrule
    \end{tabular}
\end{table}
\endgroup

\section{Confusion-Guided Noise Curriculum}
\label{sec:supp_noise}

\subsection{Construction of the Class-Transition Mapping}

To ensure that the synthetic corruptions reflect plausible real-world errors, we base our noise generation on voxel-level confusion patterns between the completed upstream semantic maps and the ground-truth semantic voxel maps. First, we compute the raw confusion frequencies:
\begin{equation}
    C_{ab} = \bigl|\{\, i \mid y_i = a,\ \hat{y}_i = b \,\}\bigr|,
\end{equation}
which counts the voxels whose ground-truth label is $a$ and whose upstream predicted label is $b$. Row normalization yields the transition frequencies:
\begin{equation}
    T_{ab} = \frac{C_{ab}}{\sum_{b'} C_{ab'}}.
\end{equation}

While a fully probabilistic approach would sample directly from $T_{ab}$, our pipeline employs a more controlled, heuristic method. We analyze the most prominent errors in $T_{ab}$ to identify the primary failure modes of the upstream model. Based on this analysis, we manually construct a class-transition mapping that links each ground-truth class to a discrete set of its most common misclassifications. For example, if the empirical statistics show that trees are frequently confused with grass, roads, and water, the transition rule is explicitly defined as \textit{tree: [grass, road, water]}. 

\begin{figure}[t]
    \centering
    \includegraphics[width=0.95\columnwidth]{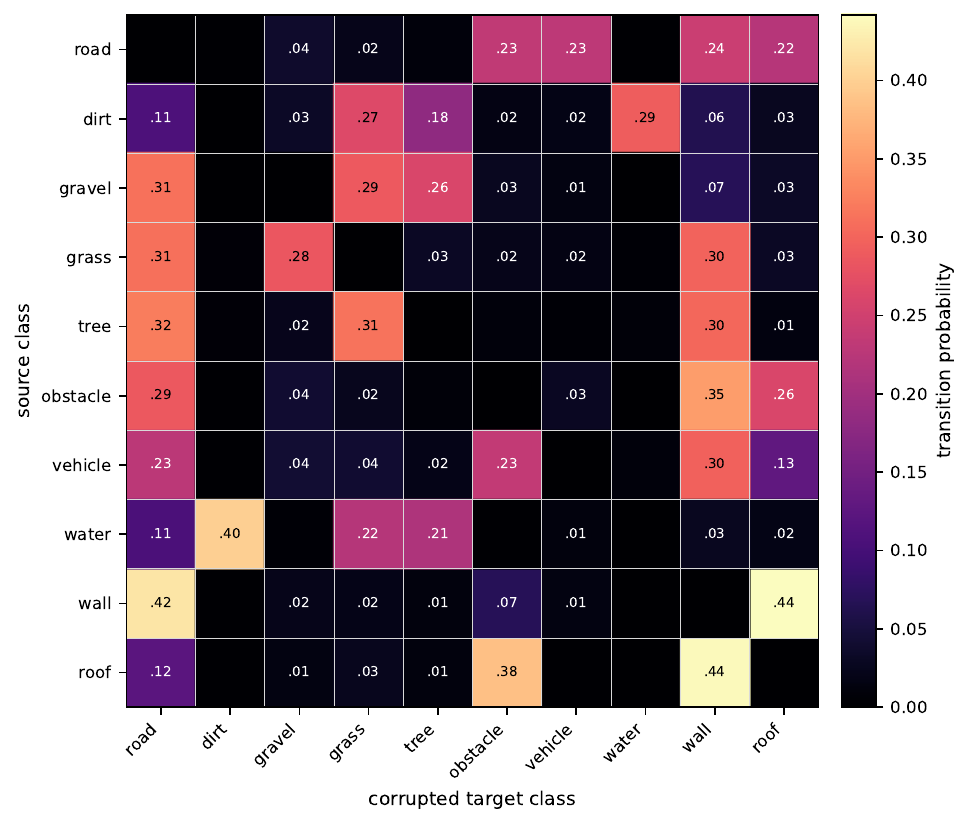}
    \caption{\textbf{Distribution of semantic substitutions for synthetic corruption.} The heatmap visualizes the transition probabilities derived from our heuristically constructed class-transition mapping. Plausible substitutions are restricted to the most prominent voxel-level confusions observed in the training scenes, resulting in a targeted prior for realistic synthetic errors.}
    \label{fig:supp_transition_matrix}
\end{figure}

During the corruption process, when a seed voxel is selected for replacement, its new label is drawn from its corresponding list of plausible errors in this handcrafted mapping. This ensures that synthetic errors follow realistic directions without being overly sensitive to long-tail statistical noise. Minor classes, such as person and bicycle, are excluded from these mapping entries. \cref{fig:supp_transition_matrix} illustrates the resulting distribution of noise changes generated using this transition mapping.

\begin{table}[t]
    \centering
    \caption{\textbf{Synthetic corruption parameters.}}
    \label{tab:supp_noise_params}
    \small
    \setlength{\tabcolsep}{5pt}
    \renewcommand{\arraystretch}{1.05}
    \begin{tabular}{@{}p{0.36\linewidth}p{0.56\linewidth}@{}}
        \toprule
        Parameter & Value \\
        \midrule
        Connectivity / neighborhood definition & Euclidean ball around a random seed voxel (KD-tree radius query), not graph-connected growth \\
        Minimum patch size & radius $2.5$ voxels ($1.25$\,m) \\
        Maximum patch size & radius $7.5$ voxels ($3.75$\,m) \\
        Patch-size distribution & radius $= 5 \times u$, $u\sim\mathcal{U}(0.5, 1.5)$ \\
        Maximum patches per map & $\lfloor N_{\mathrm{target}}/50\rfloor$, $N_{\mathrm{target}}=\text{rate}\times|V|$ \\
        Overlapping patches allowed & Yes; patch centers sampled independently, no overlap exclusion \\
        Rare-class handling & None; target class drawn $\propto$ local class frequency, only seed's own class excluded \\
        Transition-matrix sampling probability & $0.95$ (only for the 10 classes with a confusion-map entry) \\
        Scene-distribution sampling probability & $0.05$ for those 10 classes; $1.0$ for all others \\
        \bottomrule
    \end{tabular}
\end{table}

\begin{figure*}[t]
    \centering

    \begin{tabular}{ccc}
        \textbf{Clean GT} &
        \textbf{Synthetic corruption} &
        \textbf{Corruption mask}
        \\[2mm]

        \includegraphics[width=0.29\textwidth]{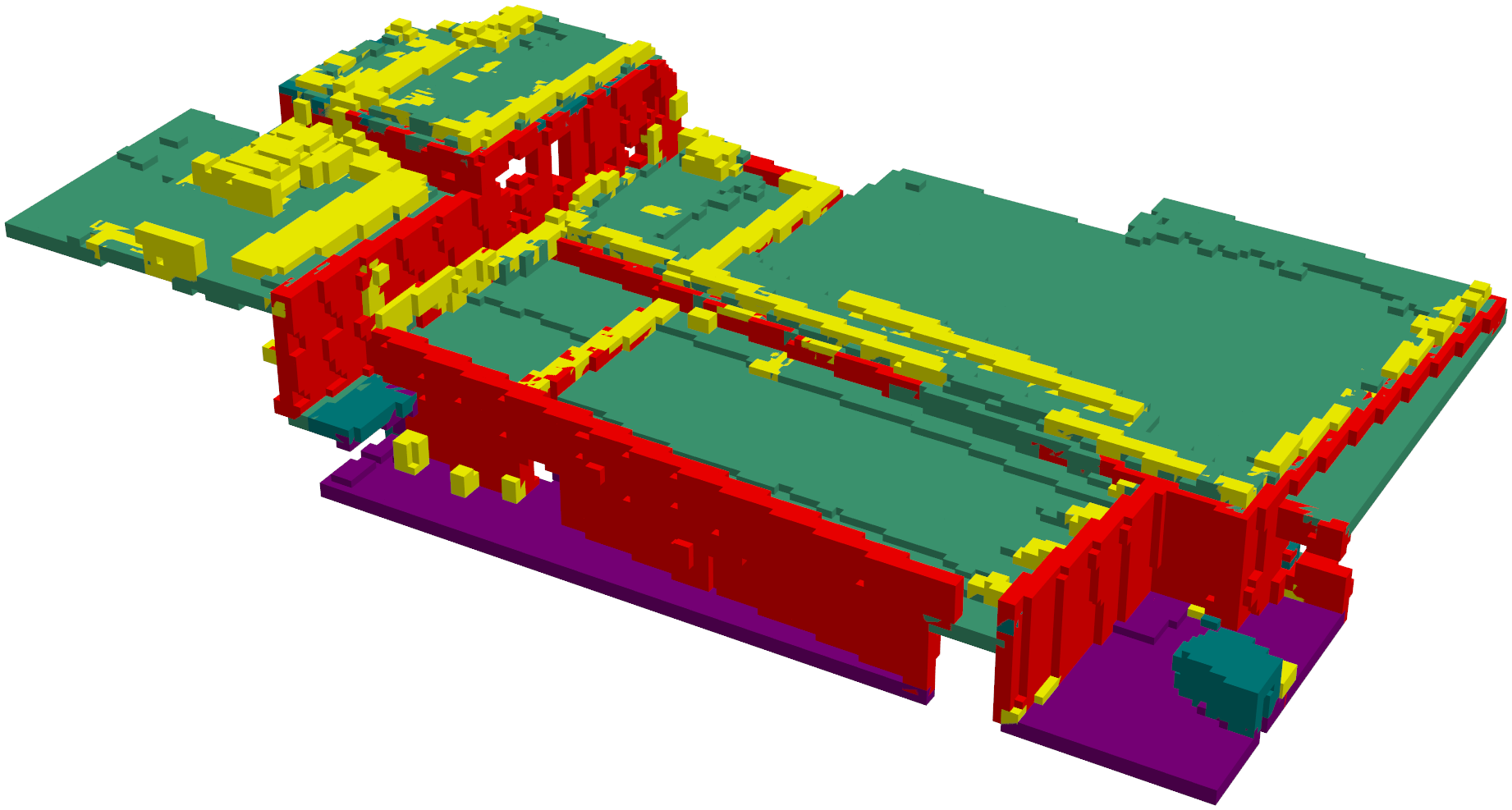} &
        \includegraphics[width=0.29\textwidth]{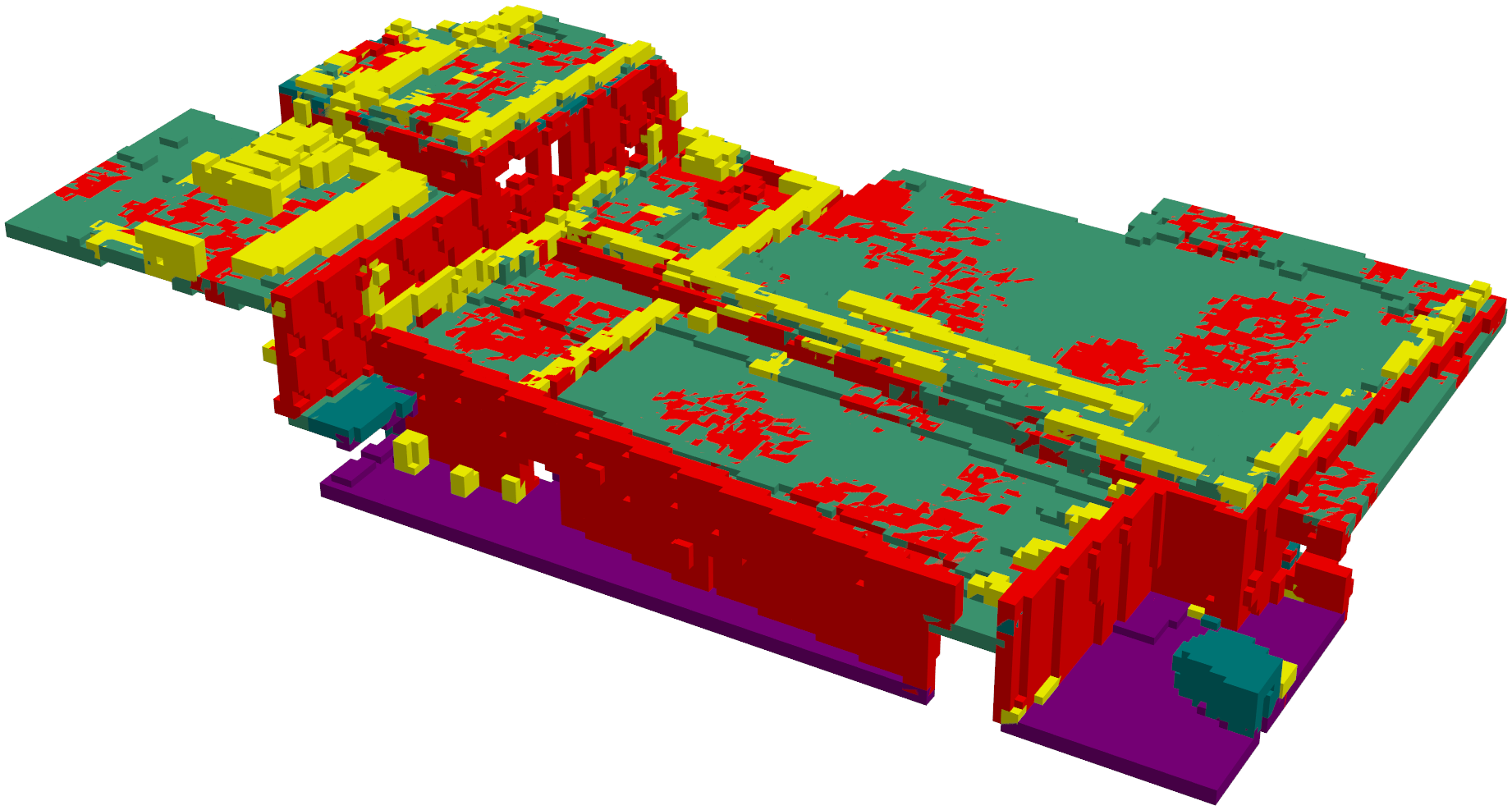} &
        \includegraphics[width=0.29\textwidth]{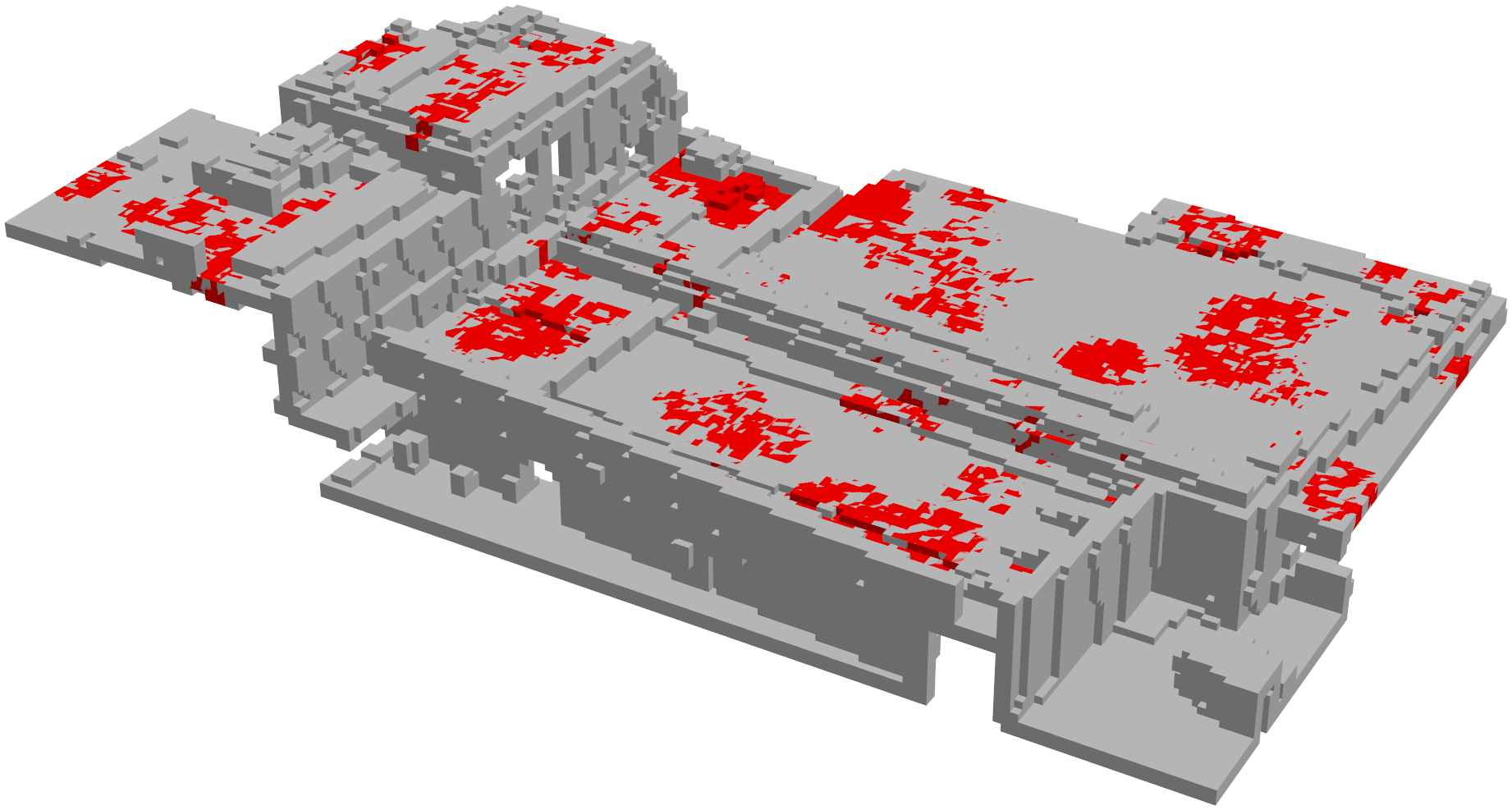} \\
        Roof region & Roof $\rightarrow$ Wall & Corrupted voxels
        \\[4mm]

        \includegraphics[width=0.29\textwidth]{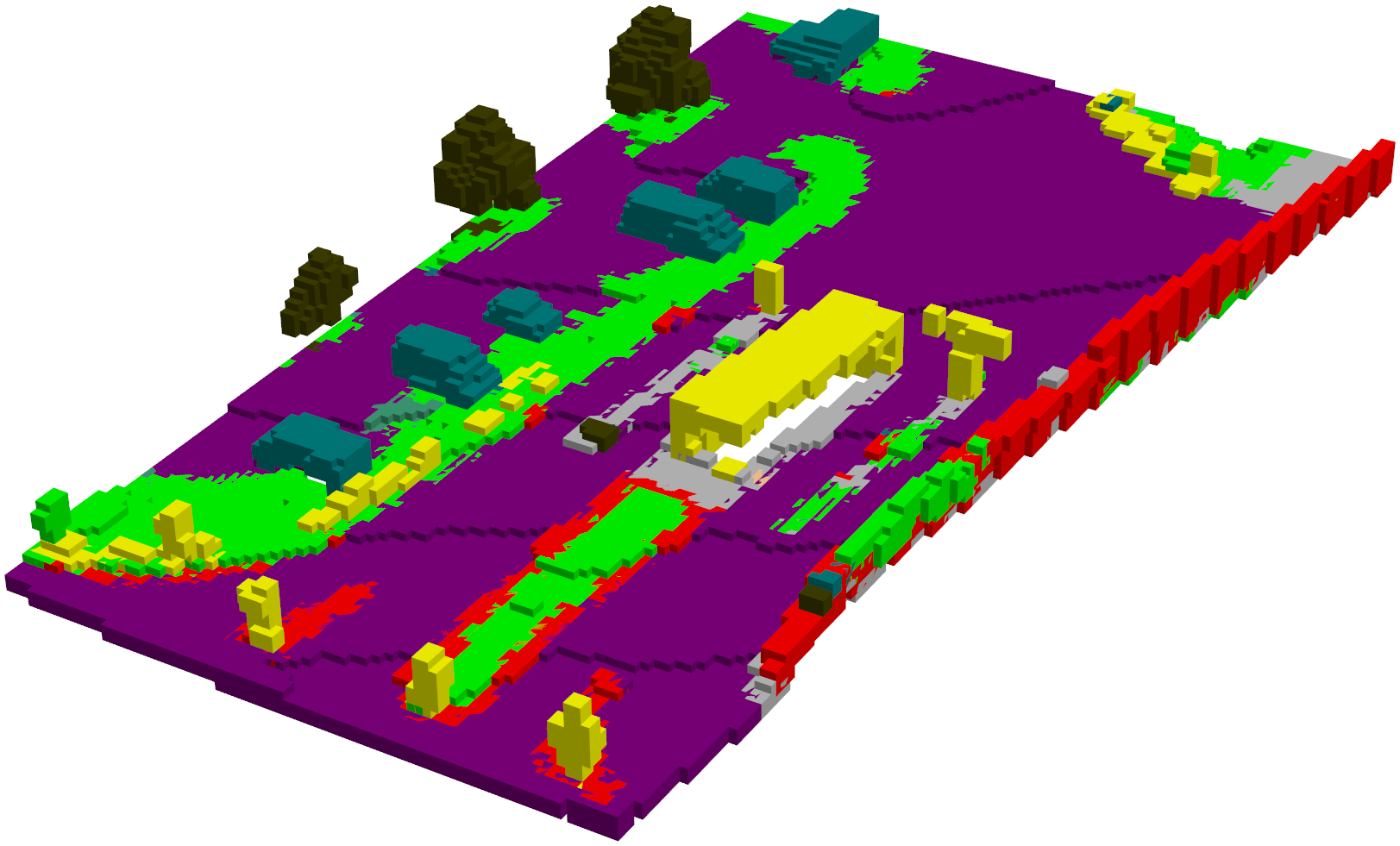} &
        \includegraphics[width=0.29\textwidth]{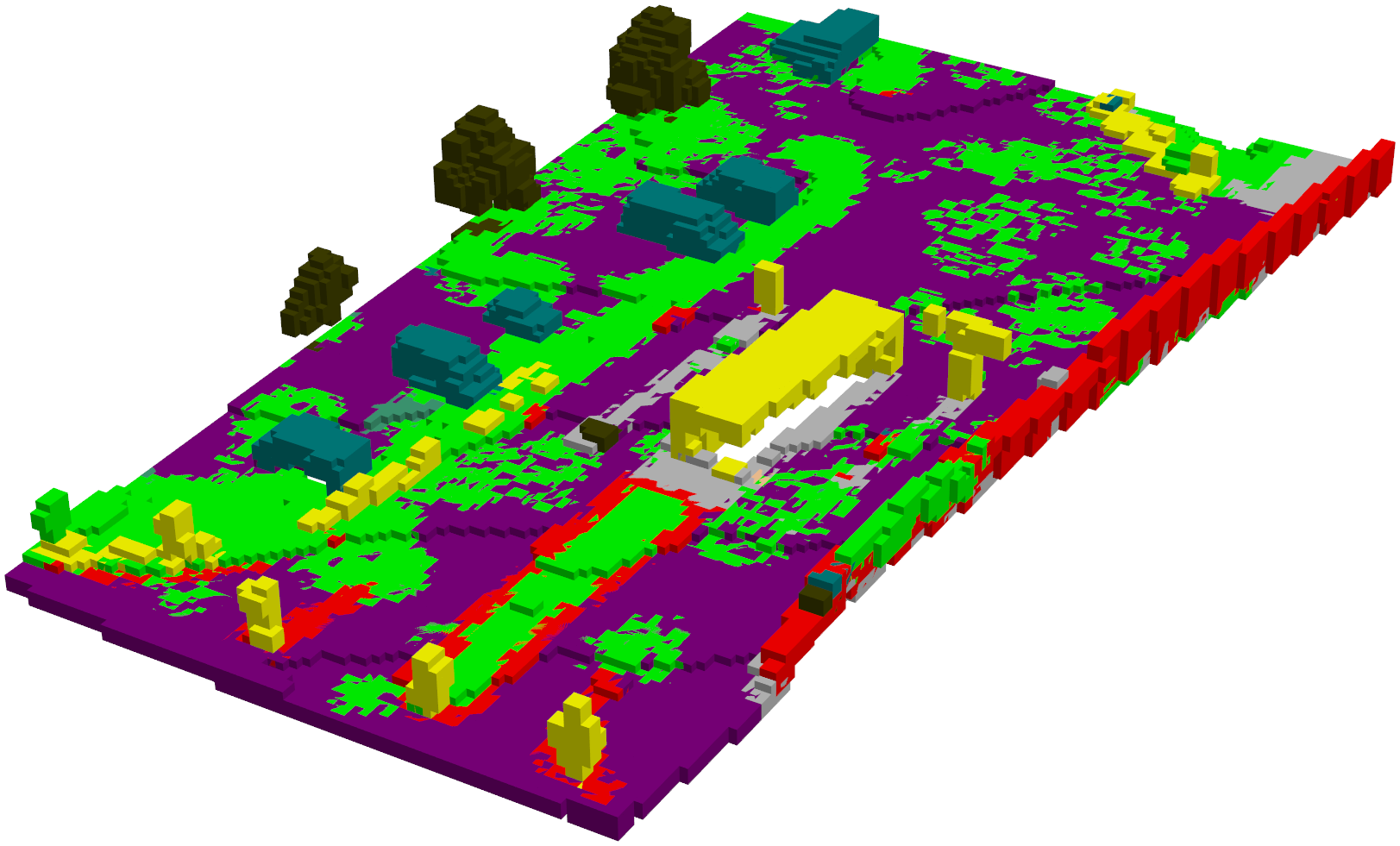} &
        \includegraphics[width=0.29\textwidth]{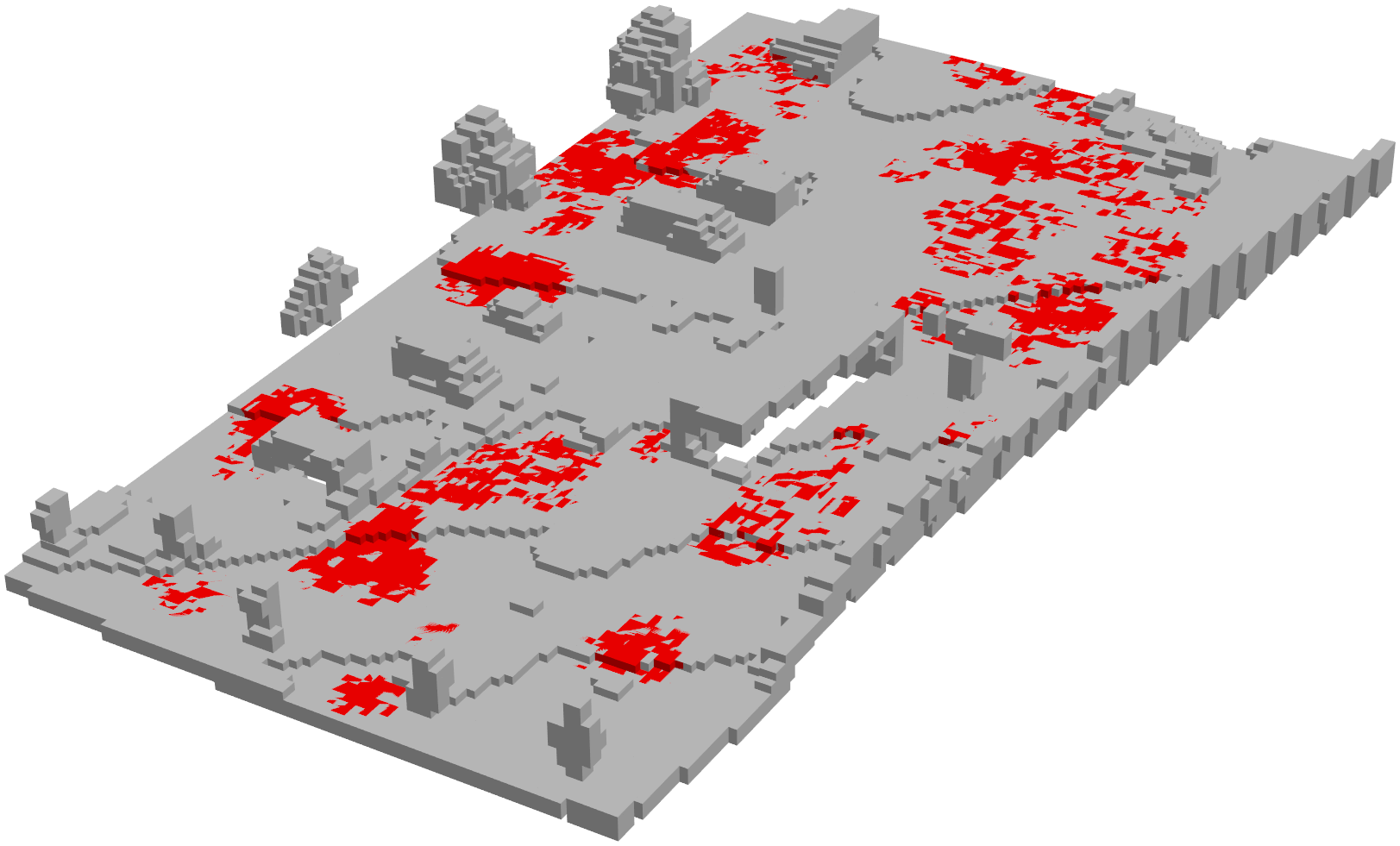} \\
        Road region & Road $\rightarrow$ Grass & Corrupted voxels
        \\[4mm]

        \includegraphics[width=0.29\textwidth]{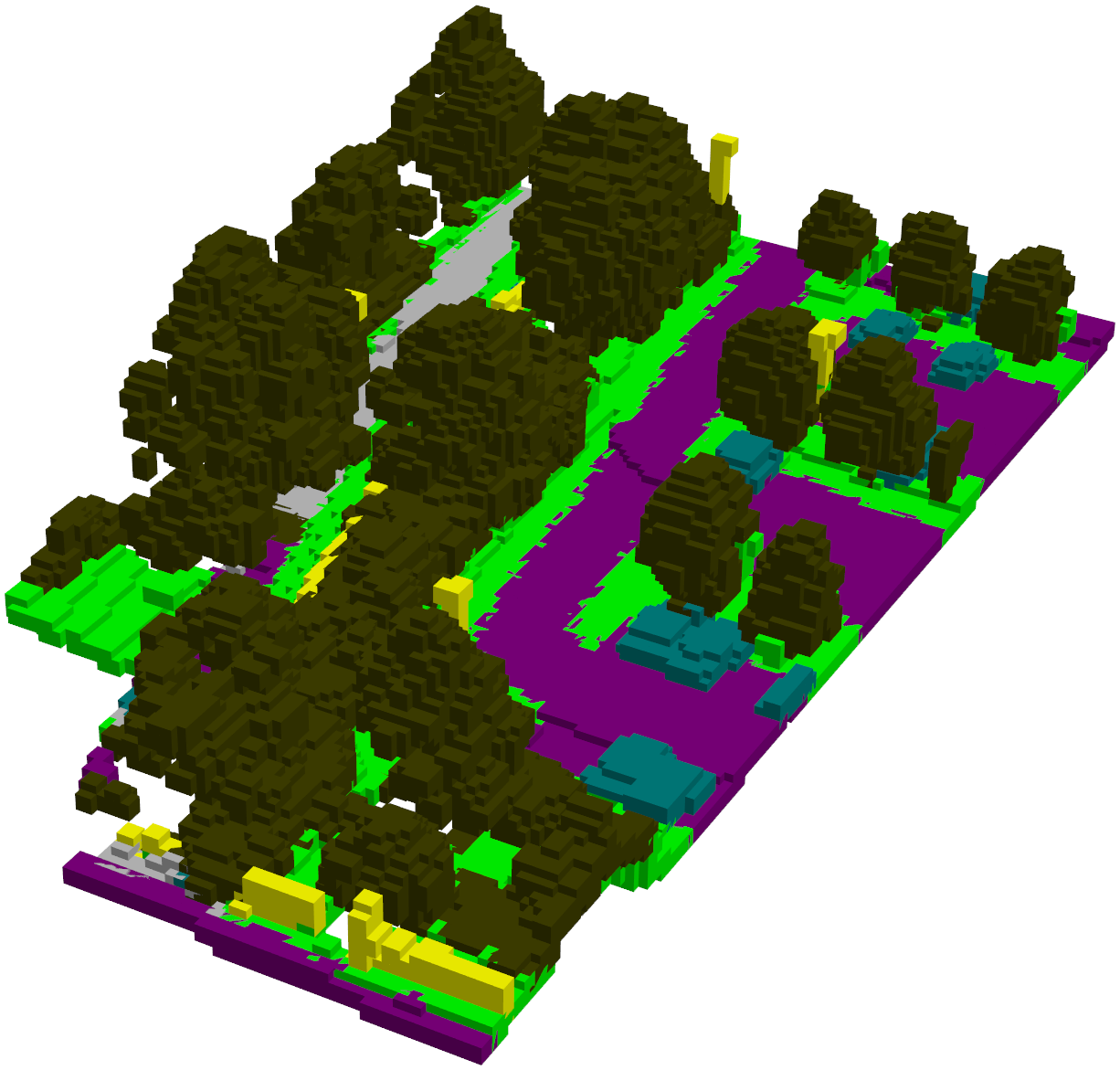} &
        \includegraphics[width=0.29\textwidth]{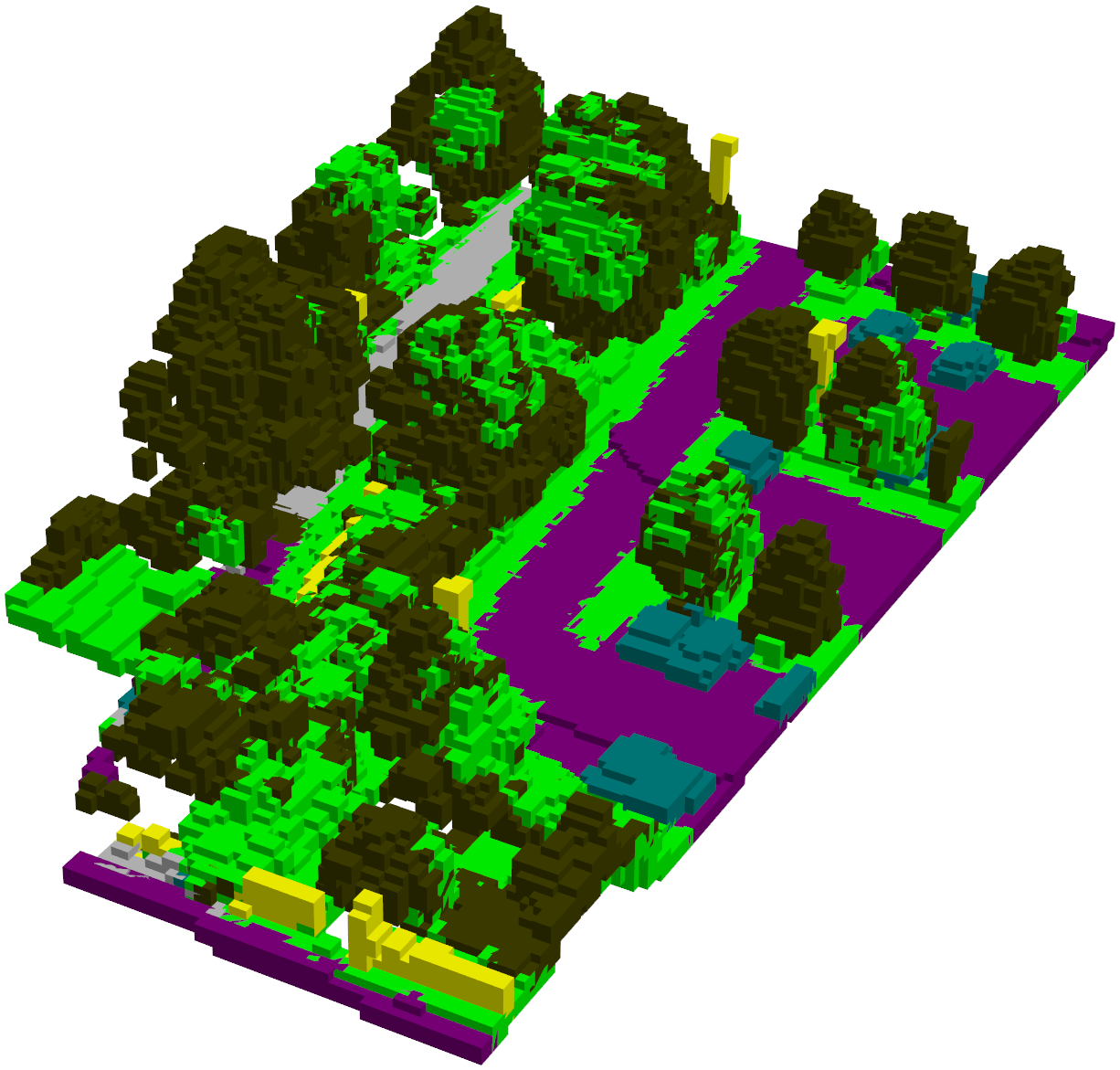} &
        \includegraphics[width=0.29\textwidth]{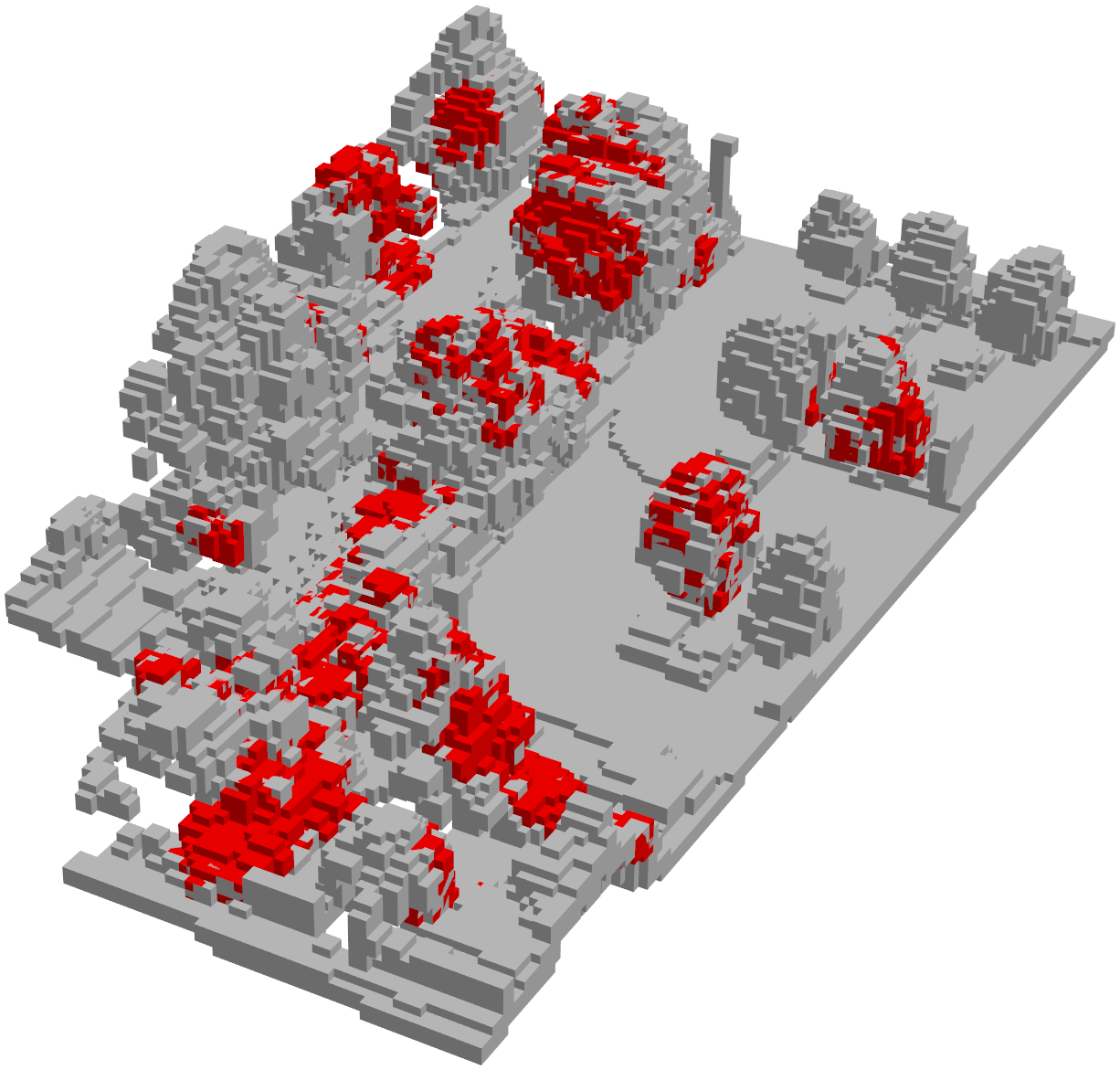} \\
        Tree region & Tree $\rightarrow$ Grass & Corrupted voxels
        \\[2mm]
    \end{tabular}

    \caption{\textbf{Examples of confusion-guided synthetic corruptions used for training.}
    Spatially contiguous voxel regions are selected from clean semantic maps
    and relabeled according to class-transition probabilities derived from
    voxel-level upstream confusion statistics. Unlike independent label
    flipping, the resulting corruptions form coherent erroneous regions
    resembling those observed in completed upstream maps.}
    \label{fig:supp_synthetic_noise}
\end{figure*}

\subsection{Corruption Rate Schedule}
\label{sec:supp_corruption_schedule}

The corruption rate is annealed linearly over training,
\begin{equation}
\rho(t) = \rho_{\max} - (\rho_{\max} - \rho_{\min})\,\frac{t}{T_{max}},
\label{eq:corruption_schedule}
\end{equation}
with $\rho_{\max} = 0.50$ and $\rho_{\min} = 0.05$. The schedule imposes a
curriculum on the correction task. In the early epochs half of the label map is
corrupted. These errors are extensive and inconsistent with their surroundings,
so the correction signal is unambiguous. The network acquires the coarse
structure of the task before it is required to resolve finer distinctions. The
corrupted fraction then decreases steadily, and in the final epochs only five
percent of the map is affected.

The end rate is deliberately non-zero. A schedule decaying to zero would present
the later epochs with an essentially clean label map. The correction objective
would then exert little pressure on the parameters, and the identity mapping
would become an increasingly favorable solution. A floor of five percent retains
a correction signal for the whole of training.

$\rho(t)$ is the fraction of the map's voxels targeted for corruption at epoch
$t$, and it fixes the corruption budget,
\begin{equation}
N_{\mathrm{corrupt}} = \left\lceil \rho(t) \times N_{\mathrm{voxels}} \right\rceil .
\label{eq:corruption_budget}
\end{equation}
This budget is not applied to individual voxels but is spent on contiguous patches, each relabeled to a single incorrect class. A patch covers a ball whose extent depends on the sampled radius and on the local voxel density, and it relabels between thirty and ninety percent of the voxels that ball contains. \cref{tab:supp_noise_params} lists the complete set of corruption parameters. Regional corruption is adopted because the errors produced by the upstream model
are themselves spatially correlated. A 2D segmentation backbone does not
misclassify scattered voxels independently, but mislabels entire surfaces, such
as a roof plane assigned to a wall or a strip of verge assigned to a road. Corrupting the map in coherent regions therefore better reflects the spatial structure of the errors the corrector is required to address at inference, whereas independent per-voxel noise would be largely removable by local smoothing and would constitute a substantially weaker training signal.

Corruption is resampled at every epoch, so no two epochs present the same
corrupted map. Since generating these maps is computationally non-trivial, the
labels for epoch $t+1$ are precomputed in a background thread while epoch $t$
trains. This removes the stalls that would otherwise occur at each epoch
boundary.

\subsection{Synthetic Corruption Examples}

\cref{fig:supp_synthetic_noise} shows representative corruptions produced by
the procedure described above, using three substitutions that occur frequently in the upstream confusion statistics: roof-to-wall, road-to-grass, and tree-to-grass. Each row shows the clean ground-truth labels, the corrupted map used as network input, and the binary error mask supervising the detection head. The corrupted regions span coherent parts of a surface rather than scattered individual voxels, which is the behavior the correction model is trained to reverse and which local smoothing cannot resolve when an entire region carries the same incorrect label.

\begin{table*}[t]
\centering
\caption{\textbf{Semantic class coverage across the datasets used to train the
upstream segmentation models.} The unlabeled class is mapped to the ignore
index and is not part of the 26-class taxonomy.}
\label{tab:upstream_data_class_mapping}
\resizebox{\textwidth}{!}{%
\begin{tabular}{l *{27}{c}}
\toprule
\textbf{Dataset} & 
\rotatebox{90}{unlabeled} & \rotatebox{90}{road} & \rotatebox{90}{dirt} & \rotatebox{90}{gravel} & 
\rotatebox{90}{rock} & \rotatebox{90}{grass} & \rotatebox{90}{vegetation} & \rotatebox{90}{tree} & 
\rotatebox{90}{ground-obstacle} & \rotatebox{90}{animals} & \rotatebox{90}{person} & \rotatebox{90}{bicycle} & 
\rotatebox{90}{vehicle} & \rotatebox{90}{water} & \rotatebox{90}{boat} & \rotatebox{90}{building} & 
\rotatebox{90}{roof} & \rotatebox{90}{sky} & \rotatebox{90}{drone} & \rotatebox{90}{train-track} & 
\rotatebox{90}{cable} & \rotatebox{90}{cable-tower} & \rotatebox{90}{wind-turbine-blade} & 
\rotatebox{90}{wind-turbine-tower} & \rotatebox{90}{walkway} & \rotatebox{90}{bridge} & \rotatebox{90}{airplanes} \\
\midrule
CART~\cite{Lee2024CARTCA} & $\checkmark$ & $\checkmark$ & $\checkmark$ & & $\checkmark$ & & $\checkmark$ & $\checkmark$ & & & $\checkmark$ & & $\checkmark$ & $\checkmark$ & & $\checkmark$ & & $\checkmark$ & & & & & & & & & \\
ICG~\cite{icg_drone_dataset} & $\checkmark$ & $\checkmark$ & $\checkmark$ & $\checkmark$ & $\checkmark$ & $\checkmark$ & $\checkmark$ & $\checkmark$ & $\checkmark$ & $\checkmark$ & $\checkmark$ & $\checkmark$ & $\checkmark$ & $\checkmark$ & & $\checkmark$ & $\checkmark$ & & & & & & & & & & \\
OkutSwiss~\cite{okutswiss_dataset} & $\checkmark$ & $\checkmark$ & & & & & $\checkmark$ & & $\checkmark$ & & $\checkmark$ & & $\checkmark$ & $\checkmark$ & & $\checkmark$ & & & & $\checkmark$ & & & & & & & \\
UAVid~\cite{DBLP:journals/corr/abs-1810-10438} & $\checkmark$ & $\checkmark$ & & & & & $\checkmark$ & $\checkmark$ & & & $\checkmark$ & & $\checkmark$ & $\checkmark$ & & $\checkmark$ & $\checkmark$ & & & & & & & & & & \\
UDD~\cite{chen2018large} & $\checkmark$ & $\checkmark$ & & & & & $\checkmark$ & & & & & & $\checkmark$ & $\checkmark$ & & $\checkmark$ & $\checkmark$ & & & & & & & & & & \\
VDD~\cite{cai2025vdd} & $\checkmark$ & $\checkmark$ & & & & & $\checkmark$ & & & & & & $\checkmark$ & $\checkmark$ & & $\checkmark$ & $\checkmark$ & & & & & & & & & & \\
Incenda~\cite{incenda_ai} & $\checkmark$ & $\checkmark$ &$\checkmark$ & & $\checkmark$& $\checkmark$ & $\checkmark$ & $\checkmark$ & $\checkmark$ & & $\checkmark$ & & $\checkmark$ & $\checkmark$ & $\checkmark$ & $\checkmark$ & $\checkmark$ & $\checkmark$ & $\checkmark$ & & $\checkmark$ & $\checkmark$ & & & $\checkmark$ & $\checkmark$ & $\checkmark$ \\
\midrule
FPV-A & $\checkmark$ & $\checkmark$ & $\checkmark$ & $\checkmark$ & $\checkmark$ & $\checkmark$ & $\checkmark$ & $\checkmark$ & $\checkmark$ & $\checkmark$ & $\checkmark$ & $\checkmark$ & $\checkmark$ & $\checkmark$ & $\checkmark$ & $\checkmark$ & $\checkmark$ & $\checkmark$ & & $\checkmark$ & $\checkmark$ & $\checkmark$ & & $\checkmark$ & $\checkmark$ & $\checkmark$ & $\checkmark$ \\
FPV-B & $\checkmark$ & $\checkmark$ & $\checkmark$ & $\checkmark$ & $\checkmark$ & $\checkmark$ & $\checkmark$ & $\checkmark$ & $\checkmark$ & $\checkmark$ & $\checkmark$ & $\checkmark$ & $\checkmark$ & $\checkmark$ & $\checkmark$ & $\checkmark$ & $\checkmark$ & $\checkmark$ & $\checkmark$ & $\checkmark$ & $\checkmark$ & $\checkmark$ & & $\checkmark$ & $\checkmark$ & $\checkmark$ & $\checkmark$ \\
\bottomrule
\end{tabular}
}
\end{table*}

\section{Upstream Semantic Segmentation}
\label{sec:supp_upstream}

This section provides additional details on the upstream semantic segmentation models, including their training data, training configuration, and performance. We further describe the semantic class mapping used to align the upstream predictions with the OccuFly taxonomy and the 12 classes considered by \method.

\subsection{Training Data}

The upstream segmentation models are trained on a combined corpus of nine
aerial datasets. Seven are public datasets: ICG~\cite{icg_drone_dataset}, OkutSwiss~\cite{okutswiss_dataset}, Incenda~\cite{incenda_ai}, UAVid~\cite{DBLP:journals/corr/abs-1810-10438}, UDD~\cite{chen2018large},
VDD~\cite{cai2025vdd}, and CART~\cite{Lee2024CARTCA}. The remaining two are internally curated datasets derived from publicly available FPV aerial videos (FPV-A and FPV-B). The two curated datasets contain 2,050 images in total. Across all datasets, the combined training and validation splits contain 5,191 and 553 images, respectively.

We adopt a unified 26-class semantic taxonomy across all datasets. Dataset-specific annotations are remapped to this common taxonomy to account for differences in class granularity and semantic definitions. Labels that cannot be mapped unambiguously, as well as unlabeled or background regions, are assigned to the ignore index and excluded from supervision. The resulting semantic class coverage is summarized in \cref{tab:upstream_data_class_mapping}.

\subsection{Training Configuration}

We use four upstream segmentation models based on three architectures. All feature extractors are initialized from pretrained weights. The models are implemented using the Hugging Face transformers library and were trained on a single NVIDIA A100 40GB GPU.

For optimization, we use AdamW with a weight decay of $0.05$ and an initial
learning rate of $5\times10^{-5}$ following a cosine annealing schedule.
The models are trained for 115 epochs with a batch size of 4 and an input
resolution of $640\times640$. For each model, we retain the checkpoint with
the highest validation mIoU. Training employs stochastic geometric, photometric, and image-degradation augmentations. These include affine scaling, rotation, perspective warping, random resized crops, color jitter, CLAHE, contrast variation, Gaussian noise, sharpening, compression, median blur, and motion blur. Each augmentation block is activated with probability $0.6$, and one transformation is sampled from the activated block.

SegFormer and UPerNet are optimized using cross-entropy together with a
context-aware leakage penalty (CALP), which we introduce here purely as part of the upstream training recipe,

\begin{equation}
\mathcal{L}_{\mathrm{tot}}
=
\mathcal{L}_{\mathrm{CE}}
+
\lambda \mathcal{L}_{\mathrm{CALP}},
\end{equation}
where $\lambda=0.25$. CALP penalizes predictions of classes present in the annotated portion of an image within regions assigned to the ignore index, reducing semantic leakage into regions without explicit supervision. The penalty is applied only when ignored pixels constitute at least $2\%$ of the image. Mask2Former is trained without the CALP component.

\subsection{Results}

We evaluate the four upstream segmentation models on both the validation split of the training data and an unseen aerial dataset containing 221 images. The latter is not used during training and provides an additional measure of generalization to unseen environments. We report mean Intersection over Union (mIoU) for both settings in \cref{tab:model_generalization}.

\begin{table}[t]
    \centering
    \caption{\textbf{Performance of the upstream segmentation models on the validation and unseen datasets.}}
    \label{tab:model_generalization}
    \small
    \setlength{\tabcolsep}{5pt}
    \renewcommand{\arraystretch}{1.05}
    \begin{tabular}{llcc}
        \toprule
        \textbf{Method} & \textbf{Backbone} &
        \multicolumn{2}{c}{\textbf{mIoU}} \\
        \cmidrule(lr){3-4}
        & & \textbf{Val} & \textbf{Unseen} \\
        \midrule
        SegFormer~\cite{xie2021segformer}
            & MiT-B3        & 66.56 & \textbf{38.91} \\
        UPerNet~\cite{xiao2018unified}
            & Swin-v2-T     & 66.53 & 38.11 \\
        UPerNet~\cite{xiao2018unified}
            & ConvNeXt-v2-T & 69.23 & 38.05 \\
        Mask2Former~\cite{cheng2022masked}
            & Swin-T & \textbf{69.55} & 36.60 \\
        \bottomrule
    \end{tabular}
\end{table}

\subsection{Semantic Class Mapping}
The upstream predictions pass through two remapping stages before
correction. The 26-class upstream taxonomy is first aligned with the
21-class OccuFly taxonomy, where most classes correspond one-to-one.

The OccuFly labels are then consolidated into the 12 classes used for map
correction by merging categories that are semantically or geometrically
close: walkway and parking lot into road; truck into vehicle; vegetation
into grass; building and constructions into wall; ground obstacle, cranes,
cable, and cable tower into obstacle; and rock into gravel. Tree, roof,
person, dirt, bicycle, and water are unchanged. \cref{tab:supp_class_mapping}
lists both stages in full.

\begin{table*}[t]
\centering
\caption{\textbf{Semantic taxonomy alignment from upstream segmentation to VoxelFix.}
The 26-class upstream predictions are first aligned with the OccuFly taxonomy
and subsequently reduced to the 12 classes used for VoxelFix training and
evaluation. Classes without a corresponding target class are ignored.}
\label{tab:supp_class_mapping}

\small
\setlength{\tabcolsep}{10pt}
\renewcommand{\arraystretch}{1.08}

\begin{minipage}[t]{0.24\textwidth}
\centering
\textbf{Upstream}\\
\footnotesize
\begin{tabular}{rl}
\toprule
ID & Class \\
\midrule
 1 & Road \\
 2 & Dirt \\
 3 & Gravel \\
 4 & Rock \\
 5 & Grass \\
 6 & Vegetation \\
 7 & Tree \\
 8 & Ground obstacle \\
 9 & Animal \\
10 & Person \\
11 & Bicycle \\
12 & Vehicle \\
13 & Water \\
14 & Boat \\
15 & Building \\
16 & Roof \\
17 & Sky \\
18 & Drone \\
19 & Train track \\
20 & Cable \\
21 & Cable tower \\
22 & Wind-turbine blade \\
23 & Wind-turbine tower \\
24 & Walkway \\
25 & Bridge \\
26 & Airplane \\
\bottomrule
\end{tabular}
\end{minipage}
\hfill
$\boldsymbol{\rightarrow}$
\hfill
\begin{minipage}[t]{0.24\textwidth}
\centering
\textbf{OccuFly}\\
\footnotesize
\begin{tabular}{l}
\toprule
Class \\
\midrule
Road \\
Walkway \\
Dirt \\
Gravel \\
Rock \\
Grass \\
Vegetation \\
Tree \\
Ground obstacle \\
Person \\
Bicycle \\
Vehicle \\
Water \\
Building \\
Roof \\
Cables \\
Cable tower \\
Parking lot \\
Constructions \\
Cranes \\
Truck \\
\bottomrule
\end{tabular}
\end{minipage}
\hfill
$\boldsymbol{\rightarrow}$
\hfill
\begin{minipage}[t]{0.40\textwidth}
\centering
\textbf{VoxelFix}\\
\footnotesize

\begin{tabular}{cll}
\toprule
& Class & Merged from \\
\midrule
\textcolor[RGB]{128,0,128}{\rule{1.3ex}{1.3ex}}
    & Road & Road, Walkway, Parking lot \\
\textcolor[RGB]{64,64,0}{\rule{1.3ex}{1.3ex}}
    & Tree & Tree \\
\textcolor[RGB]{0,128,128}{\rule{1.3ex}{1.3ex}}
    & Vehicle & 	Vehicle, Truck \\
\textcolor[RGB]{0,255,0}{\rule{1.3ex}{1.3ex}}
    & Grass & Grass, Vegetation \\
\textcolor[RGB]{255,0,0}{\rule{1.3ex}{1.3ex}}
    & Wall & Building, Constructions \\
\textcolor[RGB]{64,160,120}{\rule{1.3ex}{1.3ex}}
    & Roof & Roof \\
\textcolor[RGB]{255,255,0}{\rule{1.3ex}{1.3ex}}
    & Obstacle &
    \begin{tabular}[c]{@{}l@{}}
        Ground obstacle, Cranes,\\
        Cables, Cable tower
    \end{tabular} \\
\textcolor[RGB]{192,192,192}{\rule{1.3ex}{1.3ex}}
    & Gravel & Gravel, Rock \\
\textcolor[RGB]{255,16,255}{\rule{1.3ex}{1.3ex}}
    & Person & Person \\
\textcolor[RGB]{128,0,0}{\rule{1.3ex}{1.3ex}}
    & Dirt & Dirt \\
\textcolor[RGB]{255,204,153}{\rule{1.3ex}{1.3ex}}
    & Bicycle & Bicycle \\
\textcolor[RGB]{0,0,255}{\rule{1.3ex}{1.3ex}}
    & Water & Water \\
\bottomrule
\end{tabular}

\vspace{6mm}

\includegraphics[width=0.95\linewidth]{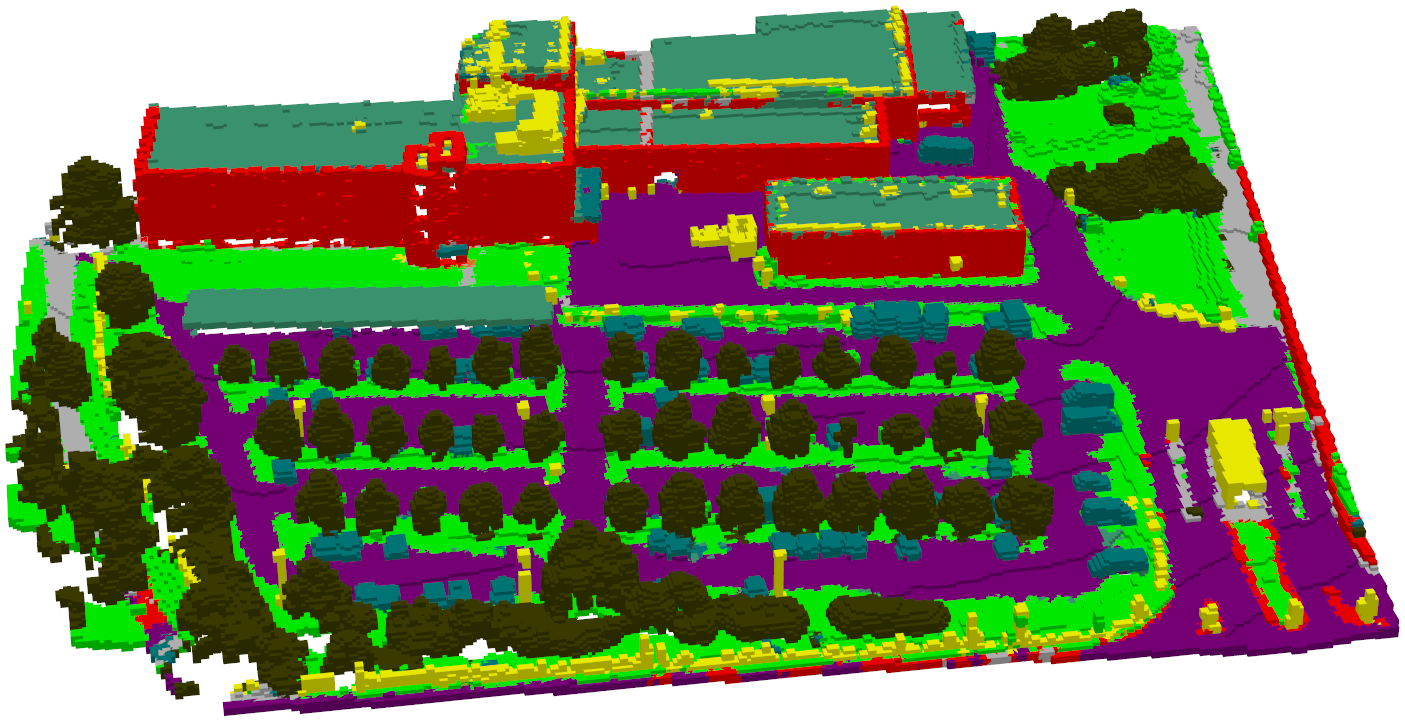}

\vspace{0.5mm}
{\scriptsize
\textit{Example semantic scene using the final 12-class VoxelFix taxonomy.}
}

\end{minipage}
\end{table*}

\section{Semantic Voxel Map Generation}
\label{sec:semantic_voxel_map_gen}

The completed maps that VoxelFix operates on are produced by a standard
modular mapping pipeline. Aerial images are undistorted using the camera
intrinsics and distortion coefficients from the OccuFly photogrammetry
calibration, and each undistorted frame is passed through one of the four
upstream segmentation models. Per-pixel predictions are remapped to the
12-class correction taxonomy and back-projected into 3D using the
ground-truth per-frame depth maps and camera poses provided by OccuFly,
yielding a semantically labeled point cloud per frame.

These per-frame clouds are aggregated with Radix, the semantic occupancy grid
mapping component \cite{scherer2026socc}, operating in its semantic mode at a voxel size of $0.5$\,m. Each occupied voxel maintains a class distribution updated by an exponential moving average of the observed semantic label frequencies, and its final label is the maximum-probability class. We use the default configuration throughout, with the single exception that ray-based cleaning (\texttt{cleaning\_ray}) is disabled: free-space carving is intended for online mapping with dynamic objects, whereas our scenes are mostly static and reconstructed offline, so retaining all accumulated observations gives a more complete map.

After construction, the class distributions are discarded, and VoxelFix
receives only the hard labels and the fixed voxel geometry. The CRF baseline
is the one exception, additionally consuming the retained per-voxel class
probabilities and therefore operating with strictly more information than
VoxelFix.

\section{Baseline Implementation Details}
\label{sec:supp_baselines}

All baselines operate under the same map-only constraint as VoxelFix, with
the exception of the CRF, which additionally receives the per-voxel class
probabilities retained during map construction. Parameters were fixed a
priori and applied unchanged across all upstream models and scenes.

\subsection{KNN Label Smoothing}

KNN is the simplest map-only correction: each voxel adopts the majority label
among its nearest spatial neighbors, with no learned component and no
geometric features. It represents the class of purely local smoothing that
our formulation is designed to improve on, and its behavior on spatially
coherent errors, where the neighborhood already agrees on the wrong label,
motivates the remaining comparisons. Parameters are listed in
\cref{tab:supp_knn}.

\begin{table}[t]
    \centering
    \caption{\textbf{KNN baseline parameters.}}
    \label{tab:supp_knn}
    \small
    \begin{tabular}{@{}p{0.36\linewidth}p{0.56\linewidth}@{}}
        \toprule
        Parameter & Value \\
        \midrule
        Number of neighbors & 5 \\
        Distance metric & Euclidean, in metres \\
        Include query voxel & Yes (counted within $k=5$) \\
        Distance weighting & None (uniform vote) \\
        Number of iterations & 1 \\
        Tie-breaking rule & Keep the voxel's current label \\
        \bottomrule
    \end{tabular}
\end{table}

\subsection{CRF Label Smoothing}

The CRF adds probabilistic regularization on top of local smoothing, combining
a unary term derived from the upstream class probabilities with a dense
pairwise term using Gaussian spatial weighting and a Potts compatibility
function, solved by mean-field inference. Unlike every other method compared
here, it consumes the class distributions retained by Radix rather than the
hard labels alone, and therefore operates on strictly richer input than
VoxelFix. Parameters are listed in \cref{tab:supp_crf}.

\begingroup
\setlength{\tabcolsep}{3pt}
\begin{table}[t]
    \centering
    \caption{\textbf{CRF baseline parameters.}}
    \label{tab:supp_crf}
    \footnotesize
    \renewcommand{\arraystretch}{0.92}
    \begin{tabular}{@{}p{0.32\linewidth}p{0.62\linewidth}@{}}
        \toprule
        Parameter & Value \\
        \midrule
        Unary potential &
        Upstream scalar confidence on the input label, remaining mass uniform
        over the other 11 classes \\
        Pairwise \newline neighborhood $k$ &
        30 (self excluded) \\
        Spatial standard \newline deviation $\sigma$ &
        $1.5$\,m \\
        Edge weighting &
        Gaussian, $\exp(-d^2/2\sigma^2)$, row-normalized \\
        Pairwise weight (Potts) &
        $15.0$ \\
        Compatibility term &
        Potts: class-independent, uniform penalty for unlike labels \\
        Number of iterations &
        20 (mean-field) \\
        Parameter selection &
        None; fixed constants across all scenes and upstream models \\
        \bottomrule
    \end{tabular}
\end{table}
\endgroup

\subsection{Geometry-Based Heuristic}

This baseline tests how much of the correction is achievable from structural
priors alone. Cloth Simulation Filtering separates ground from above-ground
voxels, after which the two partitions are corrected under different rules:
KNN majority voting restricted to valid ground classes, and DBSCAN clustering
followed by per-cluster majority relabeling for above-ground objects. The
partition prevents ground labels from propagating into elevated structures
and vice versa, which is why it outperforms unrestricted smoothing, but it
cannot resolve confusions that geometry does not determine. Parameters are
listed in \cref{tab:supp_geometry_baseline}.

\begingroup
\setlength{\tabcolsep}{3pt}
\begin{table}[t]
    \centering
    \caption{\textbf{Geometry-based heuristic parameters.}}
    \label{tab:supp_geometry_baseline}
    \footnotesize
    \renewcommand{\arraystretch}{0.92}
    \begin{tabular}{@{}p{0.32\linewidth}p{0.62\linewidth}@{}}
        \toprule
        Parameter & Value \\
        \midrule
        Ground separation \newline (CSF) &
        Cloth resolution $1.0$, rigidness $5$, classification threshold
        $0.3$\,m, 2500 iterations \\
        Ground correction &
        KNN majority vote over valid ground classes; $k=50$, widened to
        $k=100$ when too few valid neighbors are found \\
        Above-ground \newline clustering &
        DBSCAN, $\epsilon=0.5$\,m, minimum 5 samples \\
        Cluster relabeling &
        Majority vote within each cluster; dense components merged into a
        single cluster \\
        \bottomrule
    \end{tabular}
\end{table}
\endgroup


\begin{figure*}[t]
    \centering
    \footnotesize
    \setlength{\tabcolsep}{3pt}

    \begin{tabular}{@{}
        >{\centering\arraybackslash}m{0.075\textwidth}
        c c c
    @{}}
        &
        \textbf{Radix Input} &
        \textbf{VoxelFix} &
        \textbf{GT} \\[1mm]

        \textbf{Scene 1} &
        \includegraphics[width=0.24\textwidth]{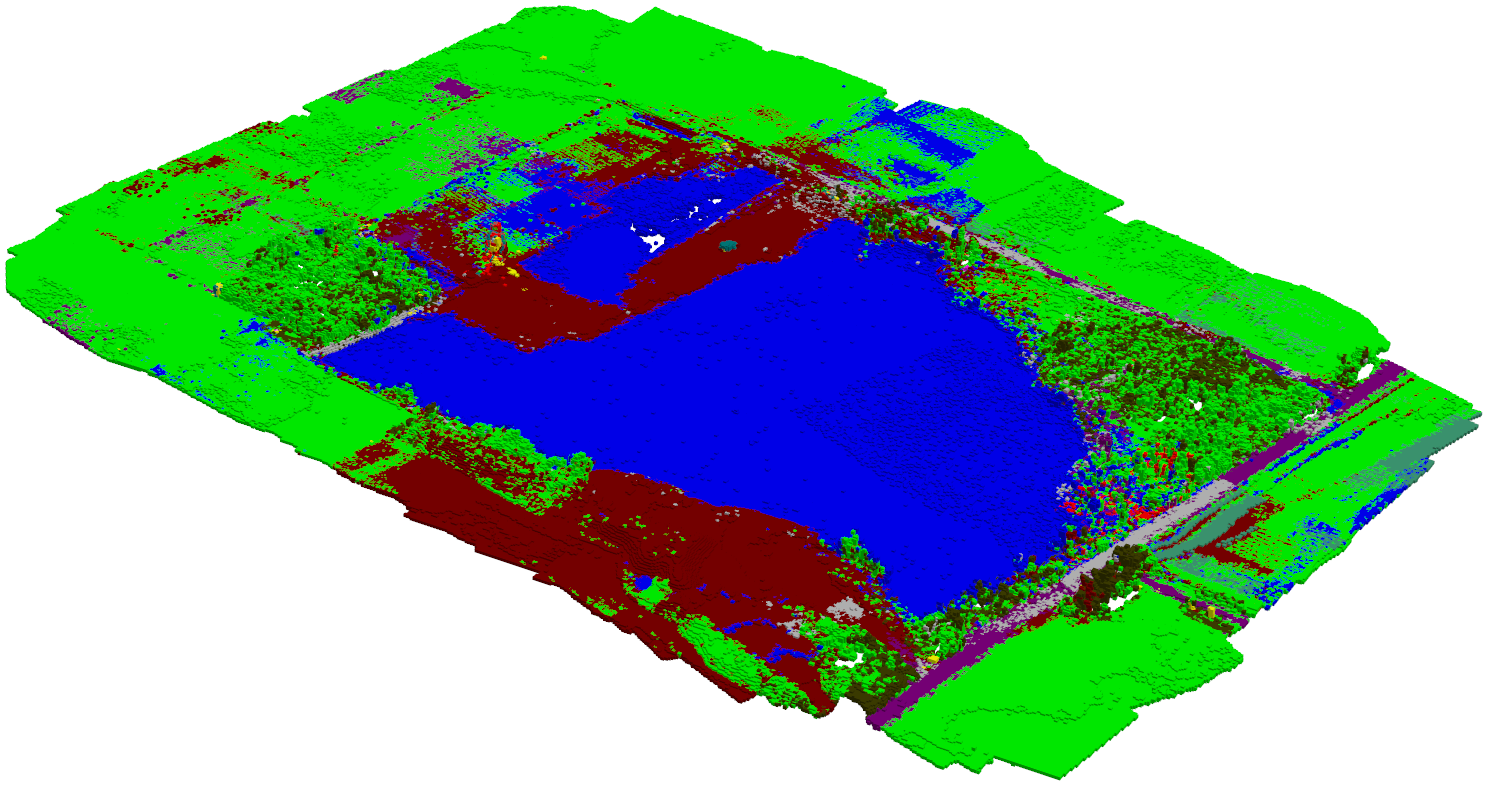} &
        \includegraphics[width=0.24\textwidth]{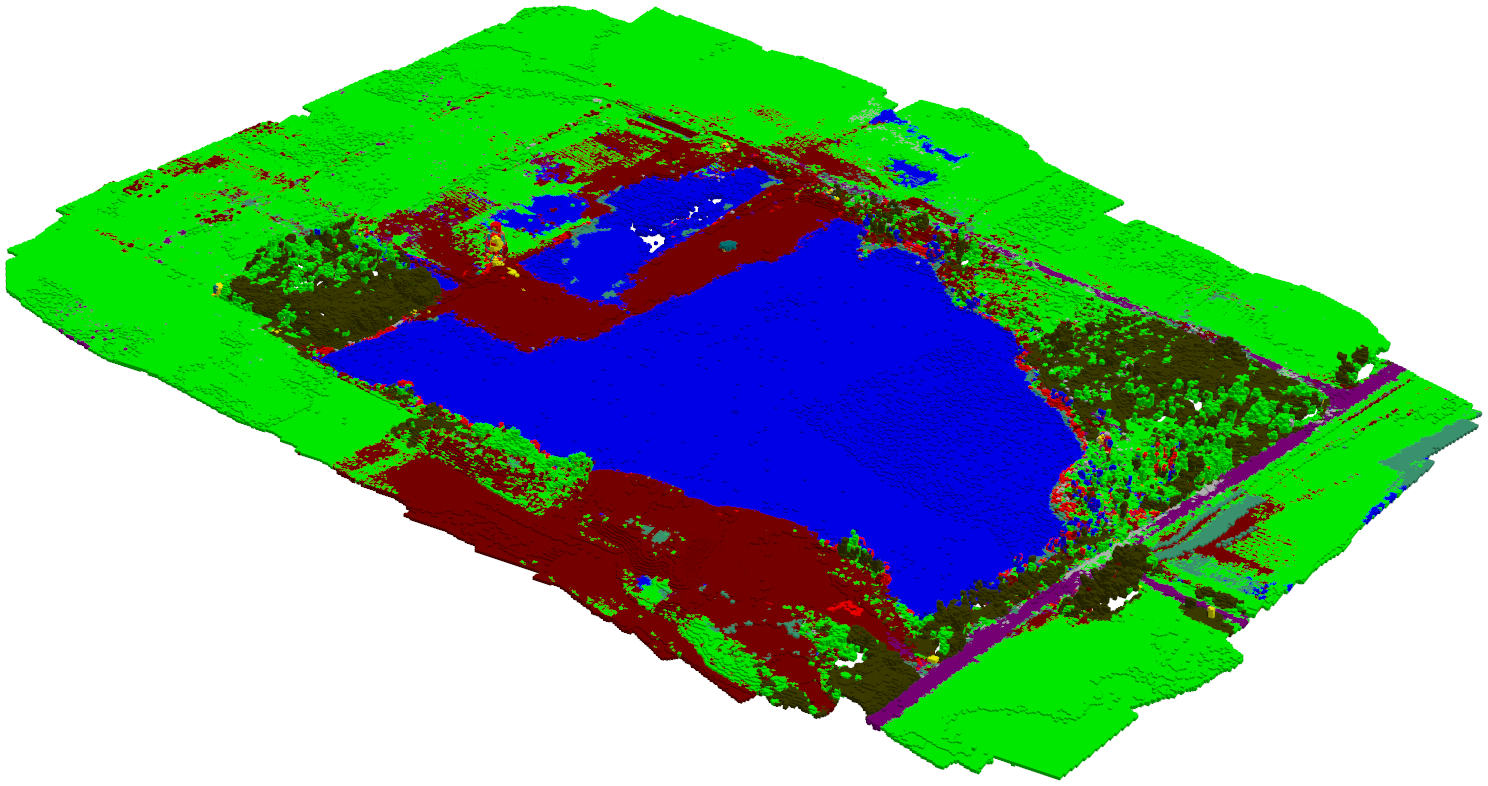} &
        \includegraphics[width=0.24\textwidth]{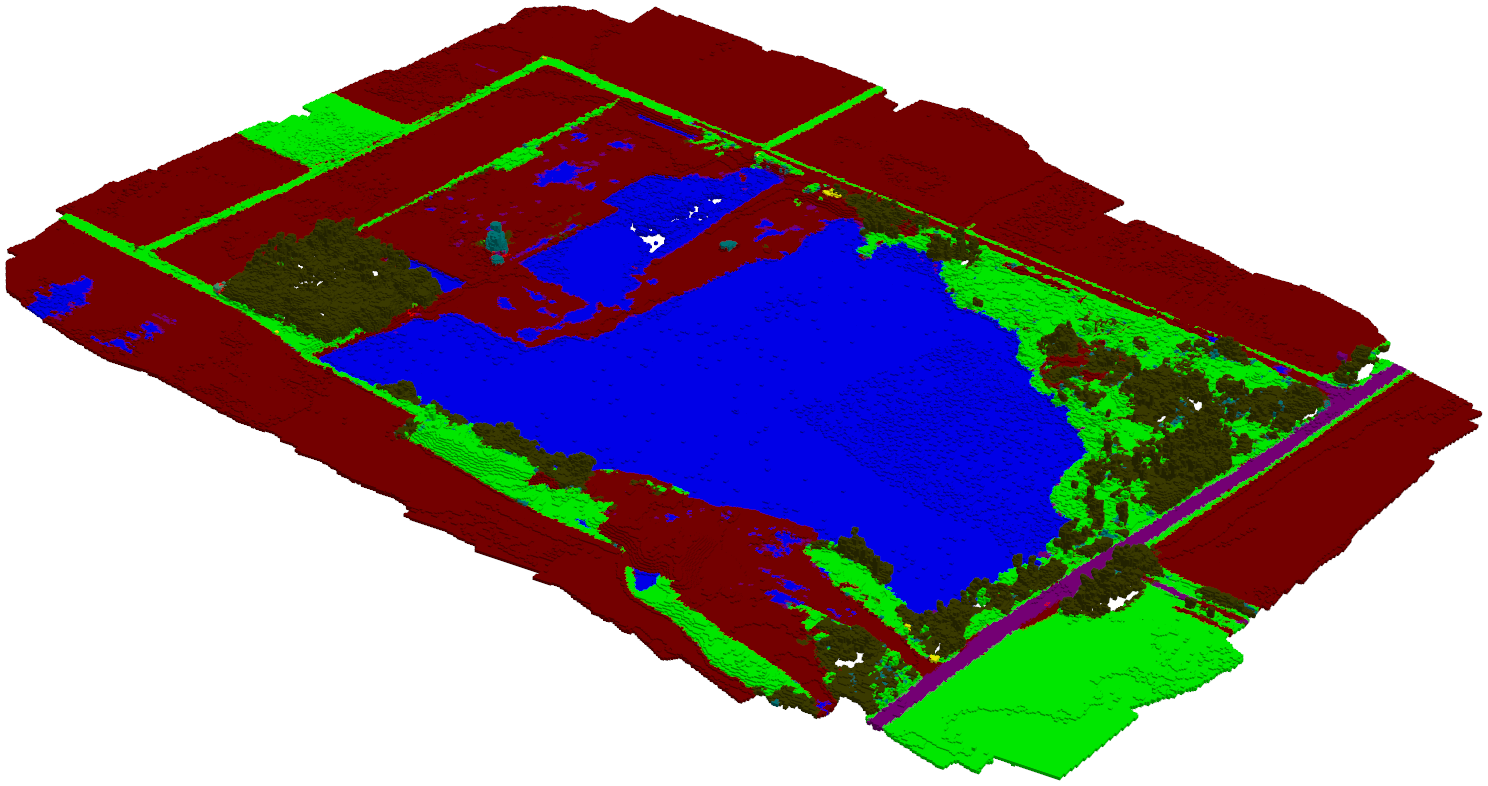}
        \\[1mm]

        \textbf{Scene 2} &
        \includegraphics[width=0.24\textwidth]{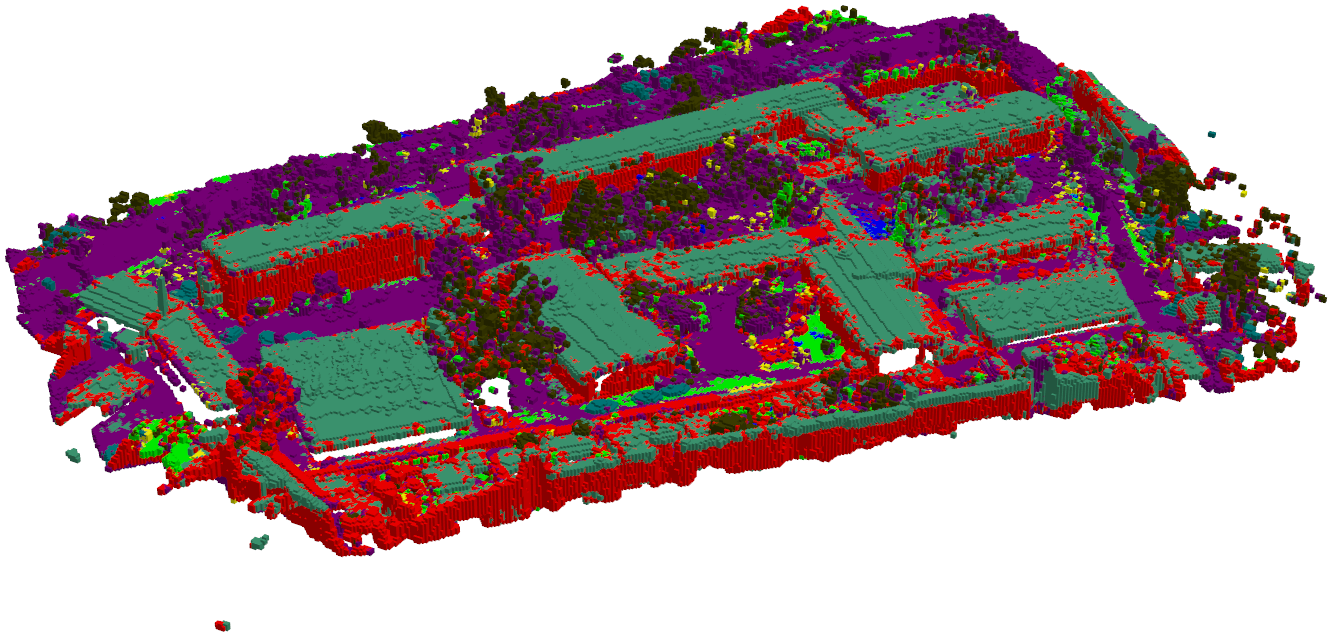} &
        \includegraphics[width=0.24\textwidth]{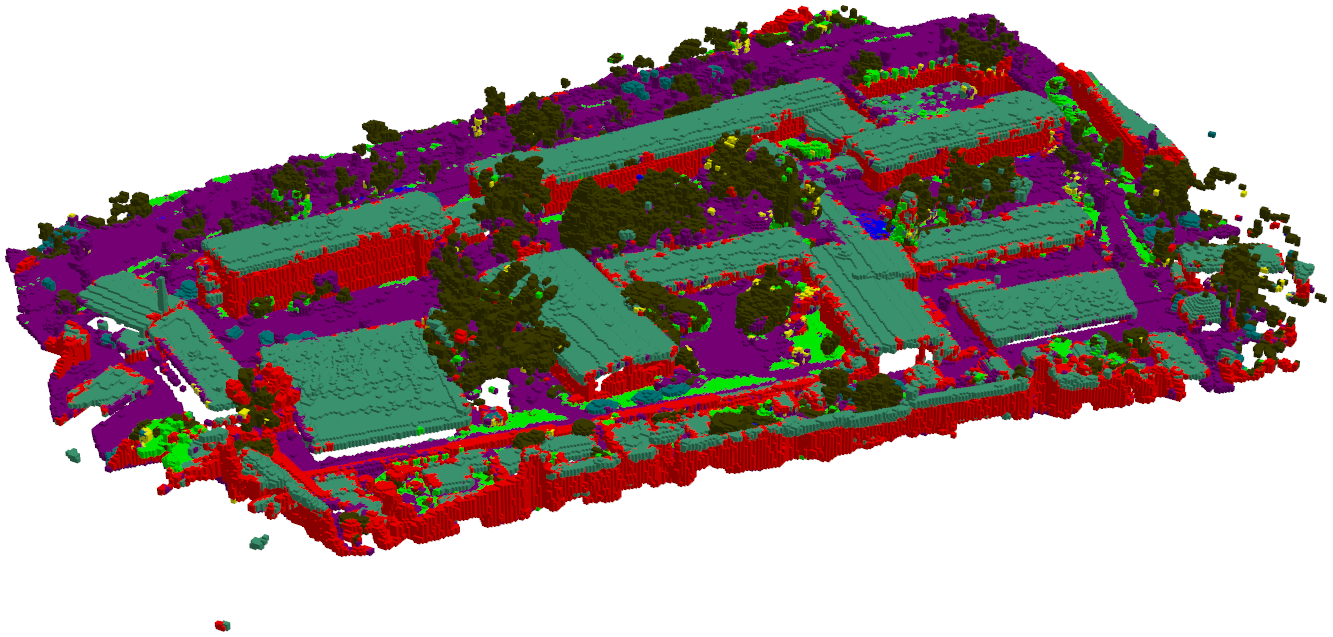} &
        \includegraphics[width=0.24\textwidth]{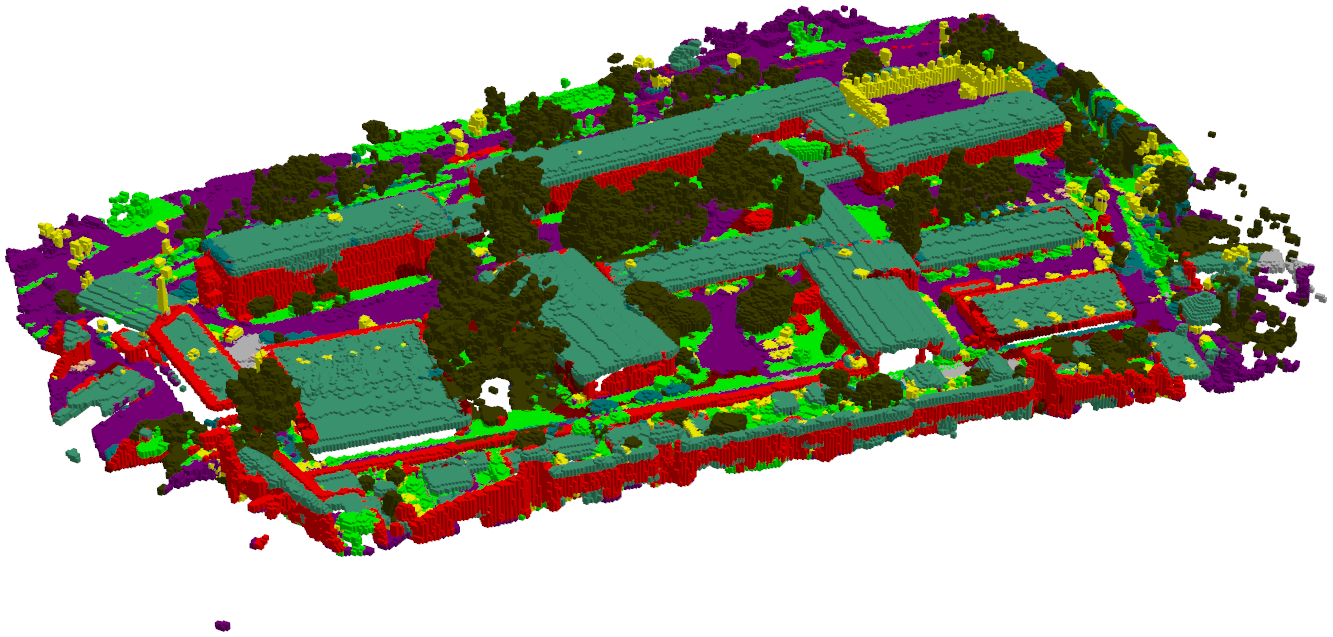}
        \\[1mm]

        \textbf{Scene 3} &
        \includegraphics[width=0.24\textwidth]{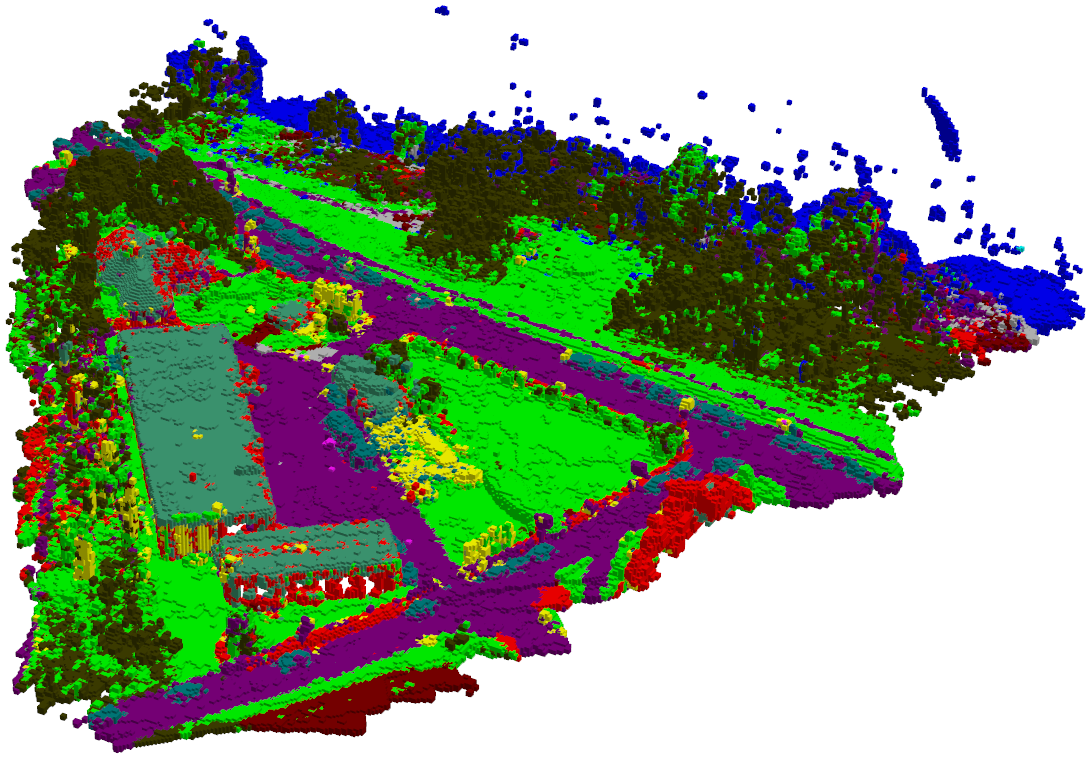} &
        \includegraphics[width=0.24\textwidth]{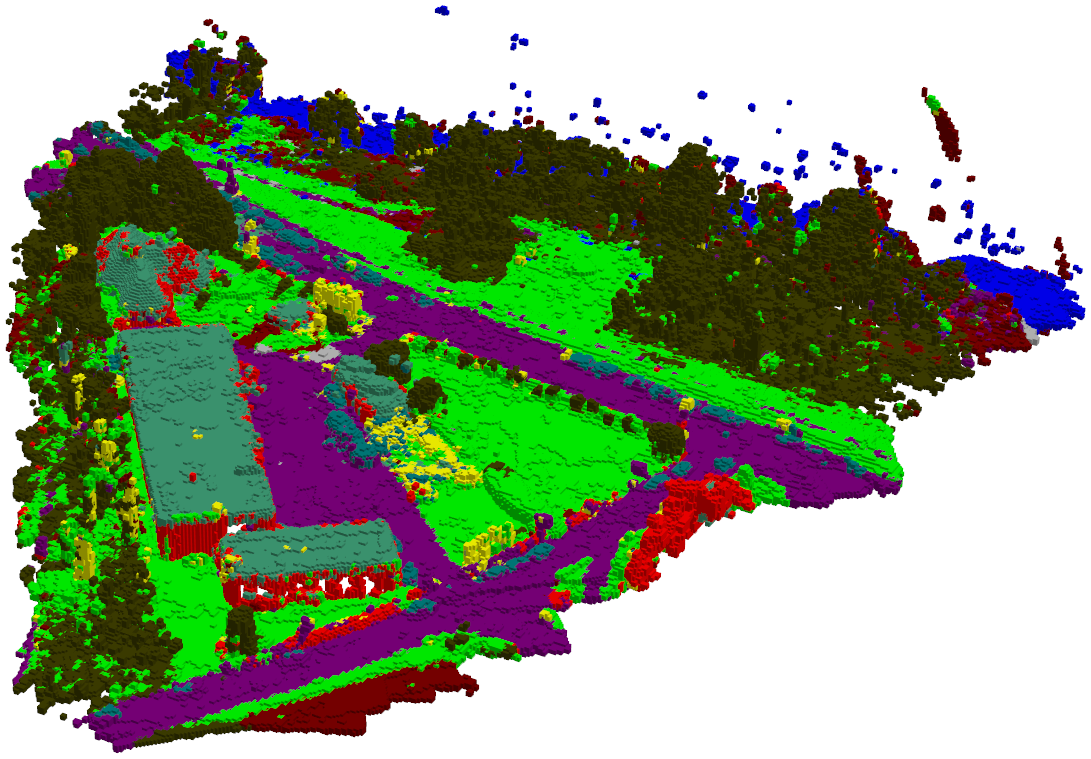} &
        \includegraphics[width=0.24\textwidth]{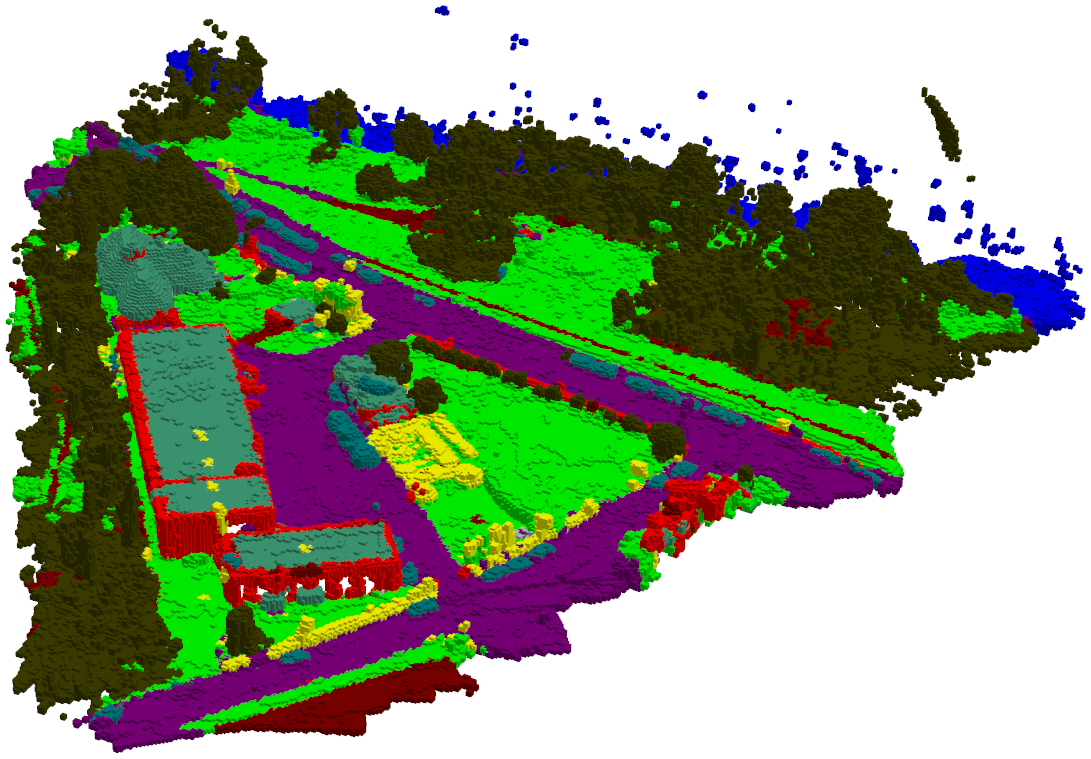}
        \\[1mm]

        \textbf{Scene 4} &
        \includegraphics[width=0.24\textwidth]{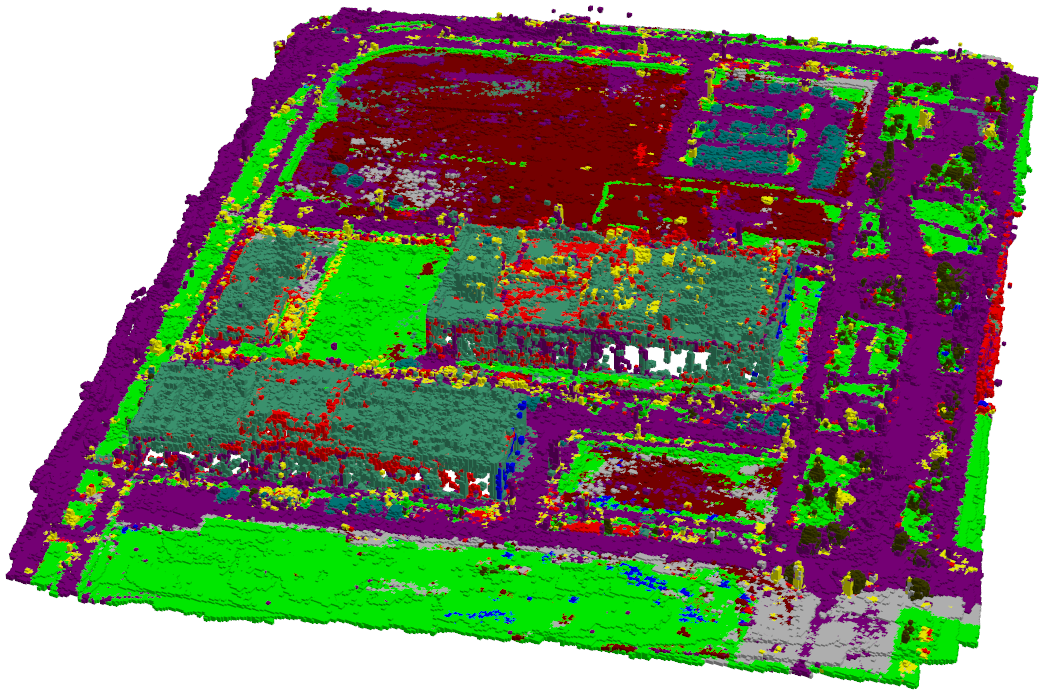} &
        \includegraphics[width=0.24\textwidth]{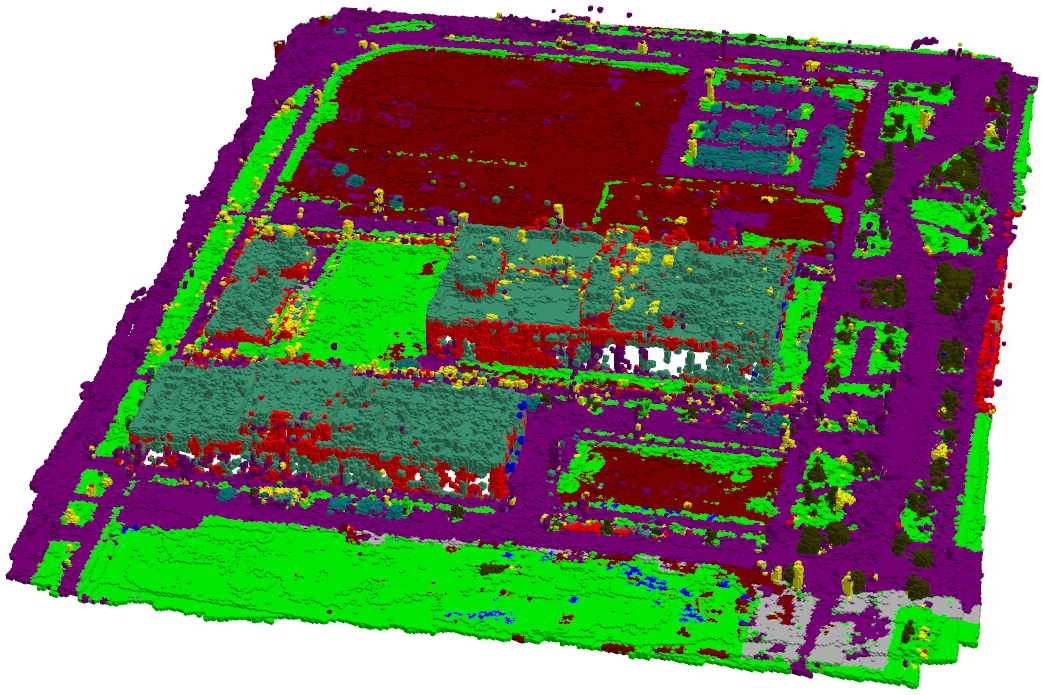} &
        \includegraphics[width=0.24\textwidth]{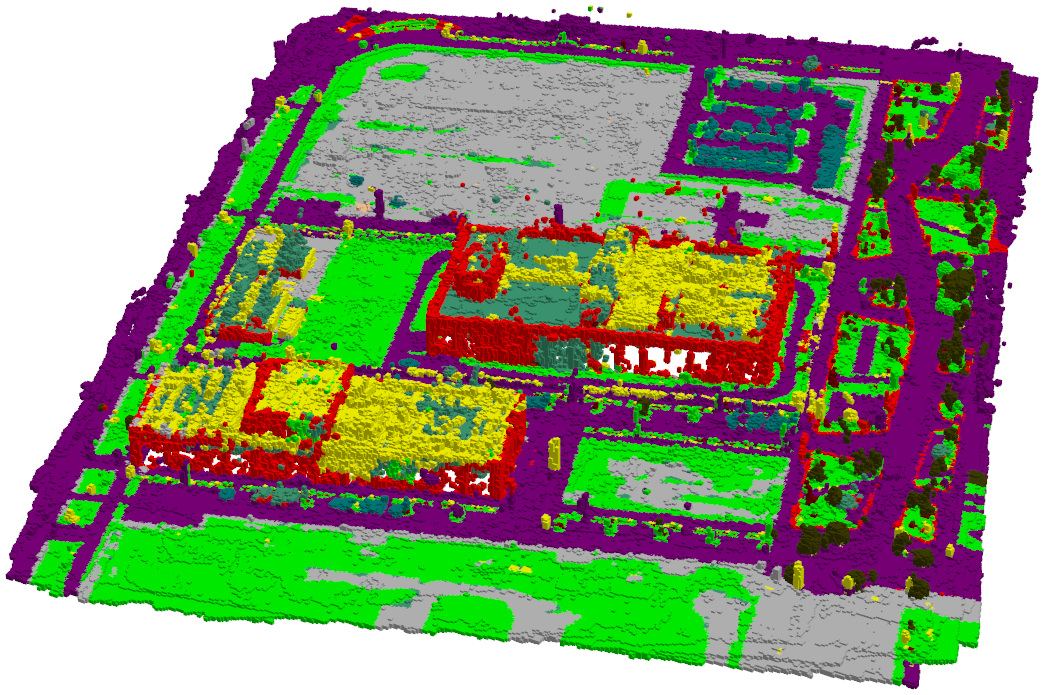}
        \\[1mm]

        \textbf{Scene 5} &
        \includegraphics[width=0.24\textwidth]{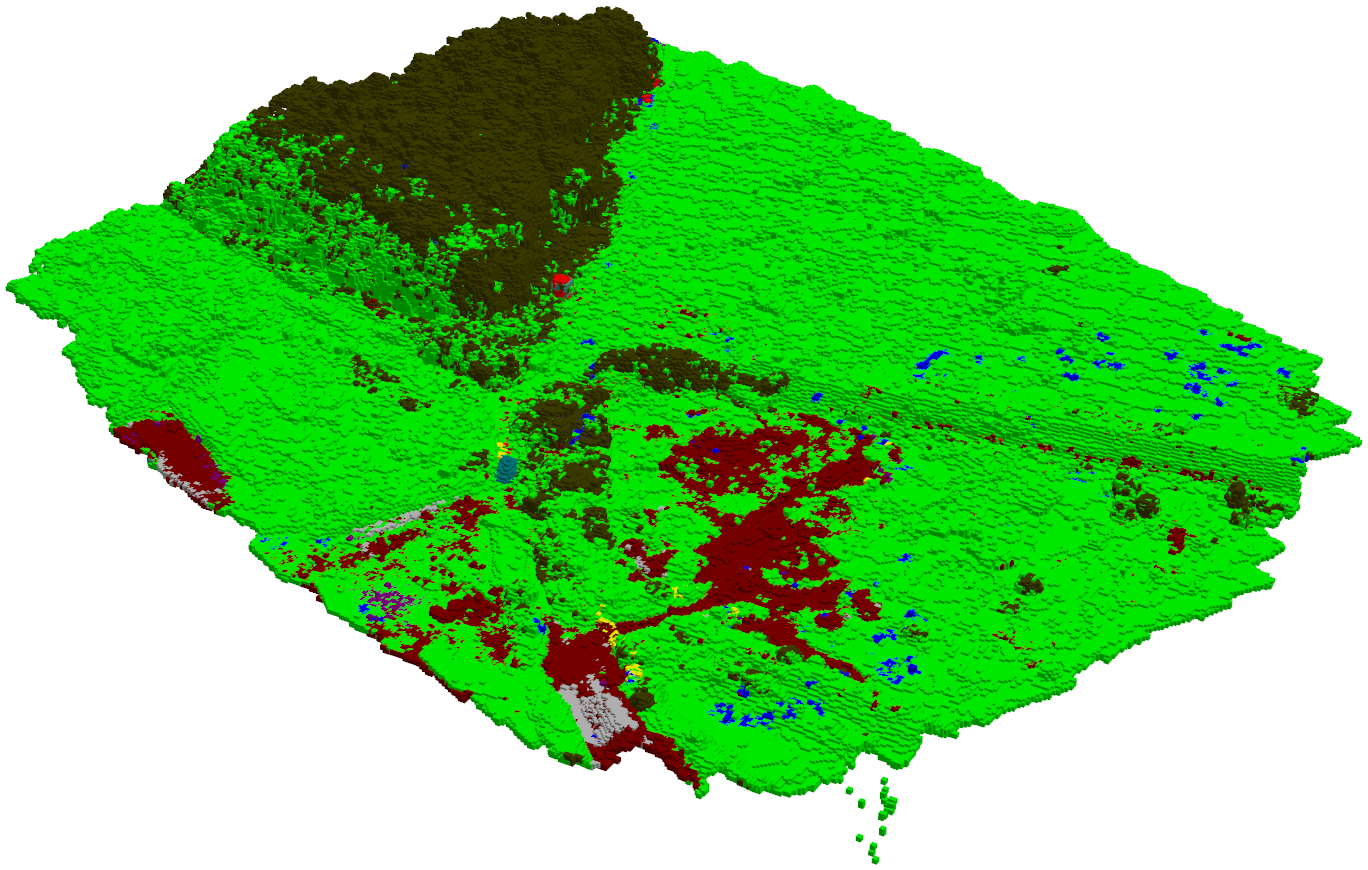} &
        \includegraphics[width=0.24\textwidth]{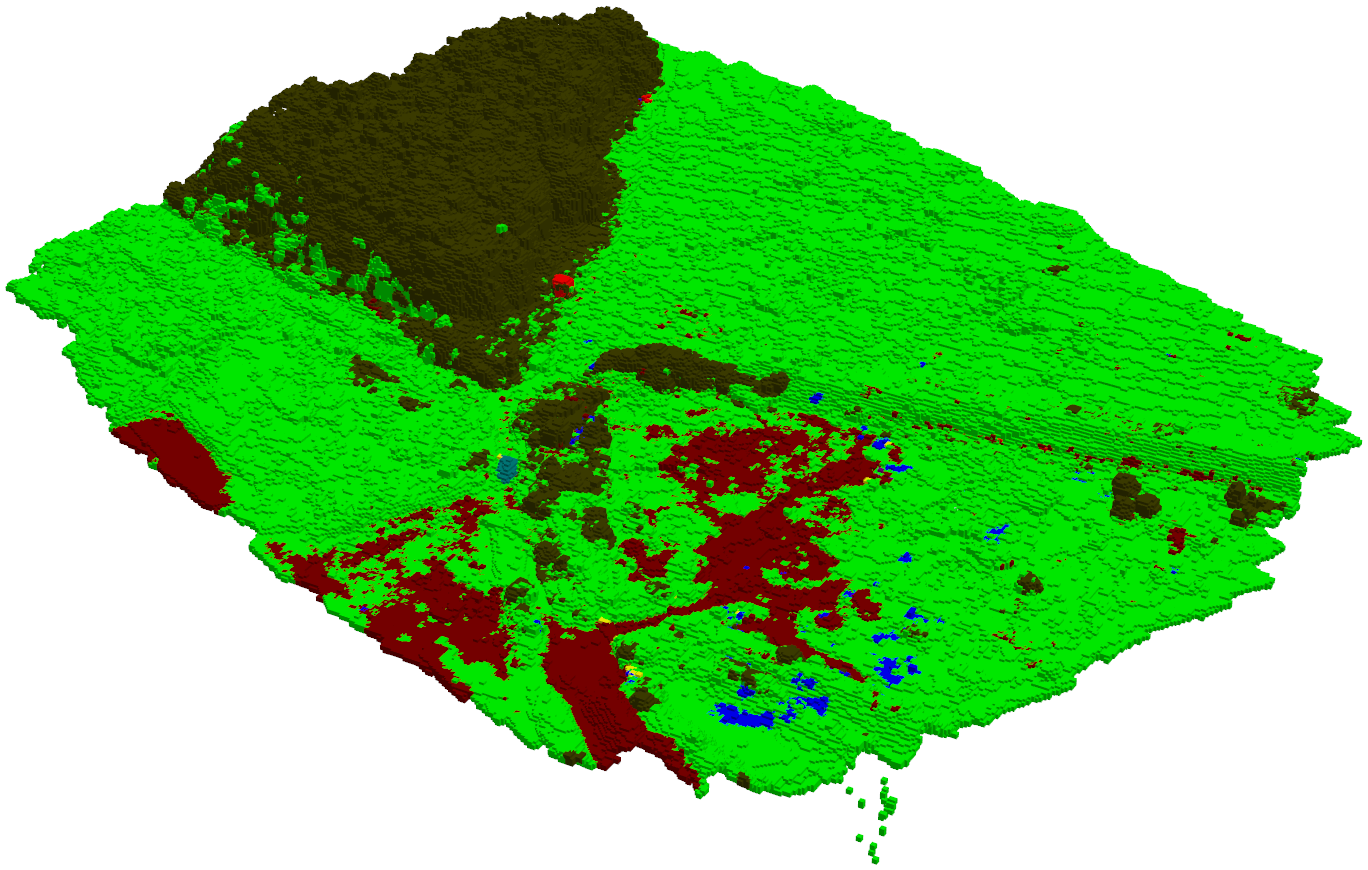} &
        \includegraphics[width=0.24\textwidth]{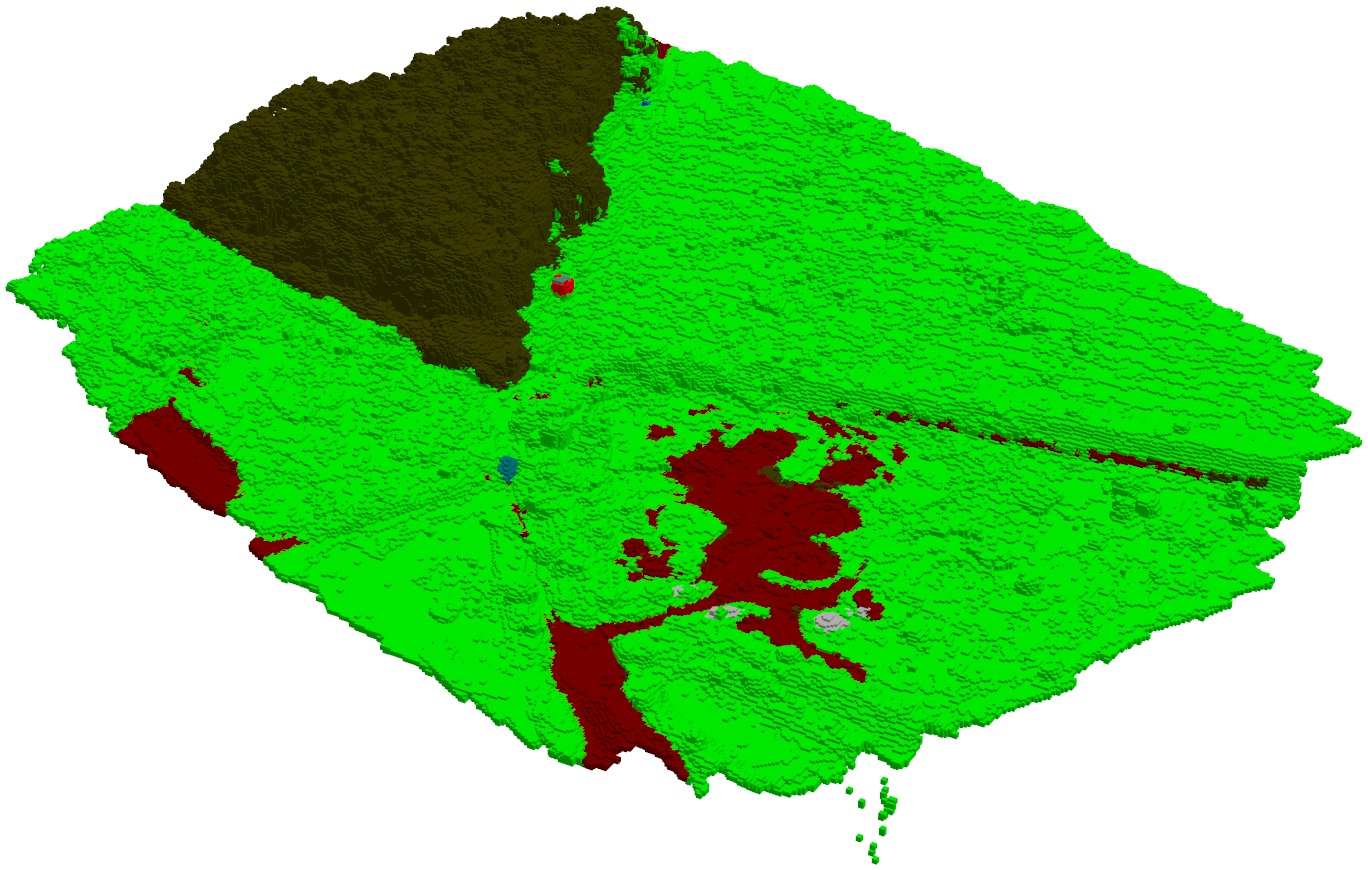}

        \\[1mm]

        \textbf{Scene 6} &
        \includegraphics[width=0.24\textwidth]{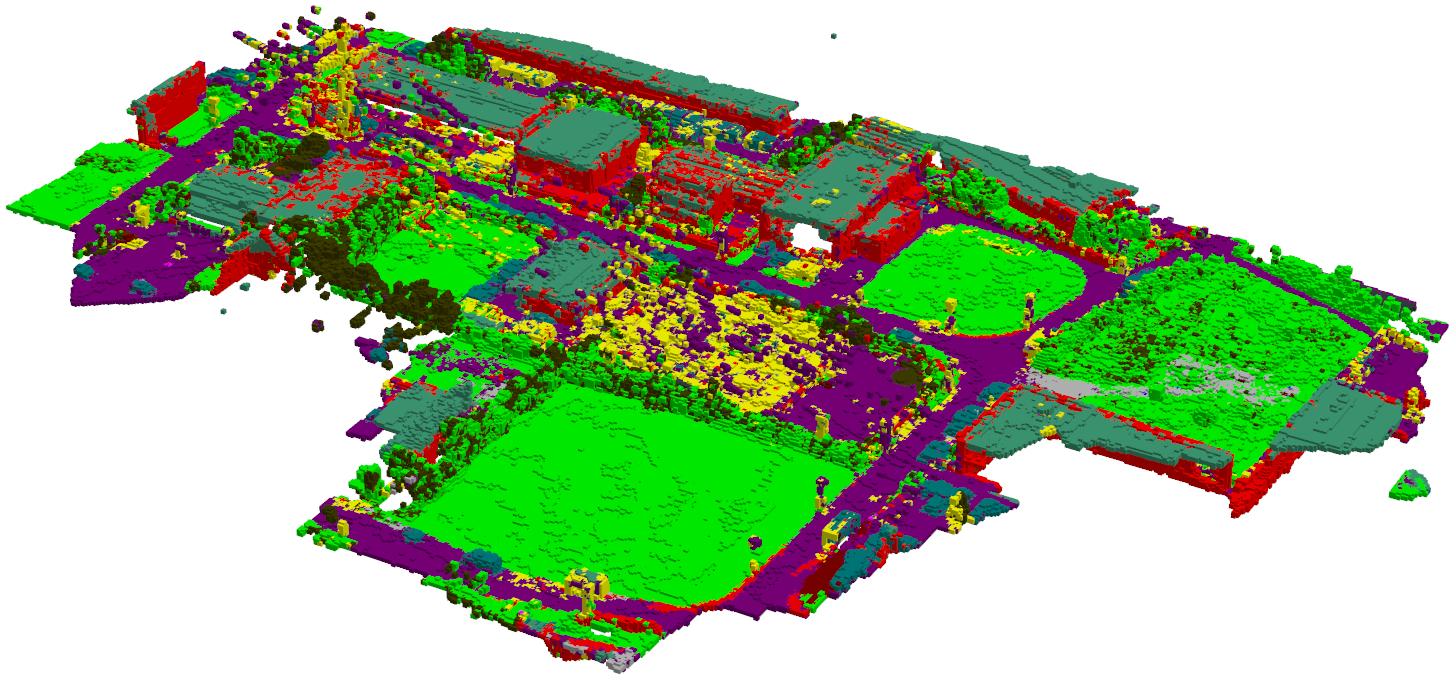} &
        \includegraphics[width=0.24\textwidth]{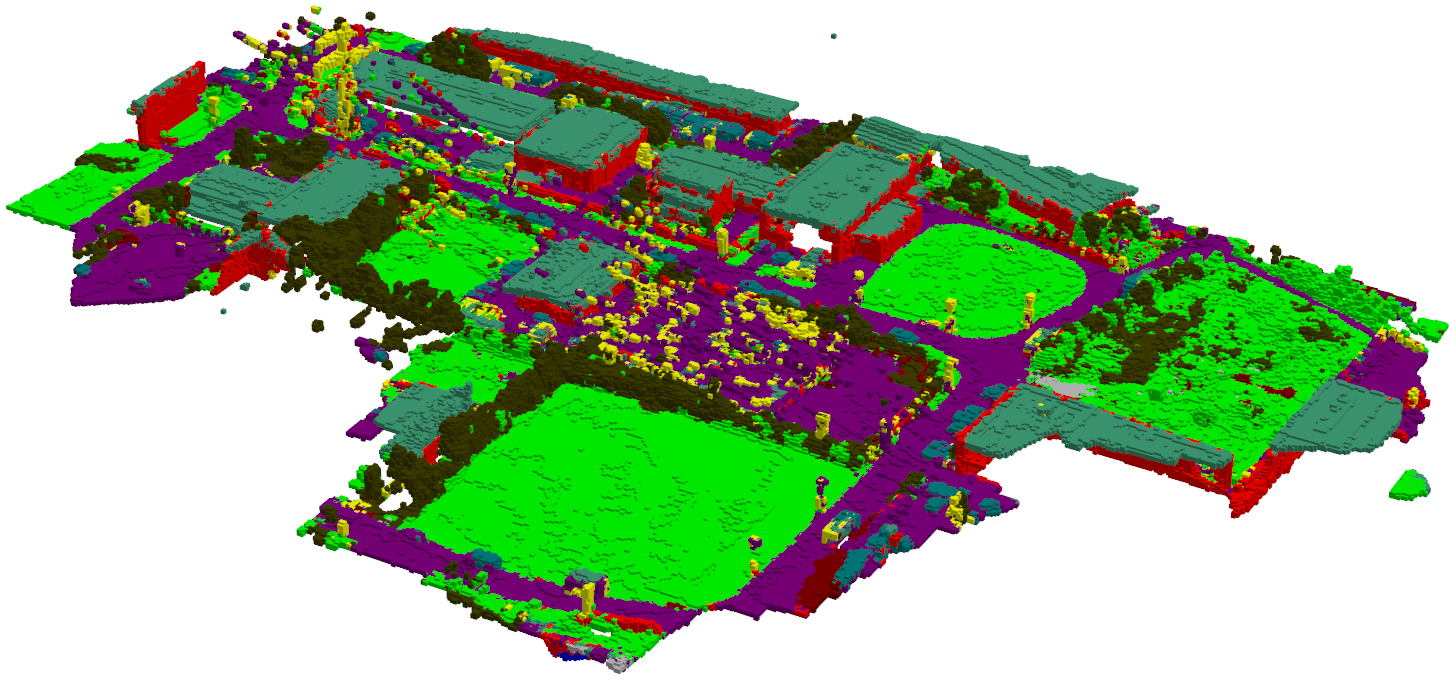} &
        \includegraphics[width=0.24\textwidth]{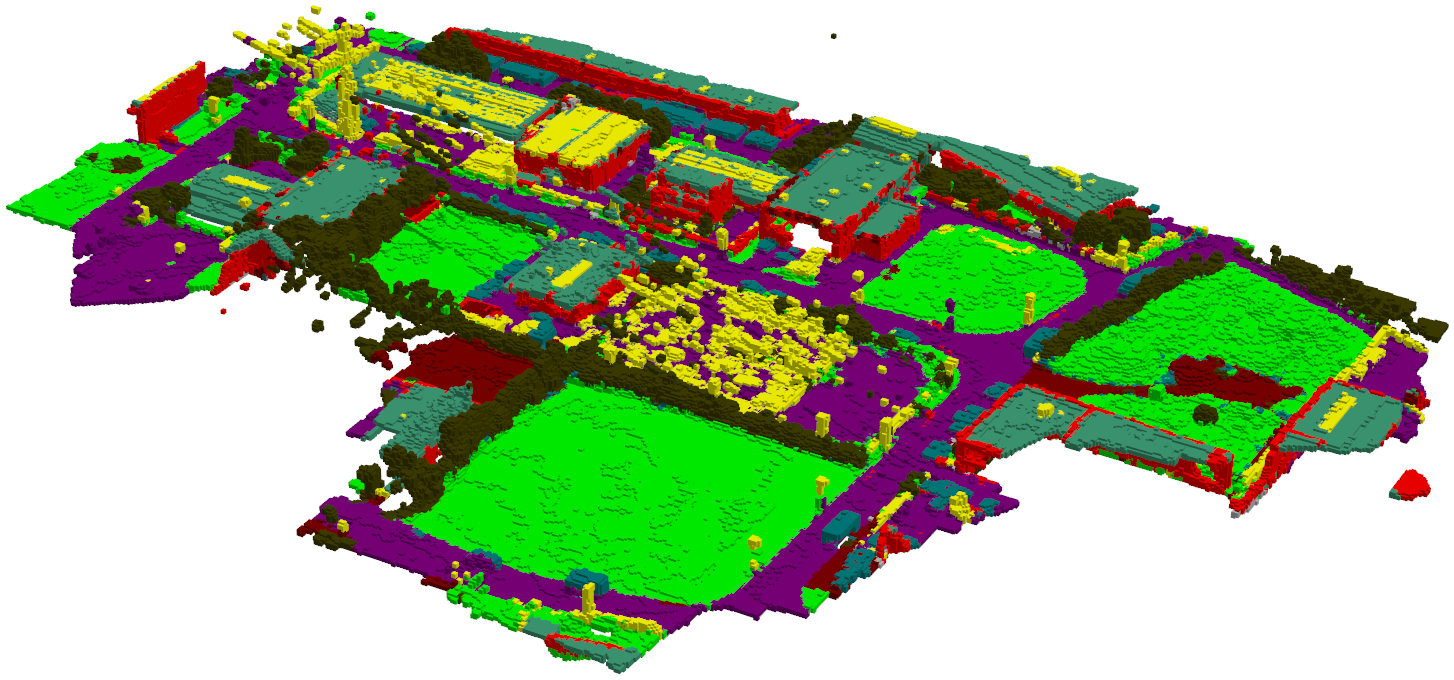}
        \\[1mm]

        \textbf{Scene 7} &
        \includegraphics[width=0.24\textwidth]{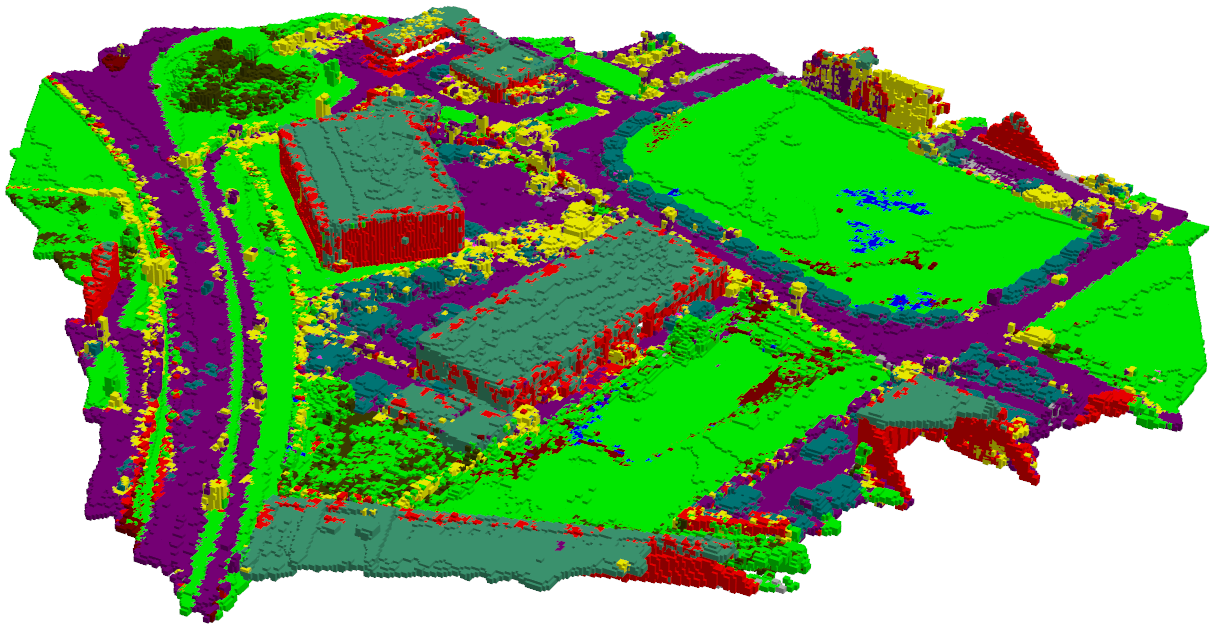} &
        \includegraphics[width=0.24\textwidth]{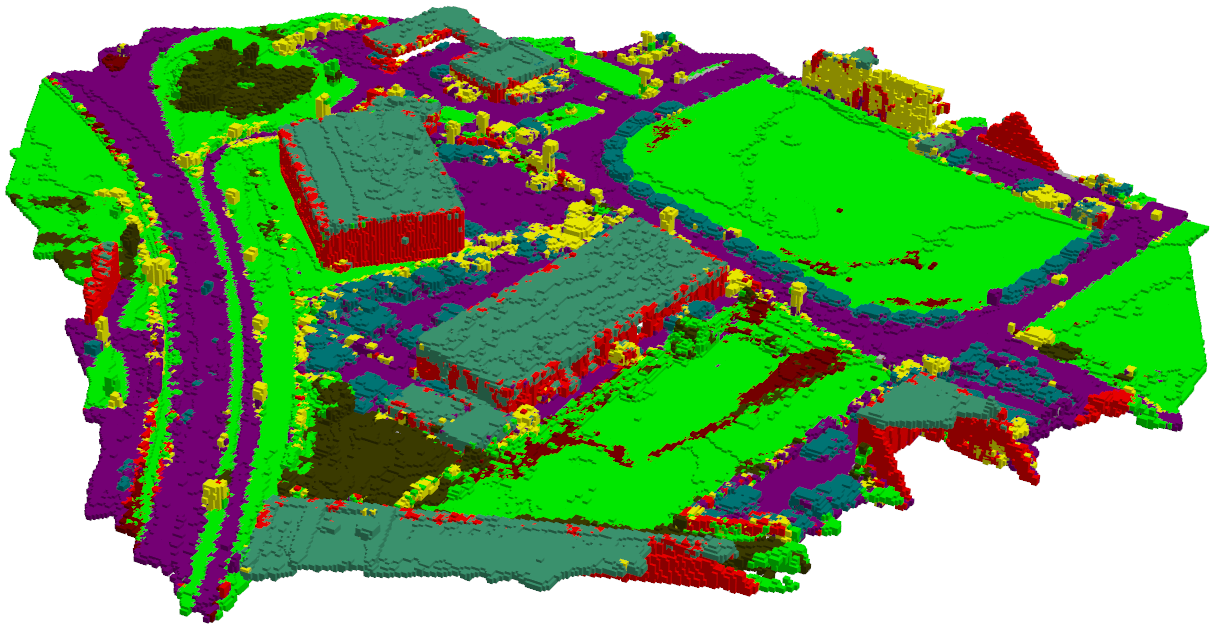} &
        \includegraphics[width=0.24\textwidth]{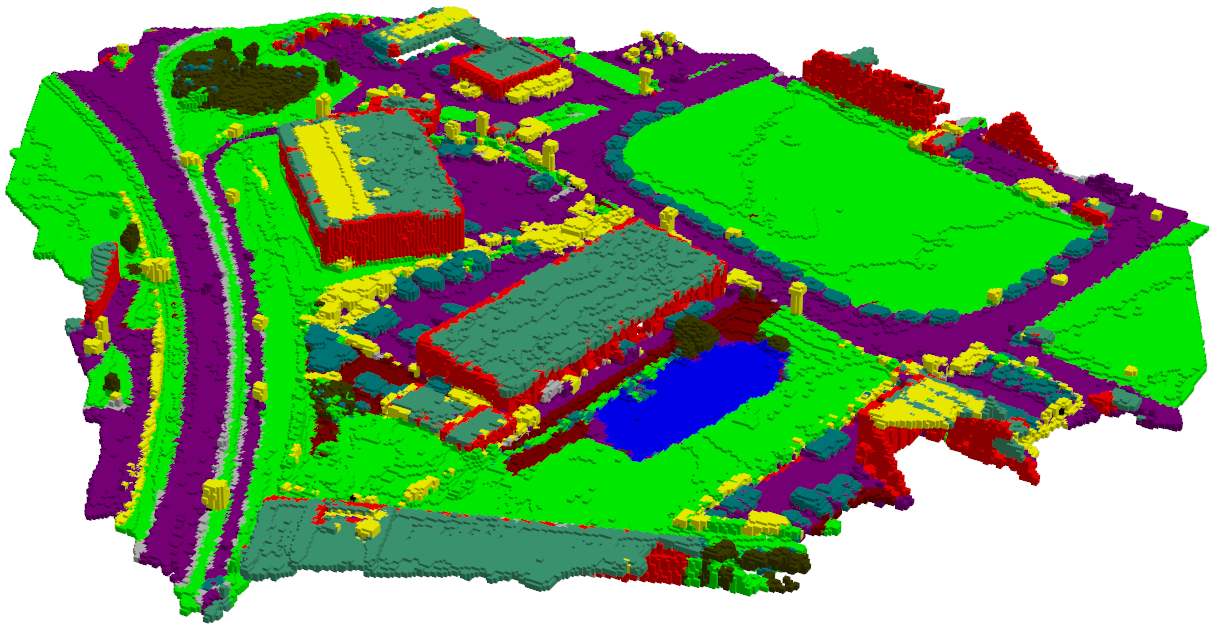}
        \\[1mm]

        \textbf{Scene 9} &
        \includegraphics[width=0.24\textwidth]{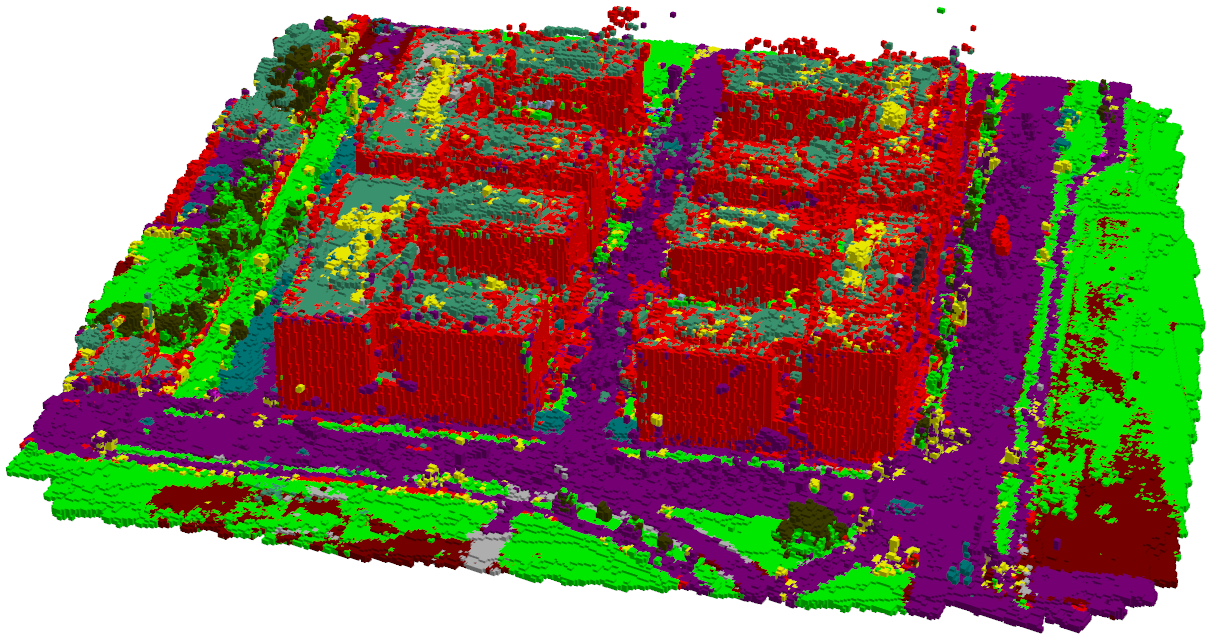} &
        \includegraphics[width=0.24\textwidth]{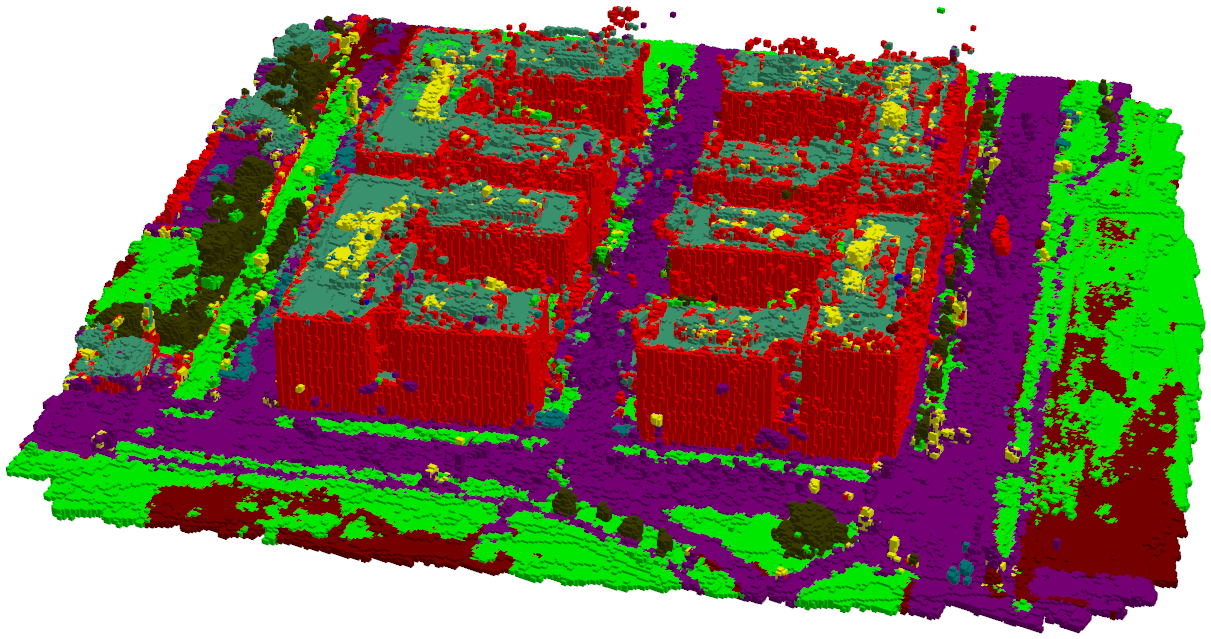} &
        \includegraphics[width=0.24\textwidth]{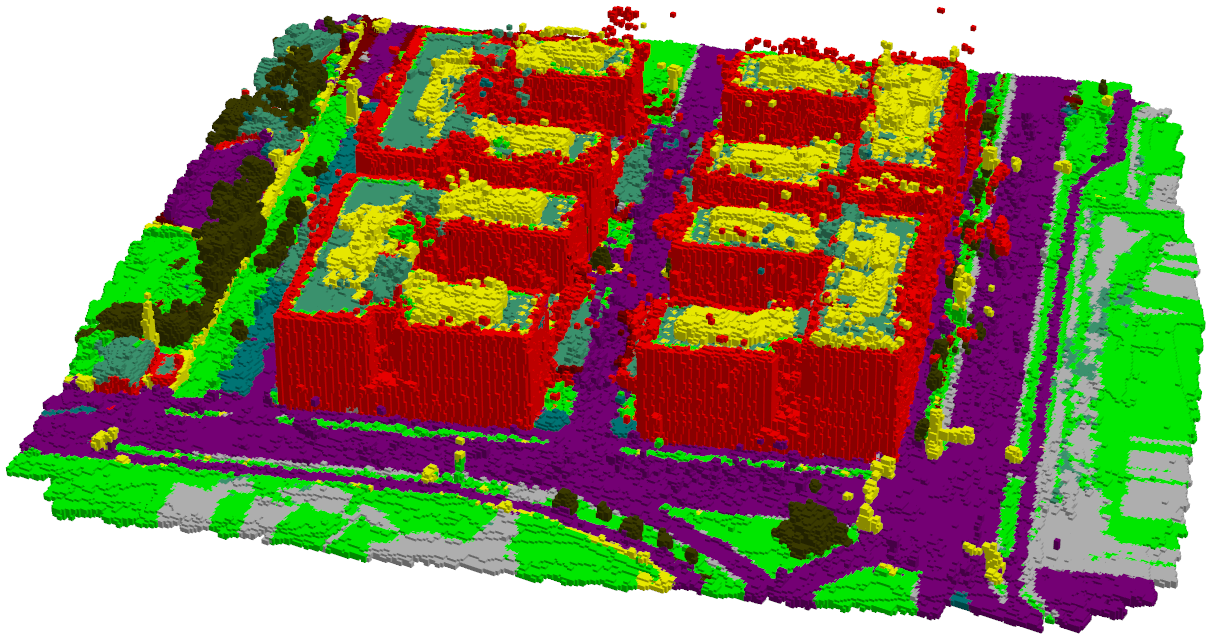}

    \end{tabular}

    \caption{\textbf{Qualitative correction on held-out OccuFly validation
    scenes using SegFormer-MiT-B3 upstream predictions.} Each scene is
    corrected by the model of the split in which it was held out. From left
    to right: uncorrected Radix input, VoxelFix, and ground truth.}
    \label{fig:supp_val_qual}
\end{figure*}


\subsection{MinkUNet}
MinkUNet is the learned map-only baseline, isolating the contribution of graph-based reasoning: it performs voxel-wise correction with sparse volumetric convolutions instead of attention over a $k$-NN graph, and is trained with the same cross-scene splits and the same correction and detection objectives as VoxelFix. Its input is the label map alone, without the geometric descriptor or the frozen embedding, so the comparison in \cref{tab:map_correction} reflects both the architectural difference and the difference in available node features. The configuration is given in \cref{tab:supp_minkunet}.

\begingroup
\setlength{\tabcolsep}{3pt}
\begin{table}[t]
    \centering
    \caption{\textbf{MinkUNet baseline configuration.}}
    \label{tab:supp_minkunet}
    \footnotesize
    \renewcommand{\arraystretch}{0.92}
    \begin{tabular}{@{}p{0.32\linewidth}p{0.62\linewidth}@{}}
        \toprule
        Parameter & Value \\
        \midrule
        Architecture &
        Sparse U-Net (MinkowskiEngine), 4 stride-2 stages, 8 channels
        throughout, additive skips; 20{,}395 parameters \\
        Input features &
        12-D one-hot of the input label map only (no geometry, no
        embeddings) \\
        Voxel size &
        $0.5$\,m \\
        Optimizer &
        AdamW \\
        Learning rate &
        $1\times10^{-3}$, ReduceLROnPlateau ($\times0.5$, patience 5) \\
        Weight decay &
        $1\times10^{-4}$ \\
        Number of epochs &
        200 max., early stop (patience 20), 30-epoch warmup \\
        Loss &
        Weighted cross-entropy $+\,0.5\times$ detection BCE
        (per-map \texttt{pos\_weight}) \\
        Batch size &
        1 scene \\
        Augmentation &
        None \\
        Checkpoint selection &
        Best validation loss \\
        \bottomrule
    \end{tabular}
\end{table}
\endgroup

\begingroup
\setlength{\tabcolsep}{2.5pt}
\renewcommand{\arraystretch}{0.9}
\begin{table*}[b]
    \centering
    \scriptsize
    \caption{\textbf{Per-class IoU (\%) on the independently reconstructed OOD
    scene using single-split Mask2Former upstream predictions.} Only classes
    present in the scene are shown; mIoU is averaged over these 10 classes and
    is therefore not directly comparable to the in-distribution values in
    \cref{tab:perclass_iou}. Best and second-best results are shown in
    \textbf{bold} and \underline{underlined}, respectively.}
    \resizebox{\textwidth}{!}{%
    \begin{tabular}{l|>{\centering\arraybackslash}c>{\centering\arraybackslash}c>{\centering\arraybackslash}c>{\centering\arraybackslash}c>{\centering\arraybackslash}c>{\centering\arraybackslash}c>{\centering\arraybackslash}c>{\centering\arraybackslash}c>{\centering\arraybackslash}c>{\centering\arraybackslash}c|>{\centering\arraybackslash}c}
        \toprule
        \textbf{Method}
        & \textbf{Road}
        & \textbf{Tree}
        & \textbf{Vehicle}
        & \textbf{Grass}
        & \textbf{Wall}
        & \textbf{Roof}
        & \textbf{Obstacle}
        & \textbf{Gravel}
        & \textbf{Person}
        & \textbf{Dirt}
        & \cellcolor{gray!20}\textbf{mIoU} \\
        & \textcolor[RGB]{128,0,128}{\rule{1.5ex}{1.5ex}}
        & \textcolor[RGB]{64,64,0}{\rule{1.5ex}{1.5ex}}
        & \textcolor[RGB]{0,128,128}{\rule{1.5ex}{1.5ex}}
        & \textcolor[RGB]{0,255,0}{\rule{1.5ex}{1.5ex}}
        & \textcolor[RGB]{255,0,0}{\rule{1.5ex}{1.5ex}}
        & \textcolor[RGB]{64,160,120}{\rule{1.5ex}{1.5ex}}
        & \textcolor[RGB]{255,255,0}{\rule{1.5ex}{1.5ex}}
        & \textcolor[RGB]{192,192,192}{\rule{1.5ex}{1.5ex}}
        & \textcolor[RGB]{255,16,255}{\rule{1.5ex}{1.5ex}}
        & \textcolor[RGB]{128,0,0}{\rule{1.5ex}{1.5ex}}
        & \\
        \midrule
        Radix (input)
        & 76.57
        & 5.65
        & 23.07
        & 87.41
        & 34.92
        & 54.92
        & 21.68
        & 0.00
        & 5.00
        & 0.28
        & \cellcolor{gray!20}30.95 \\
        KNN
        & 78.75
        & 6.52
        & 26.33
        & 88.50
        & \underline{35.75}
        & 56.32
        & \textbf{23.25}
        & 0.00
        & 3.95
        & \textbf{0.47}
        & \cellcolor{gray!20}\underline{31.98} \\
        CRF
        & \textbf{80.35}
        & \underline{8.91}
        & \textbf{30.56}
        & \textbf{90.31}
        & 32.38
        & \textbf{57.84}
        & 19.13
        & 0.00
        & 0.00
        & 0.00
        & \cellcolor{gray!20}31.95 \\
        Geometry Heuristic
        & 78.94
        & 5.14
        & 25.90
        & 85.67
        & 33.68
        & \underline{56.50}
        & 19.94
        & 0.00
        & \underline{5.00}
        & 0.27
        & \cellcolor{gray!20}31.10 \\
        MinkUNet
        & \underline{79.18}
        & 6.03
        & 26.53
        & 87.35
        & 34.92
        & 55.01
        & 20.72
        & 0.00
        & \textbf{5.75}
        & \underline{0.28}
        & \cellcolor{gray!20}31.58 \\
        \midrule
        VoxelFix (Ours)
        & 77.70
        & \textbf{10.16}
        & \underline{27.32}
        & \underline{90.30}
        & \textbf{39.64}
        & 55.66
        & \underline{20.75}
        & 0.00
        & 4.76
        & \underline{0.28}
        & \cellcolor{gray!20}\textbf{32.66} \\
        \bottomrule
    \end{tabular}%
    }
    \label{tab:supp_ood_classwise}
\end{table*}
\endgroup

\section{Additional In-Distribution Results}
\label{sec:supp_id_results}

\subsection{Qualitative Results on Validation Scenes}
\cref{fig:supp_val_qual} shows additional corrections on the OccuFly
validation scenes using SegFormer-MiT-B3 upstream predictions. Each scene is
corrected by the model of the split in which it was held out, so the eight examples come from eight independently trained models.

\section{Additional Out-of-Distribution Results}
\label{sec:supp_ood}

\subsection{OOD Scene Overview}
To assess whether the learned correction transfers beyond the environments
seen during training, we reconstruct an additional small aerial scene that is
not part of OccuFly. Imagery was captured with an EmQopter Q6500 platform
carrying a FLIR Blackfly S GigE (BFS-PGE-31S4C-C) RGB camera with a Tamron
M112FM06 lens, georeferenced by an OXTS xRED3000 INS. The scene was flown at
$30$\,m above ground in late autumn under overcast conditions, covering
approximately $50\times140$\,m ($\sim$7{,}000\,m$^2$). Camera poses and depth maps are obtained by photogrammetric reconstruction, and the ground-truth semantic voxel map is generated with the same 2D-annotation and
label-transfer procedure used to build OccuFly.

\subsection{Additional OOD Quantitative Results}
\cref{tab:supp_ood_classwise} decomposes the OOD result by semantic class.
The gain is concentrated in tree ($+4.51$) and wall ($+4.72$), the same
classes that improve most in distribution, while the smoothing baselines
remain competitive on the dominant classes. Because the OOD scene is
evaluated with a single model rather than the eight cross-validation splits,
no variance is reported.

\subsection{Additional OOD Qualitative Results}
\begin{figure*}[b]
    \centering
    \begin{tabular}{cccc}
        & \textbf{Radix Input} & \textbf{VoxelFix} & \textbf{GT} \\[1mm]

        \textbf{OOD Scene} &
        \includegraphics[width=0.25\textwidth]{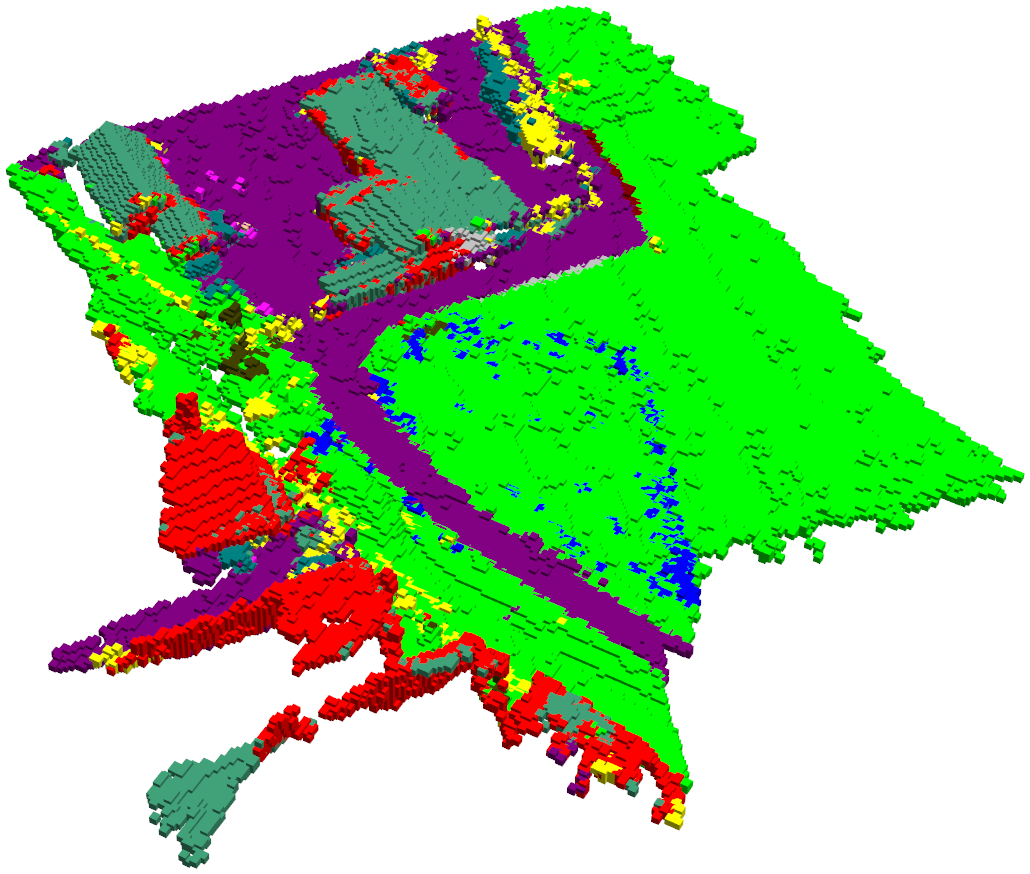} &
        \includegraphics[width=0.25\textwidth]{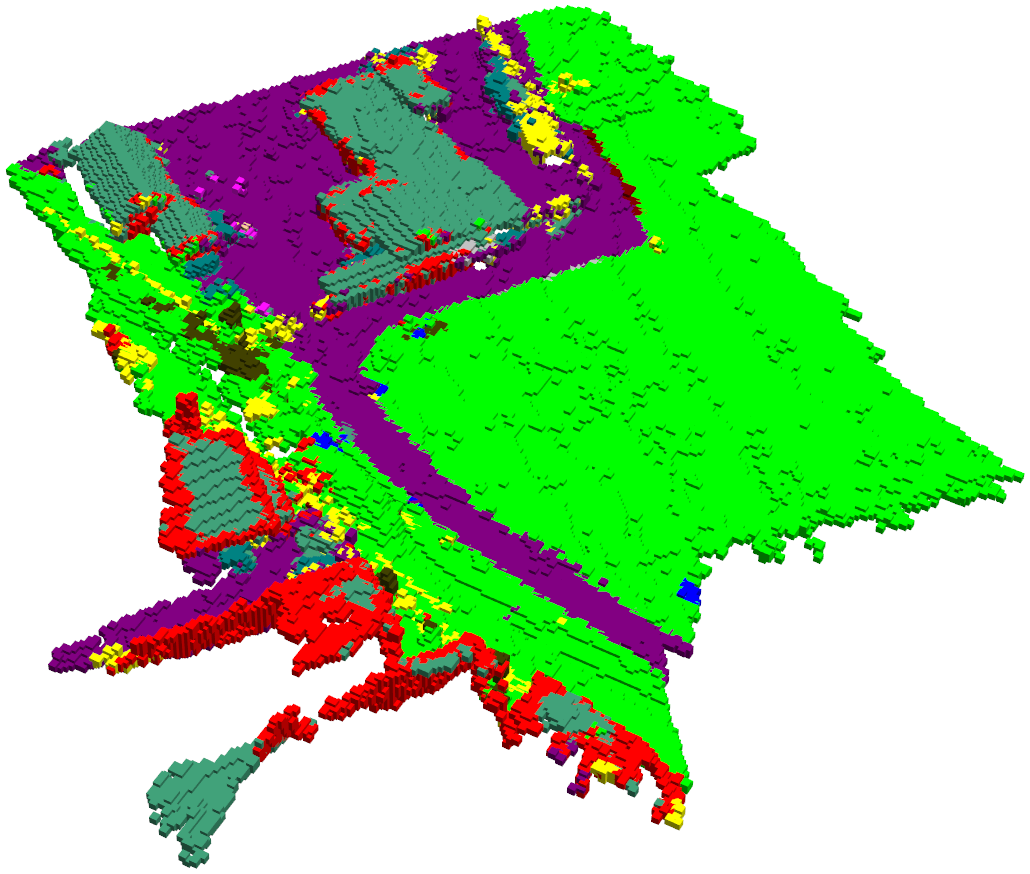} &
        \includegraphics[width=0.25\textwidth]{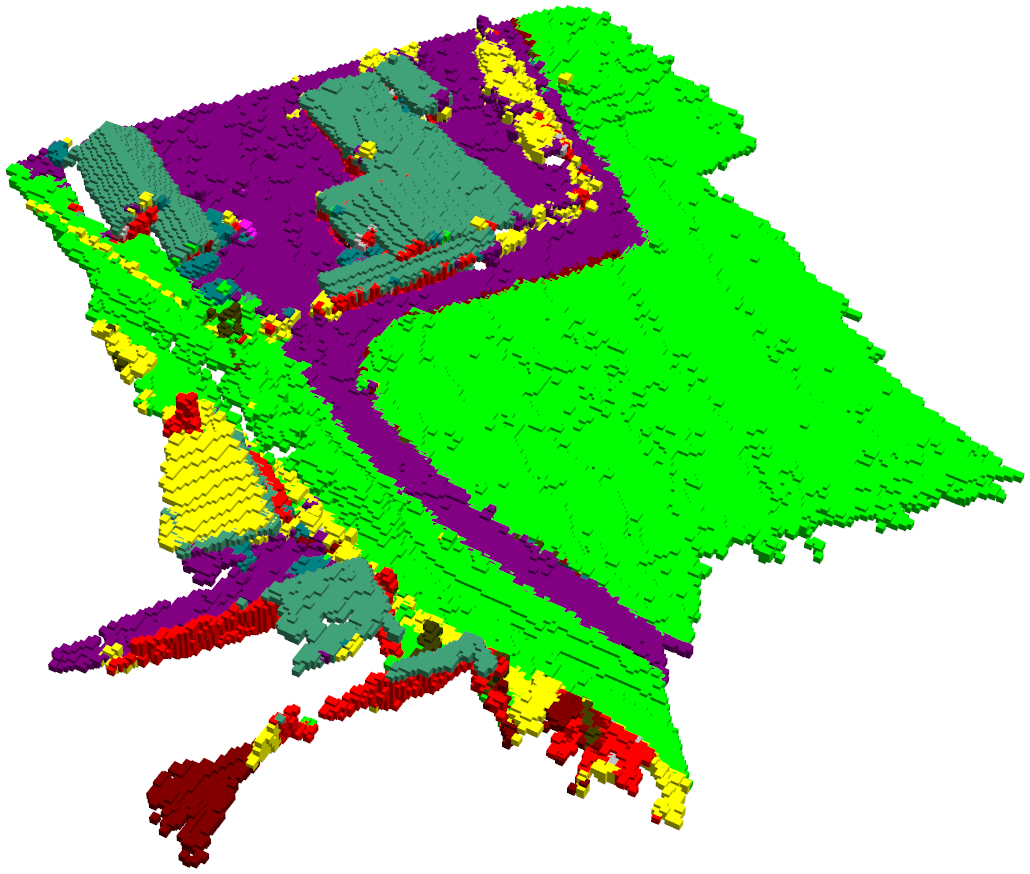}
        \\[2mm]
    \end{tabular}
    \caption{\textbf{Qualitative correction on the independently reconstructed OOD scene using Mask2Former upstream predictions.} From left to right: uncorrected Radix input, VoxelFix, and ground truth. Wall and roof regions are recovered along the building structures, and spurious water labels on the grass surface are removed. Remaining errors concentrate in structures for which the completed map carries no distinguishing evidence.}
    \label{fig:supp_ood_qual}
\end{figure*}

\cref{fig:supp_ood_qual} shows the corrected map for the full scene. Wall and roof regions are recovered along the building structures, and spurious water labels are removed from the grass surface. Remaining errors concentrate in structures for which the completed map carries no distinguishing evidence.

\section{Cross-Scene Evaluation Protocol}
\label{sec:supp_splits}

\subsection{Exact Scene Splits}

\cref{tab:supp_splits} lists the eight cross-scene splits. Scene 8 is
held out as the fixed in-distribution test scene and is never used for
training, validation, checkpoint selection, or the estimation of any
training-derived quantity. Each of the remaining eight OccuFly scenes serves
once as the validation scene, with the other seven forming the corresponding
training split, giving eight independently trained VoxelFix models that are
all evaluated on the same test map. Reported results are the mean and
standard deviation across these eight models; because the corruption seed is
fixed, the reported variance reflects the choice of training and validation
scenes rather than initialization variance.

\begin{table*}[t]
    \centering
    \caption{\textbf{OccuFly train/validation/test composition for the eight 7/1/1 cross-scene splits.}}
    \label{tab:supp_splits}
    \small
    \setlength{\tabcolsep}{5pt}
    \renewcommand{\arraystretch}{1.05}

    \begin{tabular}{clll}
        \toprule
        Split & Training scenes & Validation scene & Fixed test scene \\
        \midrule
        1 & 1, 2, 7, 3, 9, 4, 5 & 6 & 8 \\
        2 & 6, 2, 7, 3, 9, 4, 5 & 1 & 8 \\
        3 & 6, 1, 7, 3, 9, 4, 5 & 2 & 8 \\
        4 & 6, 1, 2, 3, 9, 4, 5 & 7 & 8 \\
        5 & 6, 1, 2, 7, 9, 4, 5 & 3 & 8 \\
        6 & 6, 1, 2, 7, 3, 4, 5 & 9 & 8 \\
        7 & 6, 1, 2, 7, 3, 9, 5 & 4 & 8 \\
        8 & 6, 1, 2, 7, 3, 9, 4 & 5 & 8 \\
        \bottomrule
    \end{tabular}
\end{table*}

\subsection{Training-Derived Statistics}

For every split, the following quantities are estimated exclusively from its
seven training scenes:
\begin{itemize}
    \item geometric feature-normalization statistics;
    \item class co-occurrence matrix $\mathbf{A}$;
    \item initial geometric class prototypes
          $(\boldsymbol{\mu}_c,\boldsymbol{\sigma}_c)$;
    \item confusion-derived class-transition matrix $\mathbf{T}$; and
    \item class-frequency weights used by the correction objective.
\end{itemize}

The frozen geometry encoder is the one exception: it is trained once, on the
partition of split~1, and shared across all splits (\cref{sec:supp_geo_encoder}).
The fixed test scene is excluded from its training as well, so no quantity
used at inference depends on the test map.

\end{document}